\documentclass{article}

\usepackage[utf8]{inputenc} 
\usepackage[T1]{fontenc}    
\usepackage[draft]{hyperref}       
\usepackage{url}            
\usepackage{booktabs}       
\usepackage{amsfonts, amsmath, amssymb}       
\usepackage{nicefrac}       
\usepackage{microtype}      

\usepackage{zref,zref-xr,zref-user}
\zexternaldocument*{supplementary}

\PassOptionsToPackage{numbers, compress}{natbib}

\usepackage[final]{neurips_2019}

\usepackage{graphicx}

\usepackage{subcaption}

\usepackage{bm}

\usepackage{algorithm}
\usepackage{algorithmic}

\usepackage{multicol}

\newcommand{\N}{\mathcal{N}}
\newcommand{\bigO}{\mathcal{O}}

\newcommand{\swa}{\text{SWA}}
\newcommand{\diag}{\text{diag}}
\newcommand{\lr}{\text{low-rank}}

\usepackage{color}

\title{A Simple Baseline for Bayesian Uncertainty \\ in Deep Learning}

\author{
	\parbox{\linewidth}{
		\centering
		Wesley J. Maddox\thanks{Equal contribution. Correspondence to wjm363 AT nyu.edu}\, $^1$\quad
		Timur Garipov$^{*2}$ \quad
		Pavel Izmailov$^{*1}$\quad \\
	    Dmitry Vetrov$^{2,3}$\quad
		Andrew Gordon Wilson$^{1}$
	}\\
	~\\
	\parbox{\linewidth}{
		\centering
		$^1$ New York University\\
		$^2$ Samsung AI Center Moscow\\
		$^3$ Samsung-HSE Laboratory, National Research University Higher
        School of Economics
	}
}

\begin{document}
\maketitle
\begin{abstract}
	We propose SWA-Gaussian (SWAG), a simple, scalable, and general purpose approach for uncertainty representation and calibration in deep learning. Stochastic Weight Averaging (SWA), which computes the first moment of stochastic gradient descent (SGD) iterates with a modified learning rate schedule, has recently been shown to improve generalization in deep learning.
	With SWAG, we fit a Gaussian using the SWA solution as the first moment and a low rank plus diagonal covariance also derived from the SGD iterates, forming an approximate posterior distribution over neural network weights; we then sample from this Gaussian distribution to perform Bayesian model averaging. 
	We empirically find that SWAG approximates the shape of the true posterior, in accordance with results describing the stationary distribution of SGD iterates. 
	Moreover, we demonstrate that SWAG performs well on a wide variety of 
	tasks, including out of sample detection, calibration, and transfer learning, 
	in comparison to many popular alternatives including MC dropout, KFAC Laplace, 
	SGLD, and temperature scaling.
\end{abstract}

\section{Introduction}

Ultimately, machine learning models are used to make decisions. Representing uncertainty 
is crucial for decision making. For example, in medical diagnoses and autonomous 
vehicles we want to protect against rare but costly mistakes. Deep learning models 
typically lack a representation of uncertainty, and provide overconfident and 
miscalibrated predictions \citep[e.g.,][]{kendall_what_2017, guo_calibration_2017}. 

Bayesian methods provide a natural probabilistic representation of uncertainty 
in deep learning \citep[e.g.,][]{blundell_weight_2015, kingma_variational_2015, chen_stochastic_2014}, and previously had been a gold standard for inference with neural networks \citep{neal_bayesian_1996}. However, existing approaches are often highly sensitive to hyperparameter choices, and hard to scale to modern datasets and architectures, which limits their general applicability in modern deep learning.

In this paper we propose a different approach to Bayesian deep learning: 
we use the information contained in the SGD trajectory
to efficiently approximate the posterior distribution over the weights of the neural network. 
We find that the Gaussian distribution fitted to the first two
moments of SGD iterates, with a modified learning rate schedule, 
captures the local geometry of the posterior surprisingly
well. Using this Gaussian distribution we are able to obtain convenient, efficient, 
accurate and well-calibrated predictions in a broad range of tasks in computer vision.
In particular, our contributions are the following:

\begin{itemize}
	
	\item In this work we propose SWAG (SWA-Gaussian), a scalable approximate Bayesian
	inference technique for deep learning. SWAG builds on
	Stochastic Weight Averaging \citep{izmailov_averaging_2018}, 
	which computes an average of SGD iterates with a high constant
	learning rate schedule, to provide improved generalization in deep learning and the interpretation of SGD as approximate Bayesian inference \citep{mandt_stochastic_2017}.
	SWAG additionally computes a low-rank plus diagonal approximation to the
	covariance of the iterates, which is used together with the SWA mean, to 
	define a Gaussian posterior approximation over neural network weights.
	
	\item SWAG is motivated by the theoretical analysis of the stationary 
	distribution of SGD iterates \citep[e.g.,][]{mandt_stochastic_2017,chen_statistical_2016},
	which suggests that the SGD trajectory contains useful information about the 
	geometry of the posterior. 
	In Appendix \ref{app:assumptions} we show that the assumptions of 
	\citet{mandt_stochastic_2017} do not hold for deep neural networks, 
	due to non-convexity and over-parameterization (with further analysis 
	in the supplementary material).
	However, we find in Section \ref{sec:geometry} that in the low-dimensional subspace spanned by SGD
    iterates the shape of the posterior distribution is approximately Gaussian
    within a basin of attraction.
	Further, SWAG is able to capture the geometry of this posterior 
	remarkably well.
	
	\item In an exhaustive empirical evaluation we show that SWAG can provide 
	well-calibrated uncertainty estimates for neural networks across many
	settings in computer vision. In particular
	SWAG achieves higher test likelihood compared to many state-of-the-art approaches, including 
	MC-Dropout \citep{gal_dropout_2016}, 
	temperature scaling \citep{guo_calibration_2017}, 
    SGLD \citep{welling2011bayesian},
	KFAC-Laplace \citep{ritter_scalable_2018} and SWA 
	\citep{izmailov_averaging_2018} on CIFAR-10, CIFAR-100 and ImageNet,
	on a range of architectures. 
	We also demonstrate the effectiveness of SWAG for out-of-domain 
	detection, and transfer learning. 
    While we primarily focus on image classification, we show that SWAG can 
    significantly improve test perplexities of LSTM networks on language
    modeling problems, and
    in Appendix \ref{sec:uci} we also compare SWAG with 
    Probabilistic Back-propagation (PBP) \citep{hernandez-lobato_probabilistic_2015}, Deterministic Variational Inference (DVI) \citep{wu2018fixing}, 
    and Deep Gaussian Processes \citep{bui2016deep} on regression
    problems. 
    
	\item We release PyTorch code at \url{https://github.com/wjmaddox/swa_gaussian}.
	
\end{itemize}

\section{Related Work}

\subsection{Bayesian Methods}

Bayesian approaches represent uncertainty by placing a distribution over model parameters, and then marginalizing these parameters to form a whole predictive distribution, in a procedure known as Bayesian model averaging.
In the late 1990s, Bayesian methods were the state-of-the-art approach to learning with neural networks, through the seminal works of \citet{neal_bayesian_1996} and \citet{mackay_bayesian_1992}.
However, modern neural networks often contain millions of parameters, the posterior over these parameters (and thus the loss surface) is highly non-convex, and mini-batch approaches are often needed to move to a space of good solutions \citep{keskar_large-batch_2017}.
For these reasons, Bayesian approaches have largely been intractable for modern neural networks. 
Here, we review several modern approaches to Bayesian deep learning.

\paragraph{Markov chain Monte Carlo (MCMC)} was at one time a gold standard for 
inference with neural networks, through the Hamiltonian Monte Carlo (HMC) work 
of \citet{neal_bayesian_1996}. However, HMC requires full gradients, which is 
computationally intractable for modern neural networks.
To extend the HMC framework, stochastic gradient HMC (SGHMC) was introduced by 
\citet{chen_stochastic_2014} and allows for stochastic gradients to be used in 
Bayesian inference, crucial for both scalability and exploring a space of 
solutions that provide good generalization. 
Alternatively, stochastic gradient Langevin dynamics (SGLD) 
\citep{welling2011bayesian} uses first order Langevin dynamics in the 
stochastic gradient setting.
Theoretically, both SGHMC and SGLD asymptotically sample from the posterior in 
the limit of infinitely small step sizes.
In practice, using finite learning rates introduces approximation errors (see e.g. \citep{mandt_stochastic_2017}),
and tuning stochastic gradient MCMC methods can be quite difficult.

\paragraph{Variational Inference:}
\citet{graves2011practical} suggested fitting a Gaussian variational posterior
approximation over the weights of neural networks. 
This technique was generalized by
\citet{kingma2013auto} which proposed the {\it reparameterization trick}
for training deep latent variable models; 
multiple variational inference methods based on the reparameterization trick 
were proposed for DNNs 
\citep[e.g.,][]{kingma2015variational, blundell_weight_2015, molchanov2017variational, louizos2017multiplicative}.
While variational methods achieve strong performance for moderately sized networks, 
they are empirically noted to be difficult to train on larger architectures such as 
deep residual networks \citep{he_deep_2016}; \citet{blier_description_2018} 
argue that the difficulty of training is explained by variational methods providing inusfficient data compression for DNNs despite being designed for data compression (minimum description length).
Recent key advances \citep{louizos2017multiplicative,wu2018fixing} in variational inference for deep learning typically focus
on smaller-scale datasets and architectures.
An alternative line of work re-interprets noisy versions of optimization algorithms: for example, noisy Adam \citep{khan_fast_2018} and noisy KFAC \citep{zhang_noisy_2017}, as approximate variational inference.

\paragraph{Dropout Variational Inference:}
\citet{gal_dropout_2016} used a spike and slab variational distribution to view dropout at test time as approximate variational Bayesian inference.
Concrete dropout \citep{gal_concrete_2017} extends this idea to optimize the dropout probabilities as well.
From a practical perspective, these approaches are quite appealing as they only 
require ensembling dropout predictions at test time, and they were 
succesfully applied to several downstream tasks \citep{kendall_what_2017, mukhoti_evaluating_2018}.

\paragraph{Laplace Approximations}
assume a Gaussian posterior, $\N(\theta^*, \mathcal{I}(\theta^*)^{-1}),$
where $\theta^*$ is a MAP estimate and $\mathcal{I}(\theta^*)^{-1}$ is the inverse of the Fisher information matrix (expected value of the Hessian evaluated at $\theta^*$).
It was notably used for Bayesian neural networks in \citet{mackay_practical_1992}, 
where a diagonal approximation to the inverse of the Hessian was utilized for computational reasons.  
More recently, \citet{kirkpatrick2017overcoming} proposed using diagonal Laplace
approximations to overcome catastrophic forgetting in deep learning. 
\citet{ritter_scalable_2018} proposed the use of either a diagonal 
or block Kronecker factored (KFAC) approximation to the Hessian matrix for Laplace approximations, and
\citet{ritter_online_2018} successfully applied the KFAC approach to online learning scenarios.

\subsection{SGD Based Approximations}\label{sec:sgd_approx} 
\citet{mandt_stochastic_2017} proposed to use the iterates of averaged SGD as 
an MCMC sampler, after analyzing the dynamics of SGD using tools from stochastic calculus. 
From a frequentist perspective, \citet{chen_statistical_2016} 
showed that under certain conditions a batch means estimator of the sample 
covariance matrix of the SGD iterates converges 
to $A = \mathcal{H}(\theta)^{-1} C(\theta) \mathcal{H}(\theta)^{-1}$, 
where $\mathcal{H}(\theta)^{-1}$ is the inverse of 
the Hessian of the log likelihood and 
$C(\theta) = \mathbb{E}(\nabla \log p(\theta) \nabla \log p(\theta)^T)$ is the 
covariance of the gradients of the log likelihood. 
\citet{chen_statistical_2016} then show that using $A$ and the sample average of 
the iterates for a Gaussian approximation produces well calibrated confidence 
intervals of the parameters and that the variance of these estimators achieves the Cramer Rao 
lower bound (the minimum possible variance).
A description of the asymptotic covariance of the SGD iterates dates 
back to \citet{ruppert_efficient_1988} and \citet{polyak_acceleration_1992},
who show asymptotic convergence of Polyak-Ruppert averaging.

\subsection{Methods for Calibration of DNNs}
\citet{lakshminarayanan_simple_2017} proposed using ensembles of several networks for enhanced calibration, and incorporated an adversarial loss function to be used when possible as well.
Outside of probabilistic neural networks, \citet{guo_calibration_2017} proposed temperature scaling, a procedure which uses a validation set and a single hyperparameter to rescale the logits of DNN outputs for enhanced calibration.
\citet{kuleshov_accurate_2018} propose calibrated regression using a similar rescaling technique.

\section{SWA-Gaussian for Bayesian Deep Learning}
In this section we propose SWA-Gaussian (SWAG) for Bayesian
model averaging and uncertainty estimation.
In Section \ref{sec:swa}, we review stochastic weight averaging (SWA) \citep{izmailov_averaging_2018}, which we view as estimating the mean of the stationary distribution of SGD iterates.
We then propose SWA-Gaussian in Sections \ref{sec:swag_diag} and \ref{sec:swag_cov}
to estimate the covariance of the stationary distribution, forming a Gaussian approximation to the posterior over weight parameters.
With SWAG, uncertainty in weight space is captured with minimal modifications to the SWA training procedure.
We then present further theoretical and empirical analysis for SWAG in Section \ref{sec:geometry}.

\subsection{Stochastic Gradient Descent (SGD)}

Standard training of deep neural networks (DNNs) proceeds by applying stochastic gradient descent on the model weights $\theta$ with the following update rule:
\begin{equation*}
\Delta \theta_{t} = -\eta_t\left( \frac{1}{B} \sum_{i=1}^B \nabla_\theta \log{p(y_i|f_\theta(x_i))} - \frac{\nabla_\theta \log{p(\theta)}}{N}\right),
\end{equation*}
where the learning rate is $\eta,$  the $i$th input (e.g.\, image) and label are $\{x_i, y_i\}$, the size of the whole training set is $N$, the size of the batch is $B$, and the DNN, $f,$ has
weight parameters $\theta$.\footnote{We ignore momentum for simplicity in this update; however we utilized momentum in the resulting experiments and it is covered theoretically \citep{mandt_stochastic_2017}.} The loss function is a negative log likelihood $- \sum_i \log{p(y_i | f_\theta(x_i))},$ combined with a regularizer $\log{p(\theta)}$.
This type of maximum likelihood training does not represent uncertainty in the predictions or parameters $\theta$.

\subsection{Stochastic Weight Averaging (SWA)}\label{sec:swa}

The main idea of SWA \citep{izmailov_averaging_2018} is to run SGD with a constant learning rate schedule 
starting from a pre-trained solution, and to average the weights of the models it traverses.
Denoting the weights of the network obtained after epoch $i$ of SWA training $\theta_i,$ the SWA solution after $T$ epochs is given by
$\theta_{\swa} = \frac 1 T \sum_{i=1}^T \theta_i \,.$
A high constant learning rate schedule ensures that SGD explores the set of 
possible solutions instead of simply converging to a single point in the weight
space. \citet{izmailov_averaging_2018} argue that conventional SGD training
converges to the boundary of the set of high-performing solutions; 
SWA on the other hand is able to find a more centered solution that is robust to the
shift between train and test distributions, leading to improved generalization
performance. 
SWA and related ideas have been successfully applied to a wide range of 
applications 
\citep[see e.g.][]{athiwaratkun_improving_2019, yang2019swalp, yazici_unusual_2019, nikishin_improving_2018}.
A related but different procedure is Polyak-Ruppert averaging \cite{polyak_acceleration_1992,ruppert_efficient_1988} in stochastic convex optimization, which uses a learning rate decaying to zero.
\citet{mandt_stochastic_2017} interpret Polyak-Ruppert averaging as a sampling procedure, with convergence occurring to the true posterior under certain strong conditions. Additionally, they explore the theoretical feasibility of SGD (and averaged SGD) as an approximate Bayesian inference scheme; we test their assumptions in Appendix \ref{app:normality}.

\subsection{SWAG-Diagonal} \label{sec:swag_diag}
\label{sec:swag_diag}

We first consider a simple diagonal format for the covariance matrix. 
In order to fit a diagonal covariance approximation, we maintain a running
average of the second uncentered moment for each weight, and then compute the 
covariance using the following standard identity at the end of training:
$\overline {\theta^2} = \frac 1 T \sum_{i=1}^T \theta_i^2$,
$\Sigma_{\diag} = \mbox{diag}(\overline{\theta^2} - \theta_{\swa}^2)$;
here the squares in $\theta_{\swa}^2$ and $\theta_i^2$ are applied elementwise.
The resulting approximate posterior distribution is then $\N(\theta_{\swa}, \Sigma_{\text{Diag}}).$ 
In our experiments, we term this method SWAG-Diagonal.

Constructing the SWAG-Diagonal posterior approximation requires storing two 
additional copies of DNN weights: $\theta_\swa$ and $\overline{\theta^2}$.
Note that these models do not have to be stored on the GPU. The additional
computational complexity of constructing SWAG-Diagonal compared to standard
training is negligible, as it only requires updating the running averages of weights once per epoch.

\subsection{SWAG: Low Rank plus Diagonal Covariance Structure}
\label{sec:swag_cov}
We now describe the full SWAG algorithm.
While the diagonal covariance approximation is standard in Bayesian deep learning 
\citep{blundell_weight_2015,kirkpatrick2017overcoming},
it can be too restrictive. 
We extend the idea of diagonal covariance approximations to utilize a more 
flexible low-rank plus diagonal posterior approximation. 
	SWAG approximates the sample covariance $\Sigma$ of the SGD iterates along
	with the mean $\theta_\swa$.\footnote{
		We note that stochastic gradient Monte Carlo methods \citep{chen_stochastic_2014,welling2011bayesian}
			also use the SGD trajectory to construct samples from the approximate
			posterior. However, these methods are principally different from SWAG in that
			they (1) require adding Gaussian noise to the gradients,
			(2) decay learning rate to zero and (3) do not construct a closed-form approximation
			to the posterior distribution, which for instance enables SWAG to draw new samples with minimal overhead. We include comparisons to SGLD \citep{welling2011bayesian} in the Appendix.	
	}

Note that the sample covariance matrix of the SGD iterates
can be written as the sum of outer products, 
$\Sigma = \frac{1}{T-1} \sum_{i=1}^T (\theta_i - \theta_\swa) (\theta_i - \theta_\swa)^{\top}$,
and is of rank $T$. As we do not have access to the value of $\theta_\swa$ during
training, we approximate the sample covariance with
$\Sigma \approx \frac{1}{T-1} \sum_{i=1}^T (\theta_i - \bar{\theta}_i) (\theta_i - \bar{\theta}_i)^\top = \frac{1}{T-1} DD^\top,$ 
where $D$ is the deviation matrix comprised of columns $D_i = (\theta_i - \bar{\theta}_i),$
and $\bar{\theta}_i$ is the running estimate of the parameters' mean obtained from
the first $i$ samples. 
To limit the rank of the estimated covariance matrix we only use the
last $K$ of $D_i$ vectors corresponding to the last $K$ epochs of training. 
Here $K$ is the rank of the resulting approximation and is a hyperparameter
of the method. 
We define $\widehat D$ to be the matrix with columns equal to $D_i$ for 
$i = T - K + 1, \ldots, T$. 

We then combine the resulting low-rank approximation 
$\Sigma_{\lr} = \frac{1}{K - 1} \cdot \widehat{D} \widehat{D}^\top$
with the diagonal approximation $\Sigma_\diag$ of Section \ref{sec:swag_diag}.
The resulting approximate posterior distribution is a Gaussian with the 
SWA mean $\theta_\swa$ and summed covariance:
$\N(\theta_\swa, \frac 1 2  \cdot (\Sigma_\diag  + \Sigma_\lr))$.\footnote{We use one half as the scale here because both the diagonal and low rank terms include the variance of the weights. We tested several other scales in Appendix \ref{app:hypers}.}
In our experiments, we term this method SWAG.
Computing this approximate posterior distribution requires storing $K$ vectors
$D_i$ of the same size as the model as well as the vectors $\theta_\swa$ and
$\overline{\theta^2}$. These models do not have to be stored on
a GPU.

To sample from SWAG we use the following identity
\begin{equation}
\label{eq:sample_cov}
\widetilde\theta = \theta_{\text{SWA}} + \frac 1 {\sqrt 2} \cdot \Sigma_\diag^{\frac 1 2} z_1 + \frac{1}{\sqrt{2(K-1)}} {\widehat D} z_2, 
\quad \mbox{where}~
z_1 \sim \mathcal{N}(0, I_d),~z_2 \sim \mathcal{N}(0, I_{K}).
\end{equation}
Here $d$ is the number of parameters in the network. Note that $\Sigma_\diag$ is
diagonal, and the product $\Sigma_\diag^{\frac 1 2} z_1$ can be computed in $\bigO(d)$ time.
The product ${\widehat D} z_2$ can be computed in $\bigO(Kd)$ time.

Related methods for estimating the covariance of SGD iterates were considered
in \citet{mandt_stochastic_2017} and \citet{chen_statistical_2016}, but store full-rank
covariance $\Sigma$ and thus scale quadratically in the number of parameters, 
which is prohibitively expensive for deep learning applications.
We additionally note that using the deviation matrix for online covariance matrix estimation comes from viewing the online updates used in \citet{bshouty_-line_2007} in matrix fashion. 

The full Bayesian model averaging procedure is given in Algorithm \ref{alg:bma}. 
	As in \citet{izmailov_averaging_2018} (SWA) we update the batch normalization statistics after sampling weights for models that use batch normalization \citep{ioffe2015batch}; we investigate the necessity of this update in Appendix \ref{app:batch_norm}.

\begin{minipage}[t!]{\textwidth}
	\vspace{-0.4cm}
	\centering
	\begin{algorithm}[H]
        \footnotesize
		\caption{Bayesian Model Averaging with SWAG}\label{alg:bma}
		
		\begin{algorithmic}
			\STATE $\theta_0$:~pretrained weights; $\eta$:~learning rate; 
			$T$:~number of steps; 
			$c$:~moment update frequency; 
			$K$:~maximum number of columns in deviation matrix; 
			$S$:~number of samples in Bayesian model averaging
			\vspace{-2mm}
			\begin{multicols}{2}
				\STATE \textbf{Train} SWAG
				\STATE ~~~~$\overline{\theta} \leftarrow \theta_0,~~\overline{\theta^2} \leftarrow \theta_0^2$\hfill\{Initialize moments\}
				\STATE ~~~~\textbf{for} $i \leftarrow 1, 2, ..., T$ \textbf{do}
				\STATE ~~~~~~~~$\theta_i \leftarrow  \theta_{i-1} - \eta \nabla_{\theta} \mathcal{L}(\theta_{i-1})$\{Perform SGD update\}
				\vspace{-2mm}
				\STATE ~~~~~~~~\textbf{if}~$\mathtt{MOD}(i,c) = 0$ \textbf{then}
				\STATE ~~~~~~~~~~~~$n \leftarrow i/c$\hfill\{Number of models\}
				\STATE ~~~~~~~~~~~~$\overline{\theta} \leftarrow \dfrac{n \overline{\theta} + \theta_{i}}{n+1},~~\overline{\theta^2} \leftarrow \dfrac{n \overline{\theta^2} + \theta_{i}^2}{n+1}$\hfill\{Moments\}
				\STATE ~~~~~~~~~~~~\textbf{if} $\mathtt{NUM\_COLS}(\widehat{D}) = K$ \textbf{then}
				\STATE ~~~~~~~~~~~~~~~~$\mathtt{REMOVE\_COL}(\widehat{D}[:, 1])$
				\STATE ~~~~~~~~~~~~$\mathtt{APPEND\_COL}(\widehat{D}, \theta_i - \overline{\theta})$\hfill\{Store deviation\}
				\STATE ~~~~\textbf{return} $\theta_{\text{SWA}} = \overline{\theta},~~ \Sigma_\diag = \overline{\theta^2} - \overline{\theta}^2,~~\widehat{D}$
				\columnbreak
				\STATE \textbf{Test } Bayesian Model Averaging
				\STATE ~~~~\textbf{for} $i \leftarrow 1, 2, ..., S$ \textbf{do}
				\STATE ~~~~~~~~~Draw $\widetilde{\theta}_i \sim \mathcal{N}\left(\theta_{\text{SWA}}, \frac{1}{2}\Sigma_\diag + \frac{\widehat{D}\widehat{D}^{\top}}{2(K - 1)}\right)$ \eqref{eq:sample_cov}
				\STATE ~~~~~~~~~Update batch norm statistics with new sample.
				\STATE ~~~~~~~~~$p(y^*|\text{Data})~+\!=~\frac{1}{S} p(y^*|\widetilde{\theta}_i)$
				\STATE ~~~~\textbf{return} $p(y^*|\text{Data})$
			\end{multicols}
		\end{algorithmic}
		\vspace{-0.4cm}	
	\end{algorithm}
\end{minipage}

\vspace{-2mm}
\subsection{Bayesian Model Averaging with SWAG}\label{app:prior_choice}
\vspace{-2mm}

Maximum a-posteriori (MAP) optimization is a procedure whereby one maximizes the (log) posterior with respect to parameters $\theta$:
$\log p(\theta | \mathcal{D}) = \log p(\mathcal{D} | \theta) + \log p(\theta)$.
Here, the prior $p(\theta)$ is viewed as a regularizer in optimization. However, MAP is \emph{not} Bayesian inference, since one only considers
a single setting of the parameters $\hat{\theta}_{\text{MAP}} = \text{argmax}_{\theta} p(\theta | \mathcal{D})$ in making predictions, forming 
$p(y_* | \hat{\theta}_{\text{MAP}}, x_*)$, where $x_*$ and $y_*$ are test inputs and outputs. 

A Bayesian procedure instead \emph{marginalizes} the posterior distribution over $\theta$, in a Bayesian model average, for the unconditional
predictive distribution:
$p(y_* | \mathcal{D}, x_*) = \int p(y_* | \theta, x_*) p(\theta | \mathcal{D}) d\theta$.
In practice, this integral is computed through a Monte Carlo sampling procedure:\\
$p(y_* | \mathcal{D}, x_*) \approx \frac{1}{T} \sum^T_{t=1} p(y_* | \theta_t, x_*) \,, \quad \theta_t \sim p(\theta | \mathcal{D})$.

We emphasize that in this paper we are approximating  
\emph{fully Bayesian inference}, rather than MAP optimization. We develop
a Gaussian approximation to the posterior from SGD iterates, $p(\theta | \mathcal{D}) \approx \mathcal{N}(\theta; \mu, \Sigma)$, and then
sample from this posterior distribution to perform a Bayesian model average. In our procedure, 
optimization with different regularizers, to characterize the Gaussian posterior approximation, 
corresponds to approximate Bayesian inference with different priors $p(\theta)$. 

\paragraph{Prior Choice}Typically, weight decay is used to regularize DNNs, corresponding to explicit L2 regularization when SGD without momentum is used to train the model. 
When SGD is used \textit{with} momentum, as is typically the case, implicit regularization still occurs, producing a vague prior on the weights of the DNN
in our procedure. 
This regularizer can be given an explicit Gaussian-like form (see Proposition 3 of \citet{loshchilov_decoupled_2019}), corresponding to a prior distribution on the weights.

Thus, SWAG is an approximate Bayesian inference algorithm in our experiments (see Section \ref{sec:exps}) and can be applied to most DNNs without any modifications of the training procedure (as long as SGD is used with weight decay or explicit L2 regularization).
Alternative regularization techniques could also be used, producing different priors on the weights.
It may also be possible to similarly utilize Adam and other stochastic first-order methods, which view as a promising direction for future work.

\begin{figure*}
	\centering
	\begin{subfigure}[u]{0.3\textwidth}
		\includegraphics[width=\textwidth]{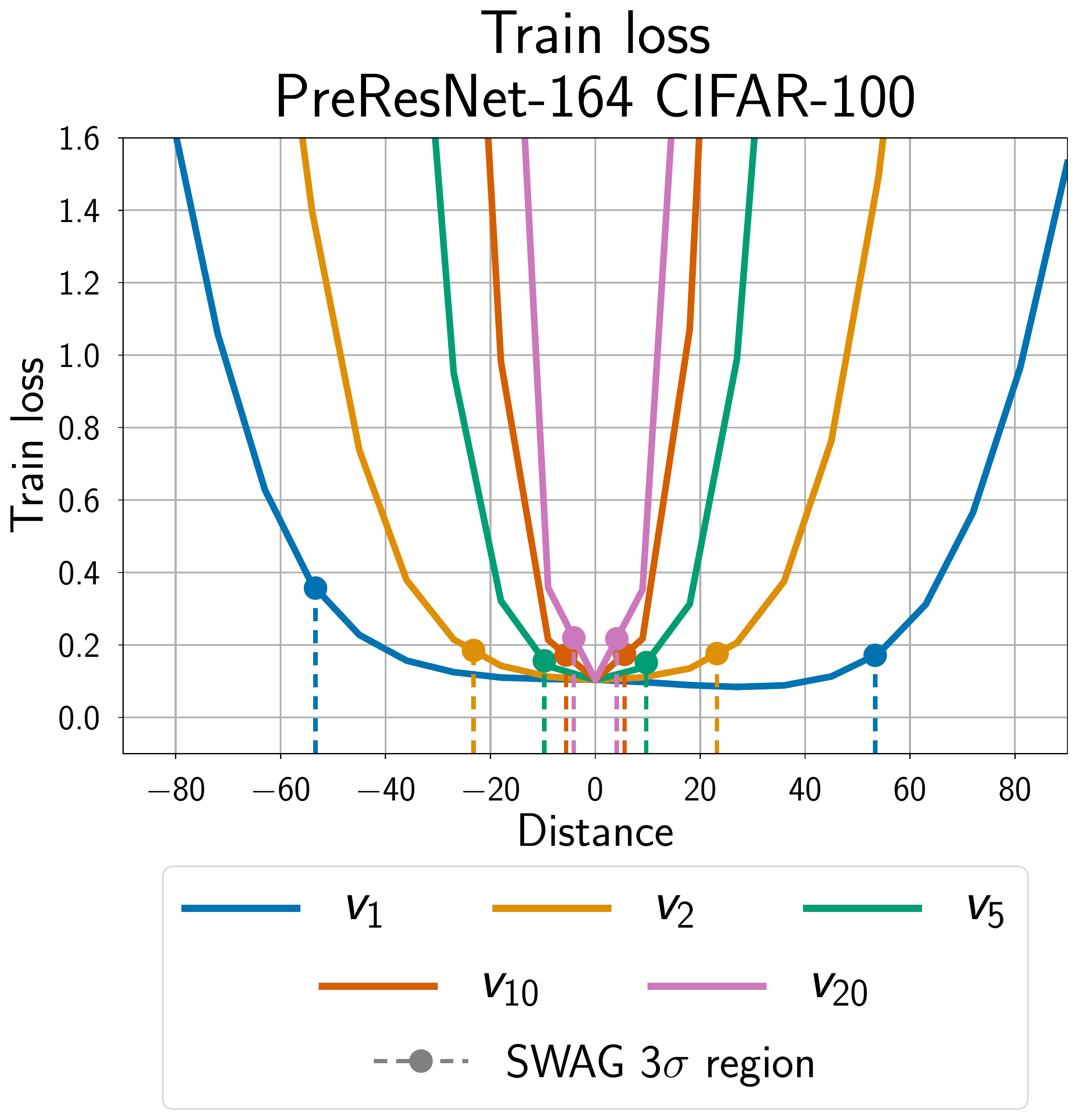}
	\end{subfigure}
	\begin{subfigure}[u]{0.3\textwidth}
		\includegraphics[width=\textwidth]{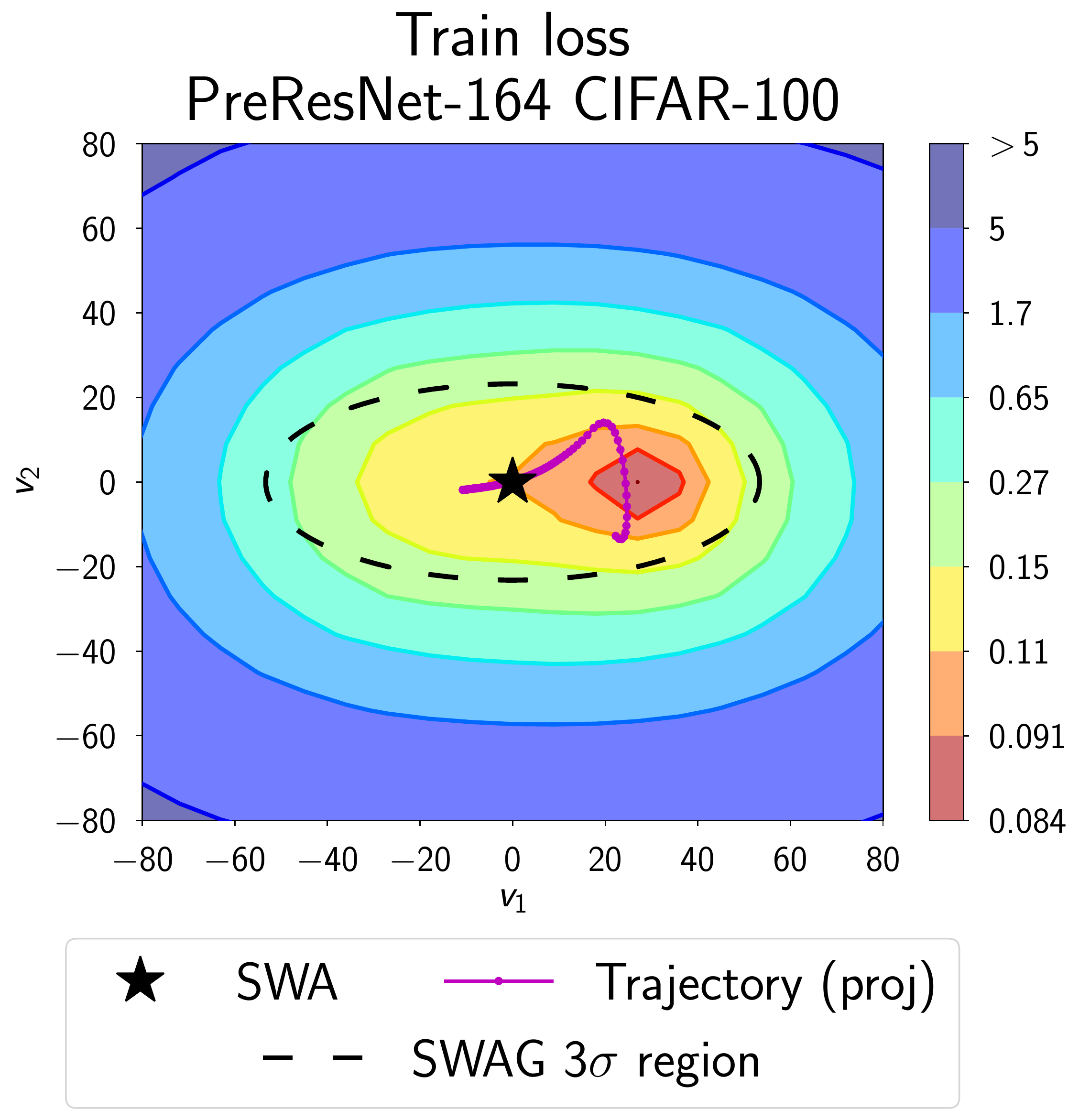}
	\end{subfigure}
	\begin{subfigure}[u]{0.3\textwidth}
		\includegraphics[width=\textwidth]{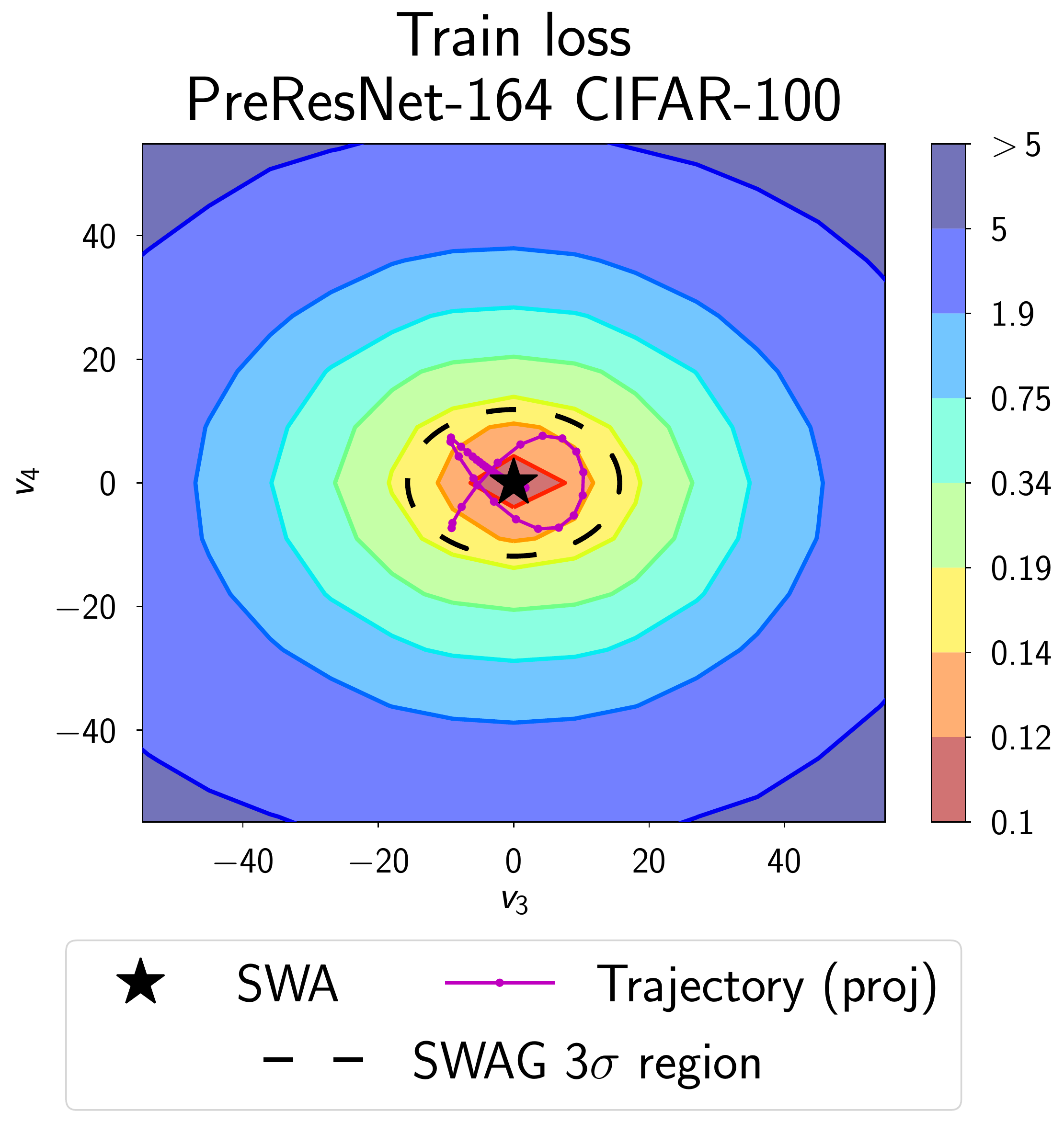}
	\end{subfigure}
	\caption{
		\textbf{Left:} Posterior joint density cross-sections along the rays corresponding to different
		eigenvectors of SWAG covariance matrix.
		\textbf{Middle:} Posterior joint density surface in the plane spanned by eigenvectors of SWAG 
		covariance matrix corresponding to the first and second largest eigenvalues and
		(\textbf{Right:}) the third and fourth largest eigenvalues. 
		All plots are produced using PreResNet-164 on CIFAR-100. The SWAG distribution projected onto these directions fits the geometry of the posterior density remarkably well.
	}
	\label{fig:geometry}
	\vspace{-.4cm}
\end{figure*}

\section{Does the SGD Trajectory Capture Loss Geometry?}\label{sec:geometry}

To analyze the quality of the SWAG approximation, we study the posterior
density along the directions corresponding to the eigenvectors of the SWAG 
covariance matrix for PreResNet-164 on CIFAR-100. 
In order to find these eigenvectors we use randomized SVD \citep{halko2011finding}.\footnote{From \texttt{sklearn.decomposition.TruncatedSVD}.} 
In the left panel of Figure \ref{fig:geometry} we visualize 
the $\ell_2$-regularized cross-entropy loss $L(\cdot)$ (equivalent to the joint density of the weights and the loss with a Gaussian prior) as a function of  
distance $t$ from the SWA solution $\theta_{\swa}$ along the $i$-th eigenvector
$v_i$ of the SWAG covariance:
$\phi(t) = L(\theta_\swa + t \cdot \frac {v_i}{\|v_i\|})$.
Figure \ref{fig:geometry} (left) shows a clear correlation between the variance of the SWAG approximation 
and the width of the posterior along the directions $v_i$. 
The SGD iterates indeed 
contain useful information about the shape of the posterior distribution, and
SWAG is able to capture this information. 
We repeated the same experiment for SWAG-Diagonal, finding that there was almost no variance in these eigen-directions.
Next, in Figure \ref{fig:geometry} (middle) we plot the posterior density surface in the 2-dimensional plane in the weight space
spanning the two top eigenvectors $v_1$ and $v_2$ of the SWAG covariance:
$\psi(t_1, t_2) = L(\theta_\swa + t_1 \cdot \frac {v_1}{\|v_1\|} + t_2 \cdot \frac {v_2}{\|v_2\|})$.
Again, 
SWAG is able to capture the geometry of the posterior. 
The contours of constant posterior density
appear remarkably well aligned with the eigenvalues of the SWAG covariance.
We also present the analogous plot for the third and fourth top eigenvectors
in Figure \ref{fig:geometry} (right).
In Appendix \ref{app:geometry}, we additionally present similar results for PreResNet-164 on CIFAR-10 and VGG-16 on CIFAR-100.

As we can see, SWAG is able to capture the geometry of the posterior in the 
subspace spanned by SGD iterates. 
However, the dimensionality of this subspace is very low compared
to the dimensionality of the weight space, and we can not guarantee that SWAG
variance estimates are adequate along all directions in weight space.
In particular, we would expect SWAG to under-estimate the variances along random
directions, as the SGD trajectory is in a low-dimensional subspace of 
the weight space, and a random vector has a close-to-zero projection on this
subspace with high probability. 
In Appendix \ref{app:normality} we visualize the trajectory 
of SGD applied to a quadratic function, and further discuss the relation between the geometry of
objective and SGD trajectory.
In Appendices \ref{app:normality} and \ref{app:assumptions}, we also 
empirically test the assumptions behind theory relating the SGD 
stationary distribution to the true posterior for neural networks.

\section{Experiments}\label{sec:exps}

We conduct a thorough empirical evaluation of SWAG, comparing to 
a range of high performing baselines, 
including MC dropout \citep{gal_dropout_2016}, temperature scaling \citep{guo_calibration_2017}, 
SGLD \citep{welling2011bayesian},
Laplace approximations \citep{ritter_scalable_2018},
deep ensembles \citep{lakshminarayanan_simple_2017}, and ensembles of SGD iterates that were used to construct the SWAG approximation.
In Section \ref{sec:exp_calibration} we evaluate SWAG
predictions and uncertainty estimates on image classification tasks. We also
evaluate SWAG for transfer learning and out-of-domain data detection.
We investigate the effect of hyperparameter choices  and practical limitations in SWAG, 
such as the effect of learning rate on the scale of uncertainty, in Appendix
\ref{app:hypers}.

\subsection{Calibration and Uncertainty Estimation on Image Classification Tasks}
\label{sec:exp_calibration}

\begin{figure*}[!t]
	\centering
	\includegraphics[width=\textwidth]{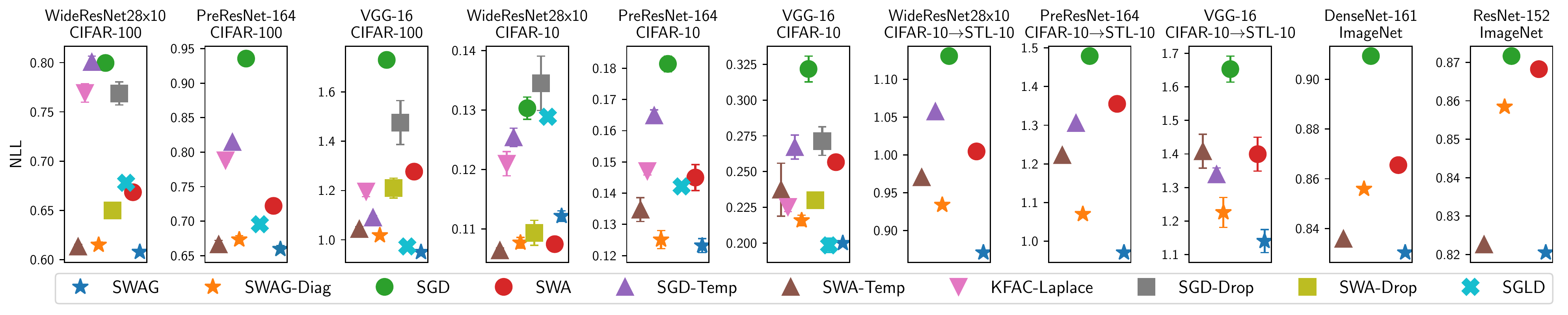}
	\caption{
		Negative log likelihoods for SWAG and baselines.
		Mean and standard deviation (shown with error-bars) over $3$ runs are 
		reported for each experiment on CIFAR datasets.
		SWAG (blue star) consistently outperforms alternatives, with lower negative log
		likelihood, with the 
		largest improvements on transfer learning. Temperature scaling applied on top of SWA (SWA-Temp) often performs close to as well on the non-transfer learning tasks, but requires a validation set.
	}
	\label{fig:results_nll}
\end{figure*}

In this section we evaluate the quality of uncertainty estimates as well as
predictive accuracy for SWAG and SWAG-Diagonal on 
CIFAR-10, CIFAR-100 and ImageNet  ILSVRC-2012 \citep{russakovsky_imagenet_2015}. 

\begin{figure*}[!t]
	\centering
	\includegraphics[width=\textwidth]{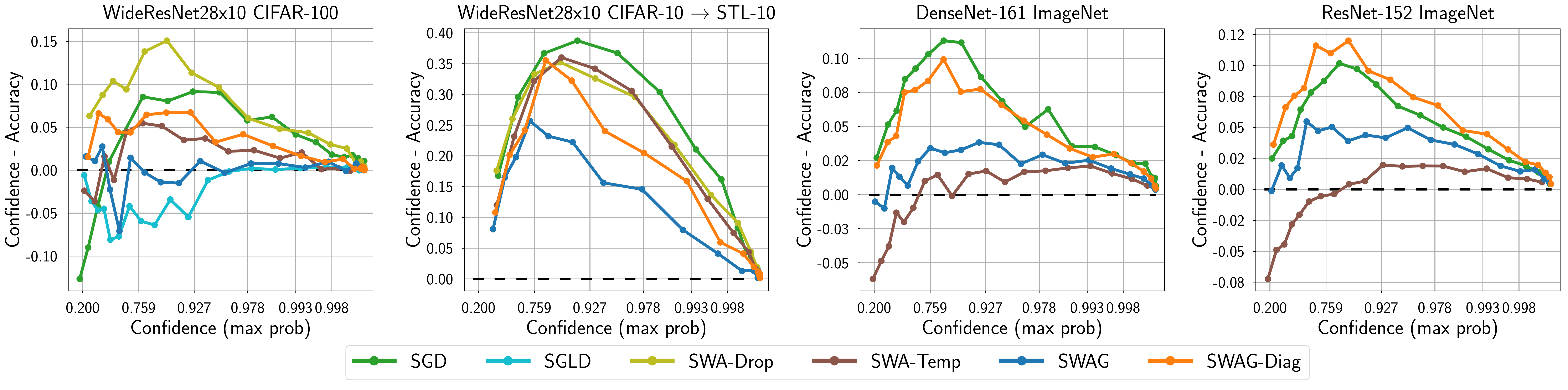}	
	\caption{
		Reliability diagrams for WideResNet28x10 on CIFAR-100 and transfer task; 
		ResNet-152 and DenseNet-161 on ImageNet. Confidence is the value of the max softmax output. A perfectly calibrated 
		network has no difference between confidence and accuracy, represented by a dashed black line. Points below this
		line correspond to under-confident predictions, whereas points above the line are overconfident predictions. 
		SWAG is able to substantially
		improve calibration over standard training (SGD), as well as SWA. Additionally, SWAG significantly outperforms temperature
		scaling for transfer learning (CIFAR-10 to STL), where the target data are not from the same distribution as the training data.
	}
	\label{fig:results_cal}
	\vspace{-.4cm}
\end{figure*}

For all methods we analyze test negative log-likelihood, which 
reflects both the accuracy and the quality of predictive uncertainty.
Following \citet{guo_calibration_2017} we also consider a variant of {\it reliability diagrams}
to evaluate the calibration of uncertainty estimates (see Figure \ref{fig:results_cal}) 
and to show the difference between a method's confidence in its predictions and its accuracy. 
To produce this plot for a given method we split 
the test data into $20$ bins uniformly based on the confidence of a method (maximum predicted probability).
We then evaluate the accuracy and mean confidence of the method on the images from each bin, and plot the difference between confidence and accuracy. 
For a well-calibrated model, this difference should be close to zero for each bin. 
We found that this procedure gives a more effective visualization of the actual confidence distribution of DNN predictions than the standard reliability diagrams used in \citet{guo_calibration_2017} and \citet{niculescu-mizil_predicting_2005}.

We provide tables containing the test accuracy, negative log likelihood and expected 
calibration error for all methods and datasets in Appendix \ref{app:tables}.
\vspace{-0.3cm}
\paragraph{CIFAR datasets}
On CIFAR datasets we run experiments with VGG-16, PreResNet-164 and WideResNet-28x10 networks.
In order to compare SWAG with existing alternatives we report the results for
standard SGD and SWA \citep{izmailov_averaging_2018} solutions (single models), 
MC-Dropout \citep{gal_dropout_2016}, 
temperature scaling \citep{guo_calibration_2017} applied to SWA and SGD solutions,
SGLD \citep{welling2011bayesian},
and K-FAC Laplace \citep{ritter_scalable_2018} methods. For all the methods we use our 
implementations in PyTorch (see Appendix \ref{app:details}). 
We train all networks for $300$ epochs, starting to collect models for SWA and SWAG 
approximations once per epoch after epoch $160$. 
For SWAG, K-FAC Laplace, and Dropout we use 30 samples at test time.
\vspace{-0.2cm}
\paragraph{ImageNet}

On ImageNet we report our results for SWAG, SWAG-Diagonal, SWA and SGD. We 
run experiments with DenseNet-161 \citep{huang_densely_2017} and Resnet-152 \citep{he_deep_2016}. 
For each model we start from a pre-trained model available in the \texttt{torchvision}
package, and run SGD with a constant learning rate for $10$ epochs. We collect
models for the SWAG versions and SWA $4$ times per epoch.
For SWAG we use $30$ samples from the posterior over network weights at test-time, and use randomly sampled
$10\%$ of the training data to update batch-normalization statistics for each of
the samples. For SGD with temperature scaling, we use the results  reported
in \citet{guo_calibration_2017}.

\paragraph{Transfer from CIFAR-10 to STL-10}
We use the models trained on
CIFAR-10 and evaluate them on STL-10 \citep{coates_analysis_2011}. 
STL-10 has a similar set of classes 
as CIFAR-10, but the image distribution is different, so adapting the model from CIFAR-10 to STL-10 is a commonly used transfer learning benchmark.
We provide further details on the architectures and hyperparameters in
Appendix \ref{app:details}.

\vspace{-0.3cm}
\paragraph{Results}
We visualize the negative log-likelihood for all methods and datasets
in Figure \ref{fig:results_nll}. 
On all considered tasks SWAG and SWAG diagonal perform comparably or better
than all the considered alternatives, SWAG being best overall. We note that
the combination of SWA and temperature scaling presents a competitive baseline.
However, unlike SWAG it requires using a validation set to tune the temperature;
further, temperature scaling is not effective when the test data distribution 
differs from train, as we observe in experiments on transfer learning from
CIFAR-10 to STL-10.

Next, we analyze the calibration of uncertainty estimates provided by different 
methods. In Figure \ref{fig:results_cal} we present reliability plots for WideResNet
on CIFAR-100, DenseNet-161 and ResNet-152 on ImageNet.
The reliability diagrams for all other datasets and architectures are presented
in the Appendix \ref{app:reliability}. As we can see, SWAG and SWAG-Diagonal both 
achieve good calibration across the board. 
The low-rank plus diagonal version of SWAG is generally better calibrated than 
SWAG-Diagonal. We also present the expected calibration error for each of the 
methods, architectures and datasets in Tables A.\ref{tab:ece_cifar},\ref{tab:ece_imagenet}.
Finally, in Tables A.\ref{tab:acc_cifar},\ref{tab:acc_imagenet} we present the predictive accuracy
for all of the methods, where SWAG is comparable with SWA and
generally outperforms the other approaches.

\subsection{Comparison to ensembling SGD solutions}\label{app:ensembling}
We evaluated ensembles of independently trained SGD solutions (Deep Ensembles, \citep{lakshminarayanan_simple_2017}) on PreResNet-164 on CIFAR-100. 
We found that an ensemble of $3$ SGD solutions has high accuracy ($82.1\%$), but only achieves NLL $0.6922$, 
which is \textit{worse than a single SWAG solution} (0.6595 NLL).
While the accuracy of this ensemble is high, SWAG solutions are much better calibrated. 
An ensemble of $5$ SGD solutions achieves NLL $0.6478$, which is \textit{competitive with a single SWAG solution, that requires $5\times$ less computation to train}. 
Moreover, we can similarly ensemble independently trained SWAG models; an ensemble of $3$ SWAG models achieves NLL of $0.6178$.

We also evaluated ensembles of SGD iterates that were used to construct the SWAG 
approximation (SGD-Ens) for all of our CIFAR models. SWAG has higher NLL than 
SGD-Ens on VGG-16, but much lower NLL on the larger PreResNet-164 and WideResNet28x10;
the results for accuracy and ECE are analogous. 

\subsection{Out-of-Domain Image Detection}
To evaluate SWAG on out-of-domain data detection we train a WideResNet as 
described in section \ref{sec:exp_calibration}
on the data from five classes of the CIFAR-10 dataset, and then analyze
predictions of SWAG variants along with the baselines on the full test set. 
We expect the outputted class probabilities on objects that belong to classes that 
were not present in the training data to have high-entropy reflecting the model's 
high uncertainty in its predictions, and considerably lower entropy on the images 
that are similar to those on which the network was trained. 
We plot the histograms of predictive entropies on the in-domain and out-of-domain 
in Figure A.\ref{fig:c5_entropy_dist} for a qualitative comparison and report 
the symmetrized KL divergence between the binned in and out of sample 
distributions in Table \ref{tab:in_out}, finding that SWAG and Dropout perform 
best on this measure.  Additional details are in Appendix 
\ref{app:oos}.

\subsection{Language Modeling with LSTMs}
We next apply SWAG to an LSTM network on language modeling tasks on Penn Treebank
and WikiText-2 datasets. In Appendix \ref{sec:lstm} we demonstrate that SWAG easily 
outperforms both SWA and NT-ASGD \citep{reglstm}, a strong baseline for LSTM 
training, in terms of test and validation perplexities.

We compare SWAG to SWA and the NT-ASGD method \citep{reglstm}, which is a strong baseline for training
LSTM models. The main difference between SWA and NT-ASGD, which is also based
on weight averaging, is that
NT-ASGD starts weight averaging much earlier than SWA: NT-ASGD switches to 
ASGD (averaged SGD) typically around epoch $100$ while with SWA we start averaging after
pre-training for $500$ epochs. We report test and validation  perplexities
for different methods and datasets in Table \ref{tab:results}.

As we can see, SWA substantially improves perplexities on both datasets
over NT-ASGD. Further, we observe that SWAG is able to substantially improve test perplexities
over the SWA solution. 
\vspace{-0.3cm}

\begin{table}[!h]
	\centering
	\caption{
		Validation and Test perplexities for NT-ASGD, SWA and SWAG on Penn Treebank
		and WikiText-2 datasets.
	}
	\label{tab:results}
	\vspace{0.2cm}
	{\small
		\begin{tabular}{lcccc}
			\toprule
			Method & PTB val & PTB test & WikiText-2 val & WikiText-2 test \\
			\midrule
			NT-ASGD &  61.2 & 58.8 & 68.7 & 65.6 \\ 
			SWA &  59.1 & 56.7 & 68.1 & 65.0 \\ 
			SWAG & \textbf{58.6} & \textbf{56.26} & \textbf{67.2} & \textbf{64.1}  \\ \bottomrule
		\end{tabular}
	}
    \vspace{-0.4cm}
\end{table}

\subsection{Regression}
Finally, while the empirical focus of our paper is classification calibration, we also 
compare to additional approximate BNN inference methods which perform well
on smaller architectures, including deterministic variational inference (DVI) \citep{wu2018fixing}, 
single-layer deep GPs (DGP) with expectation propagation \citep{bui2016deep}, 
SGLD \citep{welling2011bayesian}, and re-parameterization VI \citep{kingma2013auto} on a set of UCI regression 
tasks.
We report test log-likelihoods, RMSEs and test calibration results in 
Appendix Tables \ref{tab:small_ll} and \ref{tab:small_ca}
where it is possible to see that SWAG is competitive with these methods. Additional details are in Appendix \ref{sec:uci}.

\vspace{-2mm}
\section{Discussion}
\vspace{-2mm}

In this paper we developed SWA-Gaussian (SWAG) for approximate
Bayesian inference in deep learning. There has been a great desire to 
apply Bayesian methods in deep learning due to their theoretical properties and 
past success with small neural networks. We view SWAG as a step towards
practical, scalable, and accurate Bayesian deep learning for large modern 
neural networks.

A key geometric observation in this paper is that the posterior distribution over
neural network parameters is close to Gaussian in the subspace spanned by the trajectory of SGD. 
Our work shows Bayesian model averaging within this subspace can improve predictions over SGD or SWA solutions. 
Furthermore, \citet{gurari2019gradient} argue that the SGD trajectory lies in the subspace spanned by the eigenvectors of the Hessian corresponding to the top eigenvalues, implying that the SGD trajectory subspace corresponds to directions of rapid change in predictions. 
In recent work, \citet{izmailov2019subspace} show promising results from 
directly constructing subspaces for Bayesian inference.

\subsection*{Acknowledgements} 
WM, PI, and AGW were supported by an Amazon Research Award, Facebook Research, NSF IIS-1563887,
and NSF IIS-1910266.
WM was additionally supported by an NSF Graduate Research Fellowship under Grant No. DGE-1650441.
DV was supported by the Russian Science Foundation grant no.19-71-30020.
We would like to thank Jacob Gardner, Polina Kirichenko, and David Widmann for helpful discussions.

\bibliographystyle{apalike}
\bibliography{refs}
\appendix

\section{Asymptotic Normality of SGD}\label{app:normality}

Under conditions of decaying learning rates, smoothness of gradients, and the existence of a full rank stationary distribution, 
martingale based analyses of stochastic gradient descent 
\citep[e.g.,][Chapter 8]{asmussen_stochastic_2007} show that SGD has a Gaussian limiting distribution. 
That is, in the limit as the time step goes to infinity, 
$t^{1/2}(\theta_t - \theta^*) \rightarrow \N(0, \mathcal{H}(\theta)^{-1}\mathbb{E}(\nabla \log p(\theta) \nabla \log p(\theta)^T) \mathcal{H}(\theta)^{-1})),$
where $\mathcal{H}(\theta)^{-1}$ is the inverse of the Hessian matrix of the 
log-likelihood and $\mathbb{E}(\nabla \log p(\theta) \nabla \log p(\theta)^T) $ 
is the covariance of the gradients and $\theta^*$ is a stationary point or minima.
Note that these analyses date back to \citet{ruppert_efficient_1988} and \citet{polyak_acceleration_1992} for Polyak-Ruppert averaging, and are still popular in the analysis of stochastic gradient descent.

\citet{mandt_stochastic_2017}, \citet{chen_statistical_2016}, and \citet{babichev2018constant} all use the same style of analyses, but for different purposes. We will test the specific assumptions of \citet{mandt_stochastic_2017} in the next section.
Finally, note that the technical conditions are essentially the same conditions as for the Bernstein von Mises Theorem \citep[e.g.,][Chapter 10]{vaart_asymptotic_1998}
which implies that the asymptotic posterior will also be Gaussian.

\begin{figure}[!h]
	\centering
	\begin{subfigure}{0.3\linewidth}
		\centering
		\includegraphics[width=\textwidth]{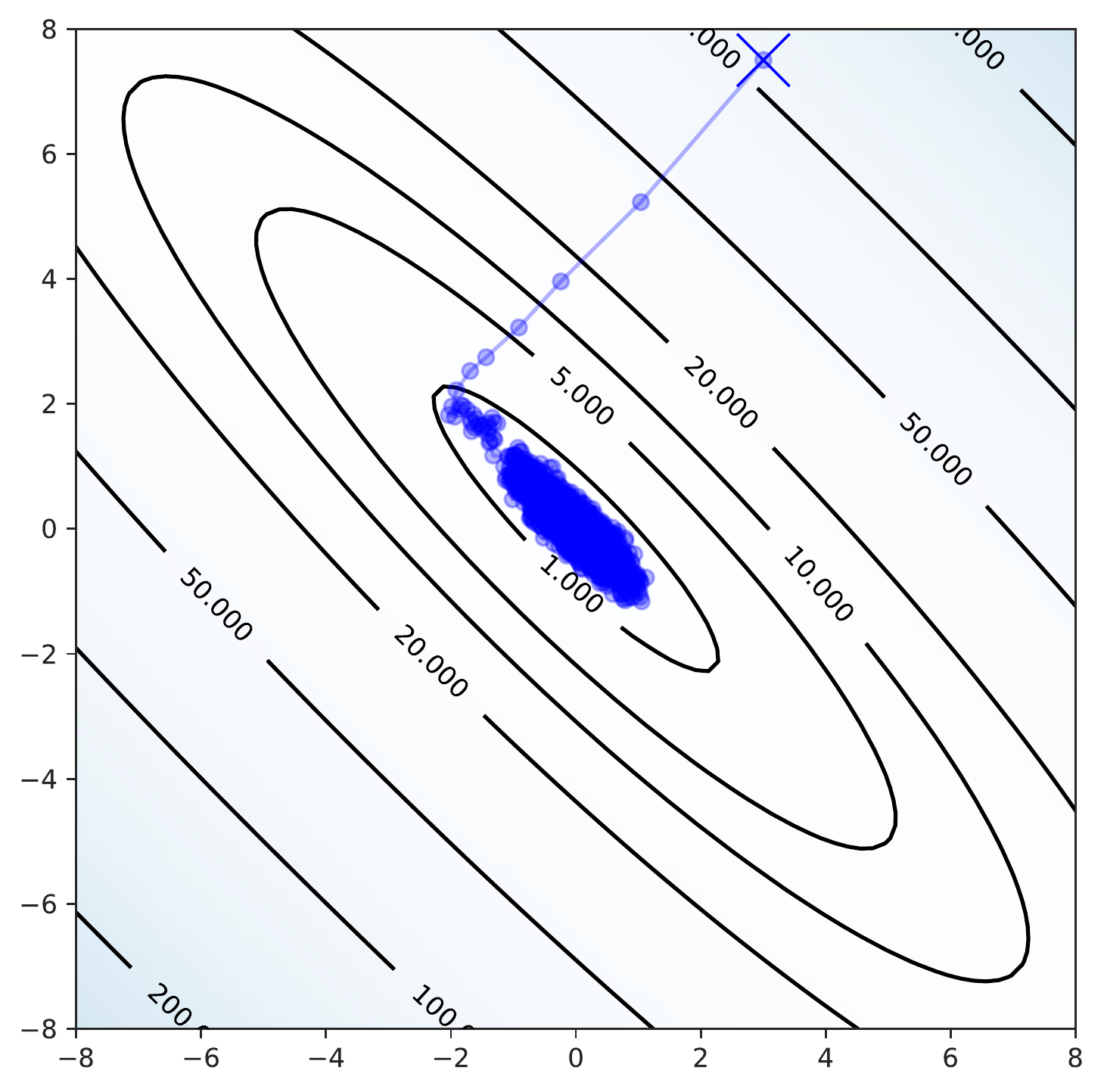}
	\end{subfigure} 
    \quad
    \quad
    \quad
	\begin{subfigure}{0.3\linewidth}
		\centering
		\includegraphics[width=\textwidth]{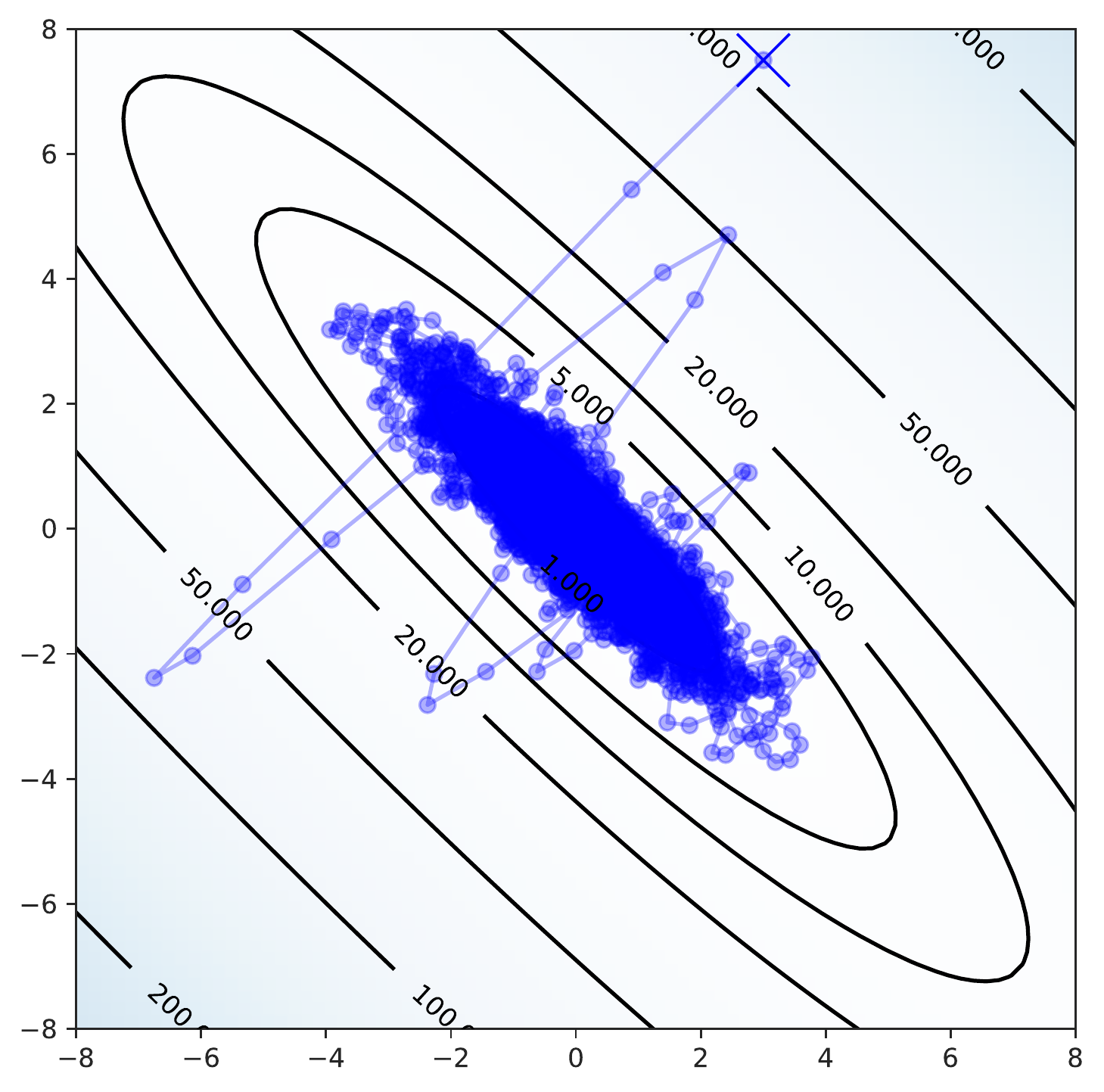}
	\end{subfigure}
	\caption{Trajectory of SGD with isotropic Gaussian gradient  noise on a quadratic
    loss function. \textbf{Left}: SGD without momentum; \textbf{Right}: SGD with 
    momentum.}
	\label{fig:quadratic}
\end{figure}

It may be counter-intuitive that, as we show in Section \ref{sec:geometry},
SWAG captures the geometry of the objective correctly. 
One might even expect SWAG estimates of variance to be inverted, as gradient 
descent would oscillate more in the sharp directions of the objective.
To gain more intuition about SGD dynamics we visualize SGD trajectory on 
a quadratic problem.
More precisely, we define a $2$-dimensional quadratic function 
$f(x, y) = (x + y)^2 + 0.05 \cdot (x - y)^2$ 
shown in 
Figure \ref{fig:quadratic}.
We then run SGD to minimize this function.

It turns out that the gradient noise plays a crucial role in forming the SGD
stationary distribution. 
If there is no noise in the gradients, we are in the full gradient descent
regime, and optimization either converges to the optimizer, or diverges 
to infinity depending on the learning rate. 
However, when we add isotropic Gaussian noise to the gradients, SGD 
converges to the correct Gaussian distribution, as we visualize in 
the left panel of Figure \ref{fig:quadratic}. 
Furthermore, adding momentum affects the scale of the distribution, but not its
shape, as we show in the right panel of Figure \ref{fig:quadratic}.
These conclusions hold as long as the learning rate in SGD is not too large.

The results we show in Figure \ref{fig:quadratic} are directly predicted
by theory in \citet{mandt_stochastic_2017}.
In general, if the gradient noise is not isotropic, the stationary distribution
of SGD would be different from the exact posterior distribution. 
\citet{mandt_stochastic_2017} provide a thorough empirical study of the 
SGD trajectory for convex problems, such as linear and logistic regression,
and show that SGD can often provide a competitive baseline on these problems.

\subsection{Other Considerations for Gaussian Approximation}

Given the covariance matrix 
$A = \mathcal{H}(\theta)^{-1}\mathbb{E}(\nabla \log p(\theta) \nabla \log p(\theta)^T) \mathcal{H}(\theta)^{-1},$ 
\citet{chen_statistical_2016} show that a batch means estimator of the iterates 
(similar to what SWAG uses) themselves will converge to $A$ in the limit of infinite time.
We tried batch means based estimators but saw no improvement; however, it could 
be interesting to explore further in future work.

Intriguingly, the covariance $A$ is the same form as sandwich estimators 
\citep[see e.g.][for a Bayesian analysis in the model mis-specification setting]{muller2013risk}, 
and so $A = \mathcal{H}(\theta)^{-1}$ under model well-specification \citep{muller2013risk,chen_statistical_2016}.
We then tie the covariance matrix of the iterates back to the well known Laplace approximation, which uses $\mathcal{H}(\theta)^{-1}$ as its covariance as described by \citet[][Chapter 28]{mackay_information_2003}, thereby justifying SWAG theoretically as a sample based Laplace approximation.

Finally, in Chapter 4 of \citet{berger2013statistical} constructs an example 
(Example 10) of fitting a Gaussian approximation from a MCMC chain, arguing that 
it empirically performs well in Bayesian decision theoretic contexts. 
\citet{berger2013statistical} give the explanation for this as the Bernstein von 
Mises Theorem providing that in the limit the posterior will itself converge to a Gaussian.
However, we would expect that even in the infinite data limit the posterior of DNNs 
would converge to something very non-Gaussian, with connected modes surrounded by gorges 
of zero posterior density \citep{garipov_loss_2018}.
One could use this justification for fitting a Gaussian from the iterates 
of SGLD or SGHMC instead.

\section{Do the assumptions of \citet{mandt_stochastic_2017} hold for DNNs?}
\label{app:assumptions}
In this section, we investigate the results of \citet{mandt_stochastic_2017} in the context of 
deep learning.
\citet{mandt_stochastic_2017} uses the following assumptions:
\begin{enumerate}
	\item Gradient noise at each point $\theta$ is $\N(0, C)$.
	\item $C$ is independent of $\theta$ and full rank.
	\item The learning rates, $\eta,$ are small enough that we can approximate the 
	dynamics of SGD by a continuous-time dynamic described by the
	corresponding stochastic differential equation.
	\item In the stationary distribution, the loss is approximately quadratic 
	near the optima, i.e. approximately $(\theta - \theta^*)^{\top} \mathcal{H}(\theta)(\theta - \theta^*),$ 
	where $\mathcal{H}(\theta^*)$ is the Hessian at the optimum; further,
	the Hessian is assumed to be positive definite.
\end{enumerate}
Assumption 1 is motivated by the central limit theorem, and Assumption 3 is necessary for the analysis in \citet{mandt_stochastic_2017}. 
Assumptions 2 and 4 may or may not hold for deep neural networks (as well as other models).
Under these assumptions, Theorem 1 of \citet{mandt_stochastic_2017} derives the optimal constant learning rate that minimizes the KL-divergence between the SGD stationary distribution and the posterior\footnote{An optimal diagonal preconditioner is also derived; our empirical work applies to that setting as well. A similar analysis with momentum holds as well, adding in only the momentum coefficient.}:
\begin{equation}
\eta^* = 2 \frac{B}{N}\frac{d}{tr(C)}, 
\label{eq:optlr}
\end{equation}
where $N$ is the size of the dataset, $d$ is the dimension of the model, $B$ is 
the minibatch size and $C$ is the gradient noise covariance.

We computed Equation \ref{eq:optlr} over the course of training for two neural networks in Figure A.\ref{fig:optlr}, finding that the predicted optimal learning rate was an order of magnitude 
larger than what would be used in practice to explore the loss surface in a reasonable time (about $4$ compared to $0.1$). 

We now focus on seeing how Assumptions 2 and 4 fail for DNNs; this will give further insight into what portions of the theory do hold, and may give insights into a corrected version of the optimal learning rate.

\subsection{Assumption 2: Gradient Covariance Noise.}

\begin{figure}
	\centering
	\begin{subfigure}{0.48\linewidth}
		\centering
		\includegraphics[width=\textwidth]{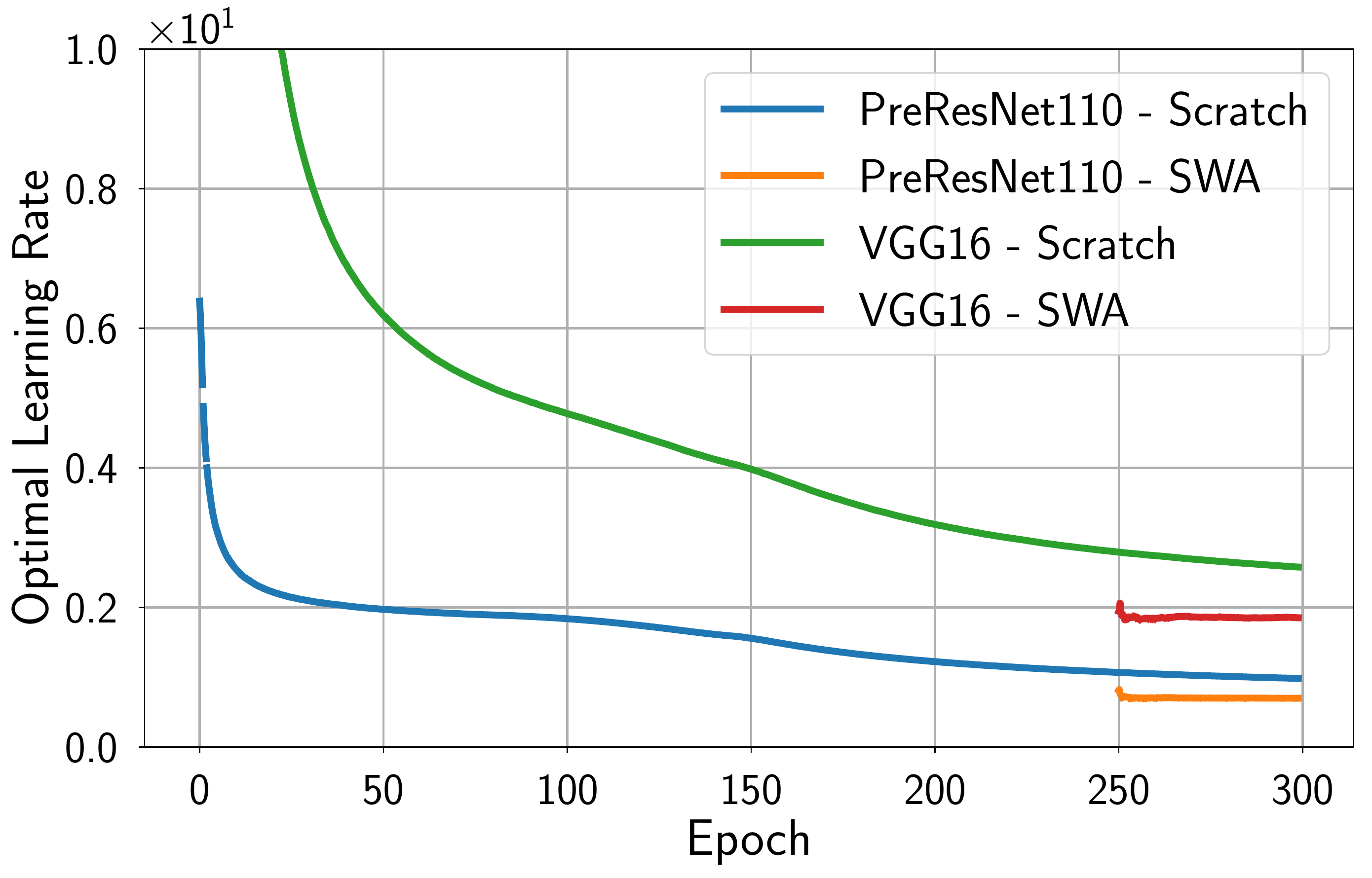}
		\caption{Optimal learning rate.}
		\label{fig:optlr}
	\end{subfigure} 
	\vspace{-0.3cm}
	\begin{subfigure}{0.48\linewidth}
		\centering
		
		\includegraphics[width=\textwidth]{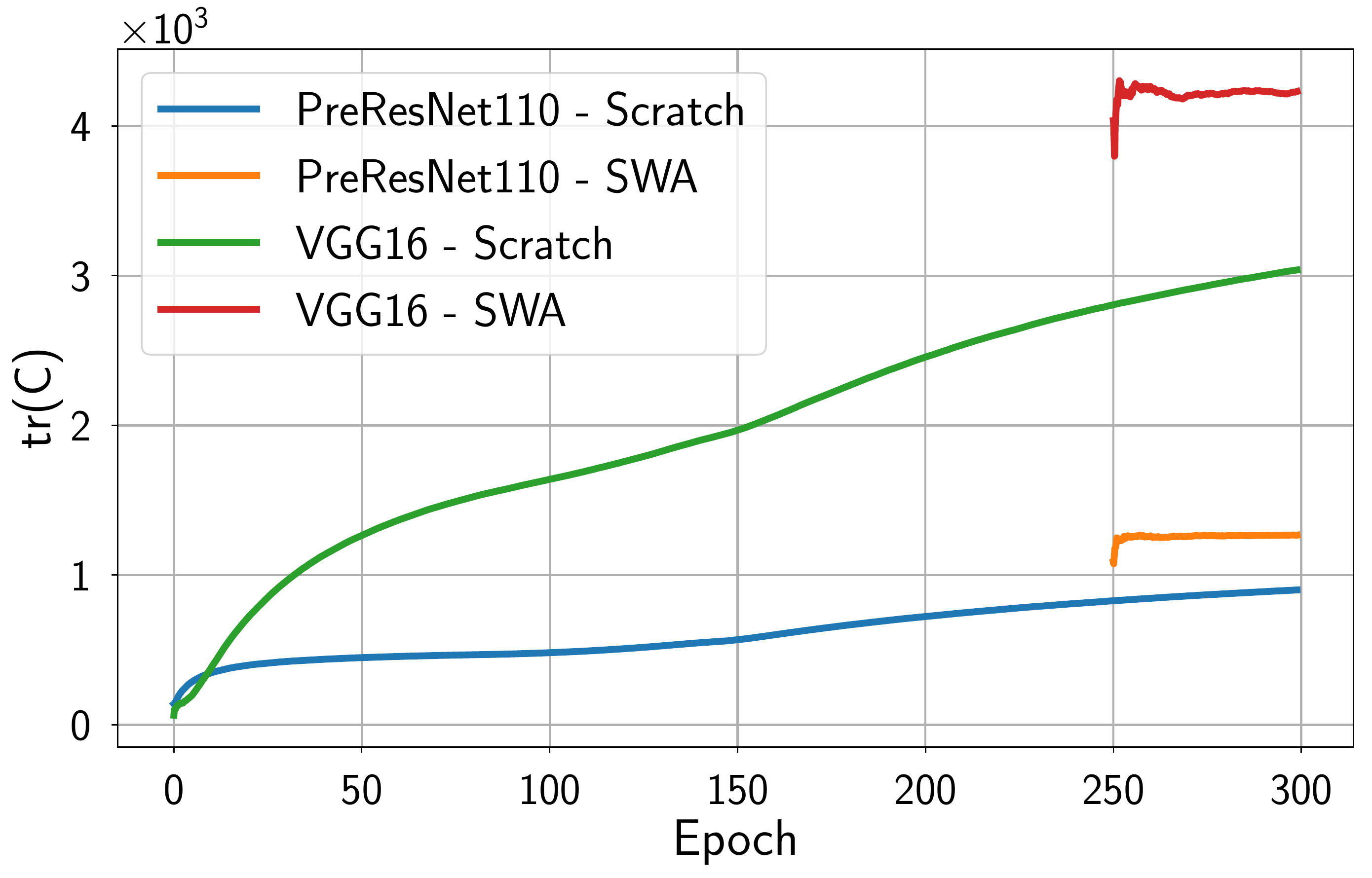}
		
		\caption{$tr(C)$}
		\label{fig:grad_norm}
	\end{subfigure}
	\caption{Gradient variance norm and computed optimal learning rates for VGG-16 and PreResNet-164. The computed optimal learning rates are always too large by a factor of $10,$ while the gradient variance stabilizes over the course of training. }
\end{figure}

In Figure A.\ref{fig:grad_norm}, the trace of the gradient noise covariance and thus the optimal learning rates \textit{are} nearly constant; however, the total variance is much too small to induce effective learning rates, probably due to over-parameterization effects inducing non full rank gradient covariances as was found in \citet{chaudhari_stochastic_2018}.
We note that this experiment is not sufficient to be fully confident that $C$ is independent of the parameterization near the local optima, but rather that $tr(C)$ is close to constant; further experiments in this vein are necessary to test if the diagonals of $C$ are constant.
The result that $tr(C)$ is close to constant suggests that a constant learning rate could be used for sampling in a stationary phase of training.
The dimensionality parameter in Equation \ref{eq:optlr} could be modified to use the number of effective parameters or the rank of the gradient noise to reduce the optimal learning rate to a feasible number.

To estimate $tr(C)$ from the gradient noise we need to divide the estimated variance by the batch size (as $V(\hat{g}(\theta)) = B C(\theta)$),
for a correct version of Equation \ref{eq:optlr}.
From Assumption 1 and Equation 6 of \citet{mandt_stochastic_2017}, we see that 
\begin{equation*}
\hat{g}(\theta) \approx g(\theta) + \frac{1}{\sqrt{B}} \nabla g(\theta), \nabla g(\theta) \sim N(0, C(\theta)),
\end{equation*}
where $B$ is the batch size.
Thus, collecting the variance of $\hat{g}(\theta)$ (the variance of the stochastic gradients) will give estimates that are upscaled by a factor of $B$, leading to a cancellation of the batch size terms:
\begin{equation*}
\eta \approx \frac{2}{N} \frac{d}{tr(V(\hat{g}(\theta)))}.
\end{equation*}
To include momentum, we can repeat the analysis in Sections 4.1 and 4.3 of \citet{mandt_stochastic_2017} finding that this also involves scaling the optimal learning rate but by a factor of $\mu,$ the momentum term.\footnote{Our experiments used $\mu = 0.1$ corresponding to $\rho = 0.9$ in PyTorch's SGD implementation.}
This gives the final optimal learning rate equation as 
\begin{equation}
\eta \approx \frac{2 \mu }{N} \frac{d}{tr(V(\hat{g}(\theta)))}.
\label{eq:corr_optlr}
\end{equation}
In Figure \ref{fig:grad_norm}, we computed $tr(C)$ for VGG-16 and PreResNet-164 on CIFAR-100 beginning from the start of training (referred to as from scratch), as well as the start of the SWAG procedure (referred to in the legend as SWA).
We see that $tr(C)$ is never quite constant when trained from scratch, while for a period of constant learning rate near the end of training, referred to as the stationary phase, $tr(C)$ is essentially constant throughout.
This discrepancy is likely due to large gradients at the very beginning of training, indicating that the stationary distribution has not been reached yet.

Next, in Figure \ref{fig:optlr}, we used the computed $tr(C)$ estimate for all four models and Equation \ref{eq:corr_optlr} to compute the optimal learning rate under the assumptions of \citet{mandt_stochastic_2017}, finding that these learning rates are not constant for the estimates beginning at the start of training and that they are too large (1-3 at the minimum compared to a standard learning rate of 0.1 or 0.01).

\subsection{Assumption 4: Hessian Eigenvalues at the Optima}

To test assumption 4, we used a GPU-enabled Lanczos method from GPyTorch \citep{gardner2018gpytorch} and used restarting to compute the minimum eigenvalue of the train loss of a pre-trained PreResNet-164 on CIFAR-100.
We found that even at the end of training, the minimum eigenvalue was $-272$ 
(the maximum eigenvalue was $3580$ for comparison), indicating that the Hessian 
is not positive definite.
This result harmonizes with other work analyzing the spectra of the Hessian for DNN training \citep{li_visualizing_2018,sagun_empirical_2018}.
Further, \citet{garipov_loss_2018} and \citet{draxler_essentially_2018} 
argue that the loss surfaces of DNNs have directions along which the 
loss is completely flat, suggesting that the loss is nowhere near a
positive-definite quadratic form.

\section{Further Geometric Experiments}\label{app:geometry}

\begin{figure*}
	\centering
	\begin{subfigure}[u]{0.30\textwidth}
		\centering
		\includegraphics[width=\textwidth]{plots/c100_resnet110_swag_width_big_font.pdf}
	\end{subfigure}
	\quad
	\begin{subfigure}[u]{0.30\textwidth}
		\centering
		\includegraphics[width=\textwidth]{plots/c100_resnet110_swag_2d_01_big_font.pdf}
	\end{subfigure}
	\quad
	\begin{subfigure}[u]{0.30\textwidth}
		\centering
		\includegraphics[width=\textwidth]{plots/c100_resnet110_swag_2d_23_big_font.pdf}
	\end{subfigure}
	\begin{subfigure}[u]{0.30\textwidth}
		\centering
		\includegraphics[width=\textwidth]{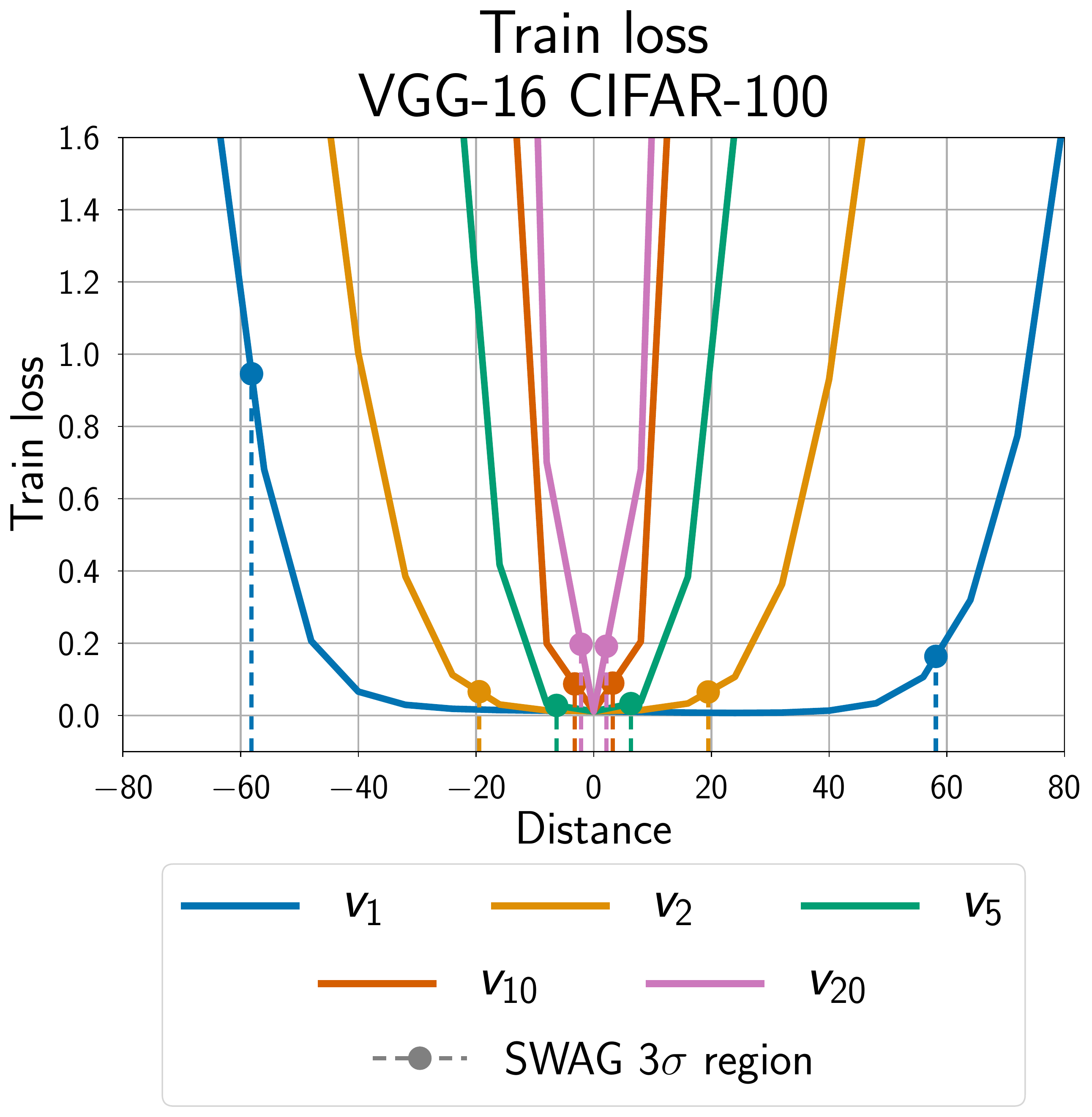}
	\end{subfigure}
	\quad
	\begin{subfigure}[u]{0.30\textwidth}
		\centering
		\includegraphics[width=\textwidth]{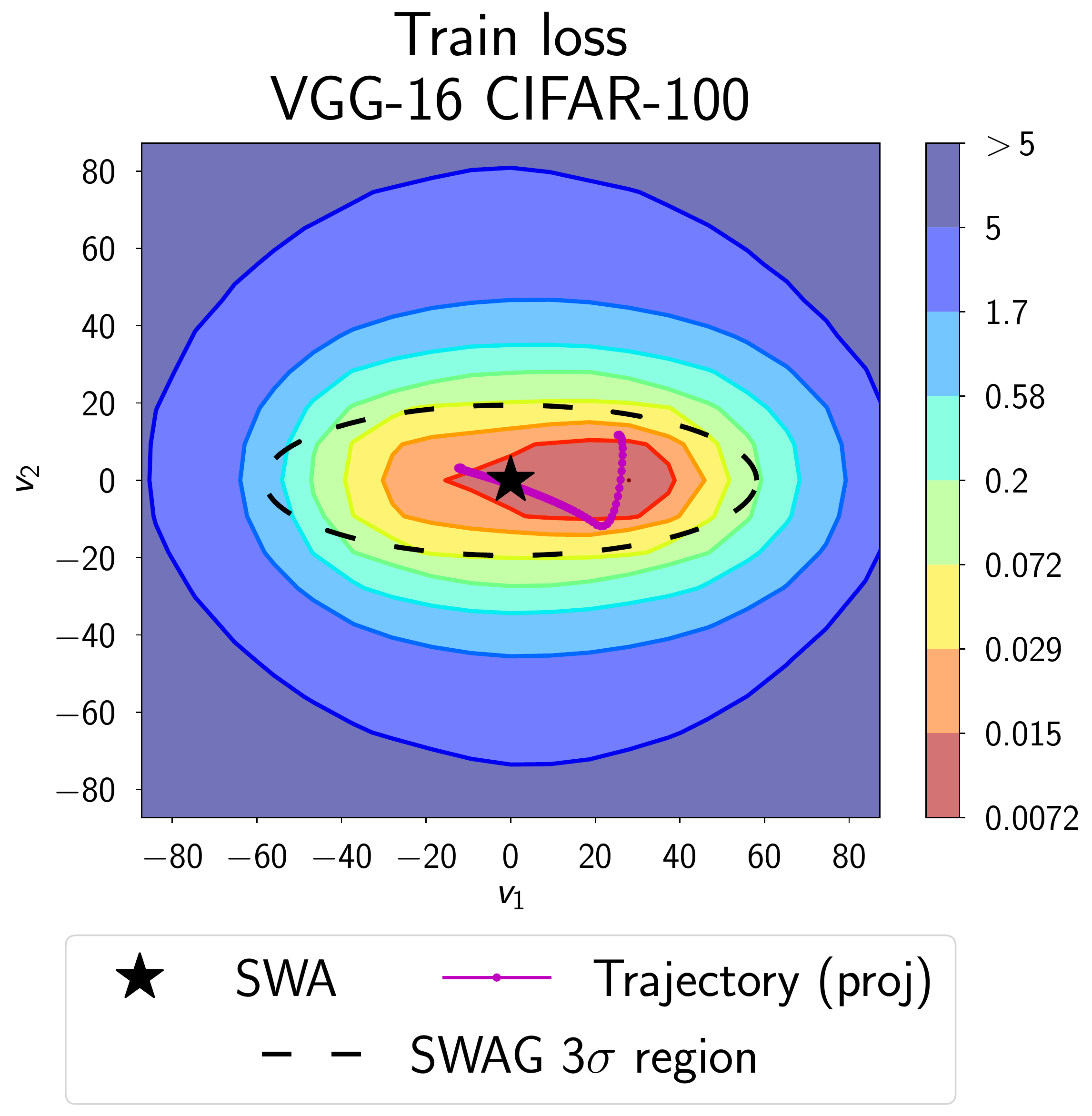}
	\end{subfigure}
	\quad
	\begin{subfigure}[u]{0.30\textwidth}
		\centering
		\includegraphics[width=\textwidth]{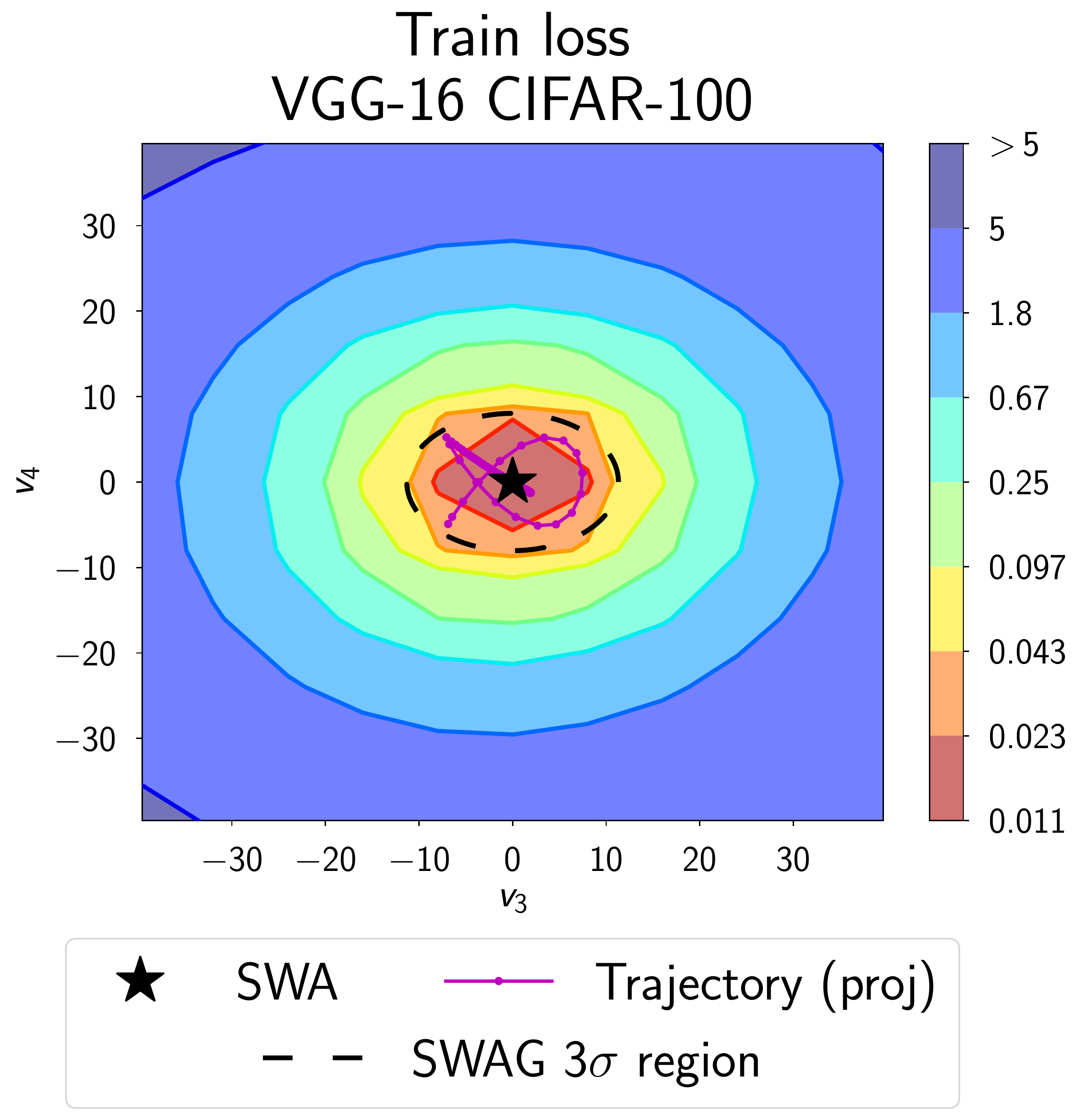}
	\end{subfigure}
	\begin{subfigure}[u]{0.30\textwidth}
		\centering
		\includegraphics[width=\textwidth]{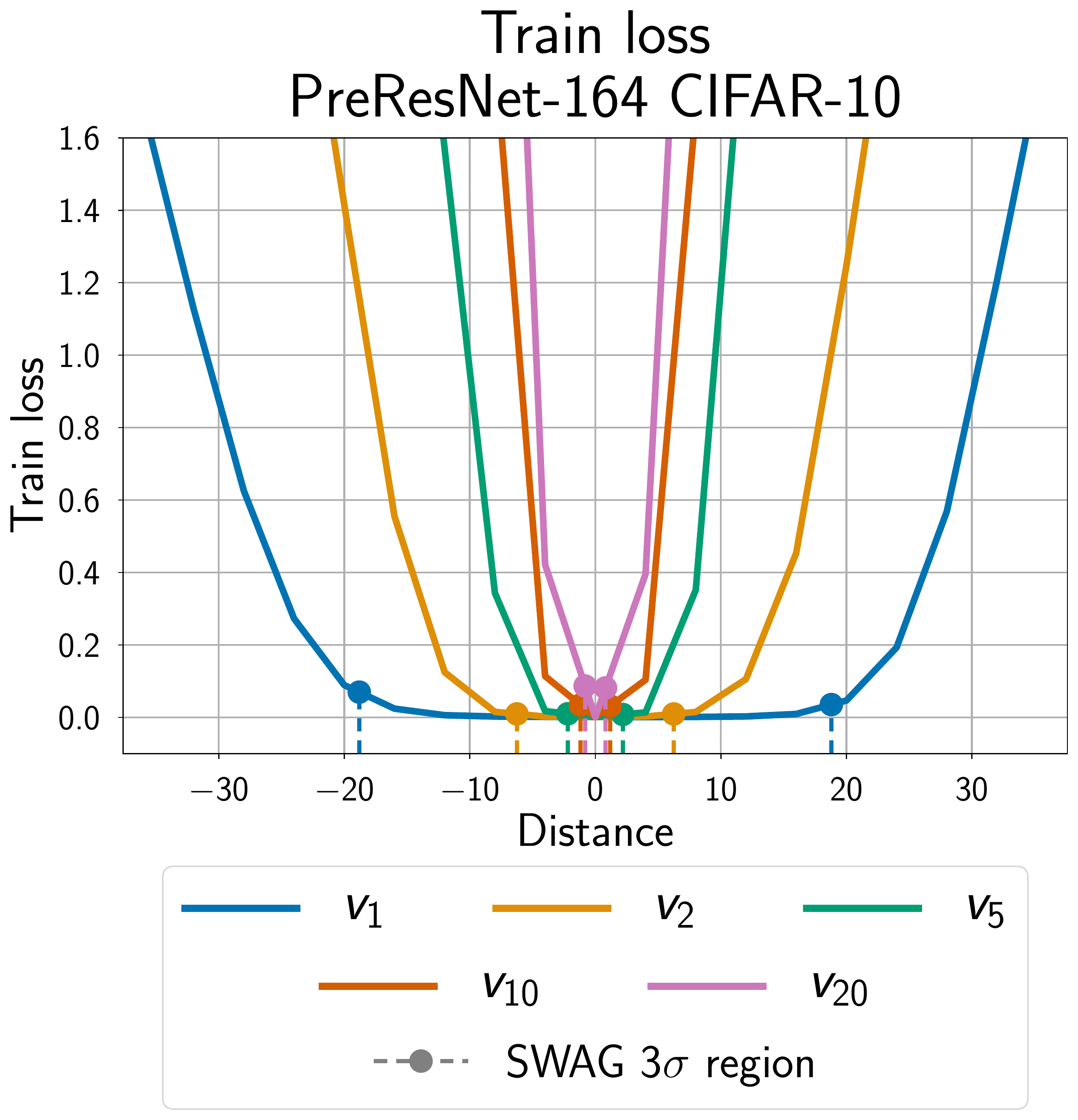}
	\end{subfigure}
	\quad
	\begin{subfigure}[u]{0.30\textwidth}
		\centering
		\includegraphics[width=\textwidth]{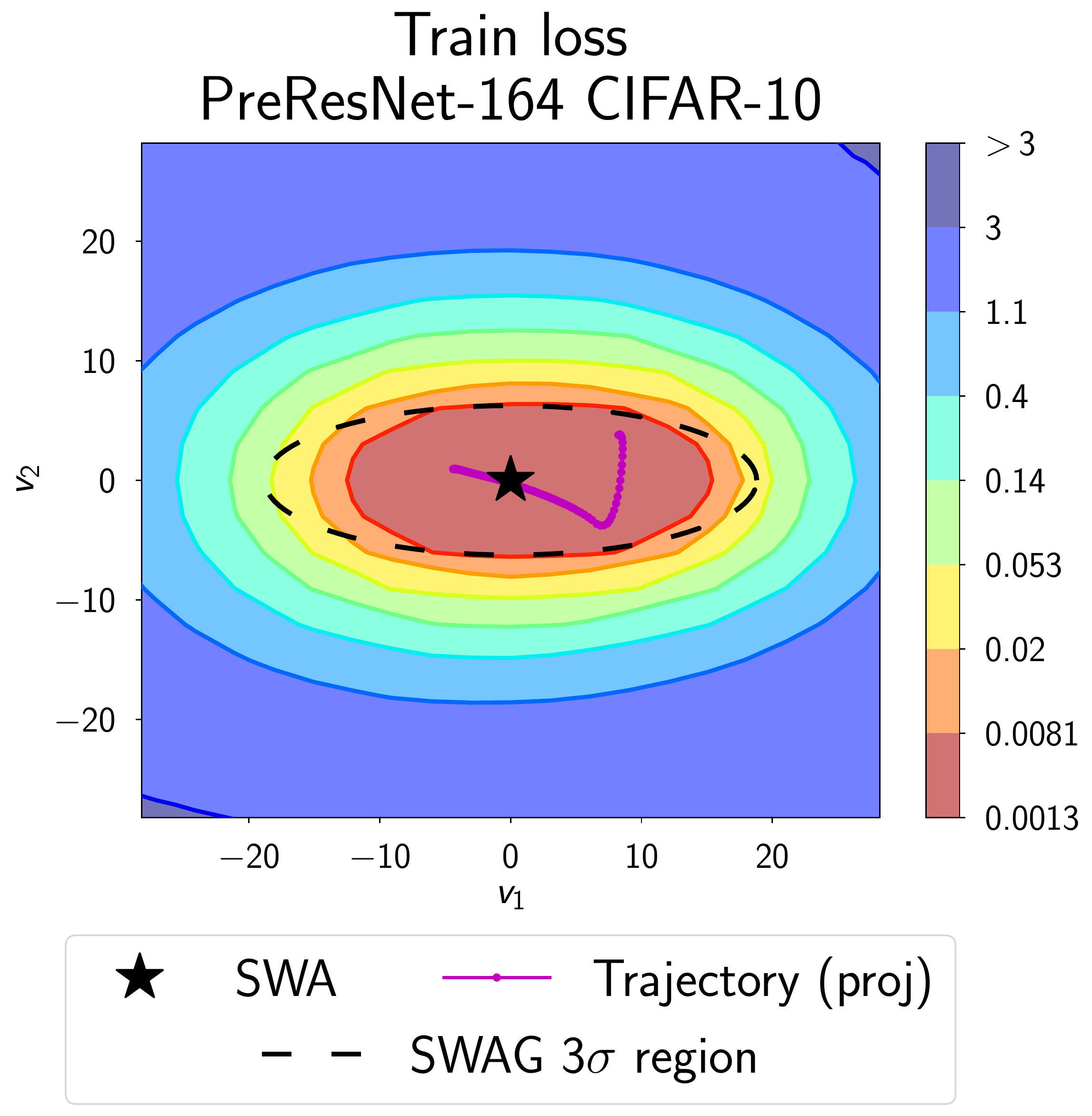}
	\end{subfigure}
	\quad
	\begin{subfigure}[u]{0.30\textwidth}
		\centering
		\includegraphics[width=\textwidth]{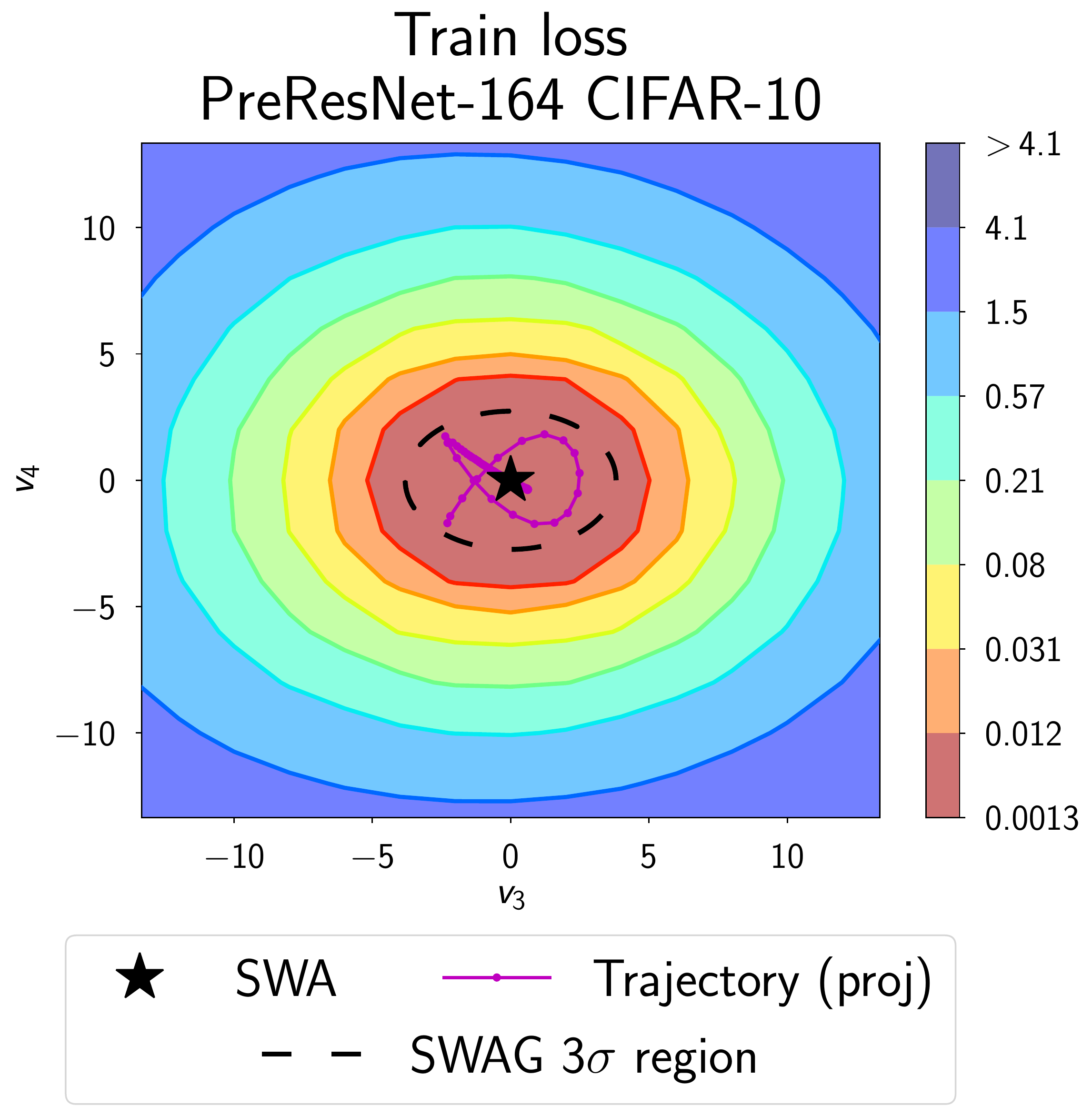}
	\end{subfigure}
	\begin{subfigure}[u]{0.30\textwidth}
		\centering
		\includegraphics[width=\textwidth]{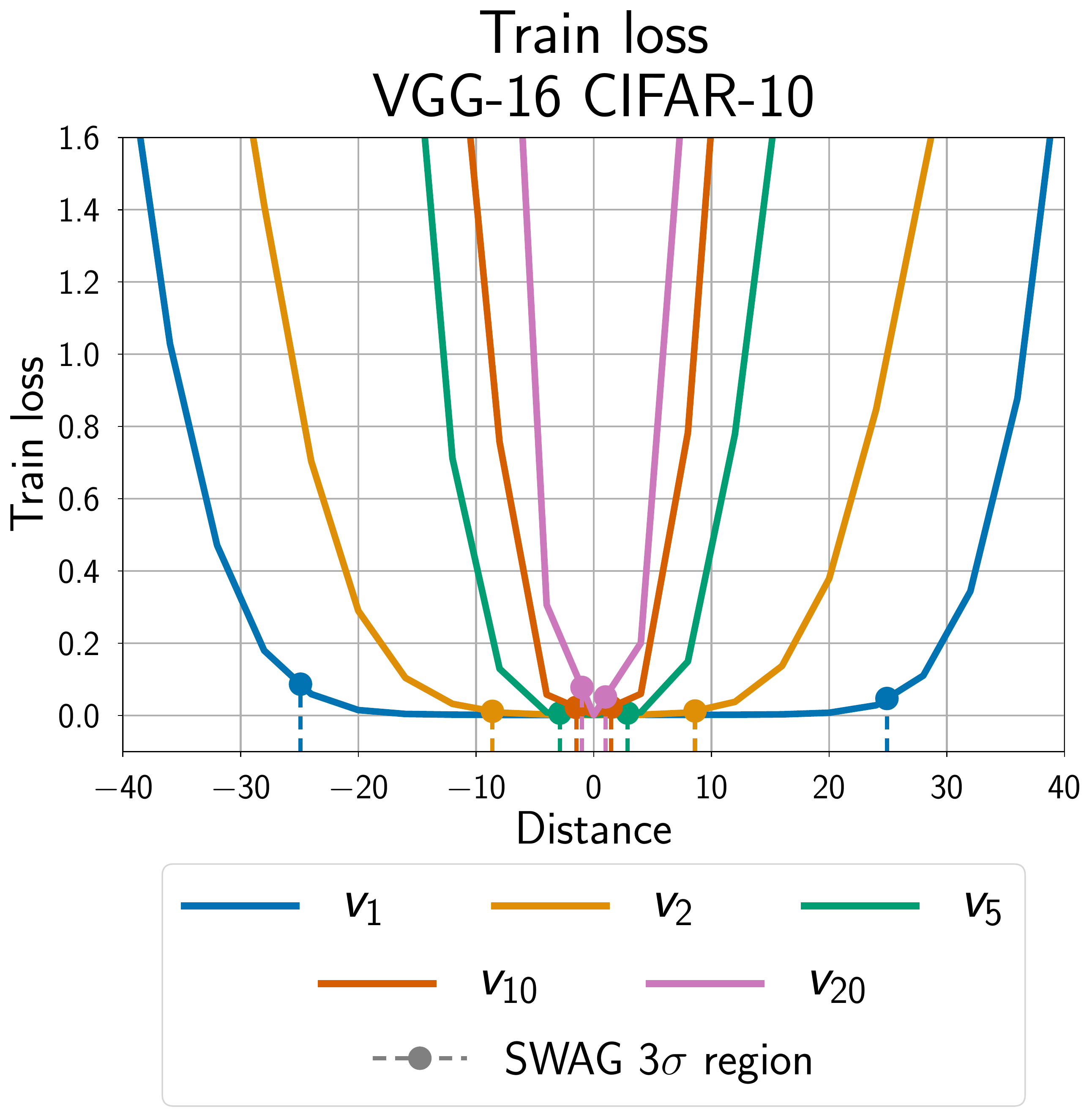}
	\end{subfigure}
	\quad
	\begin{subfigure}[u]{0.30\textwidth}
		\centering
		\includegraphics[width=\textwidth]{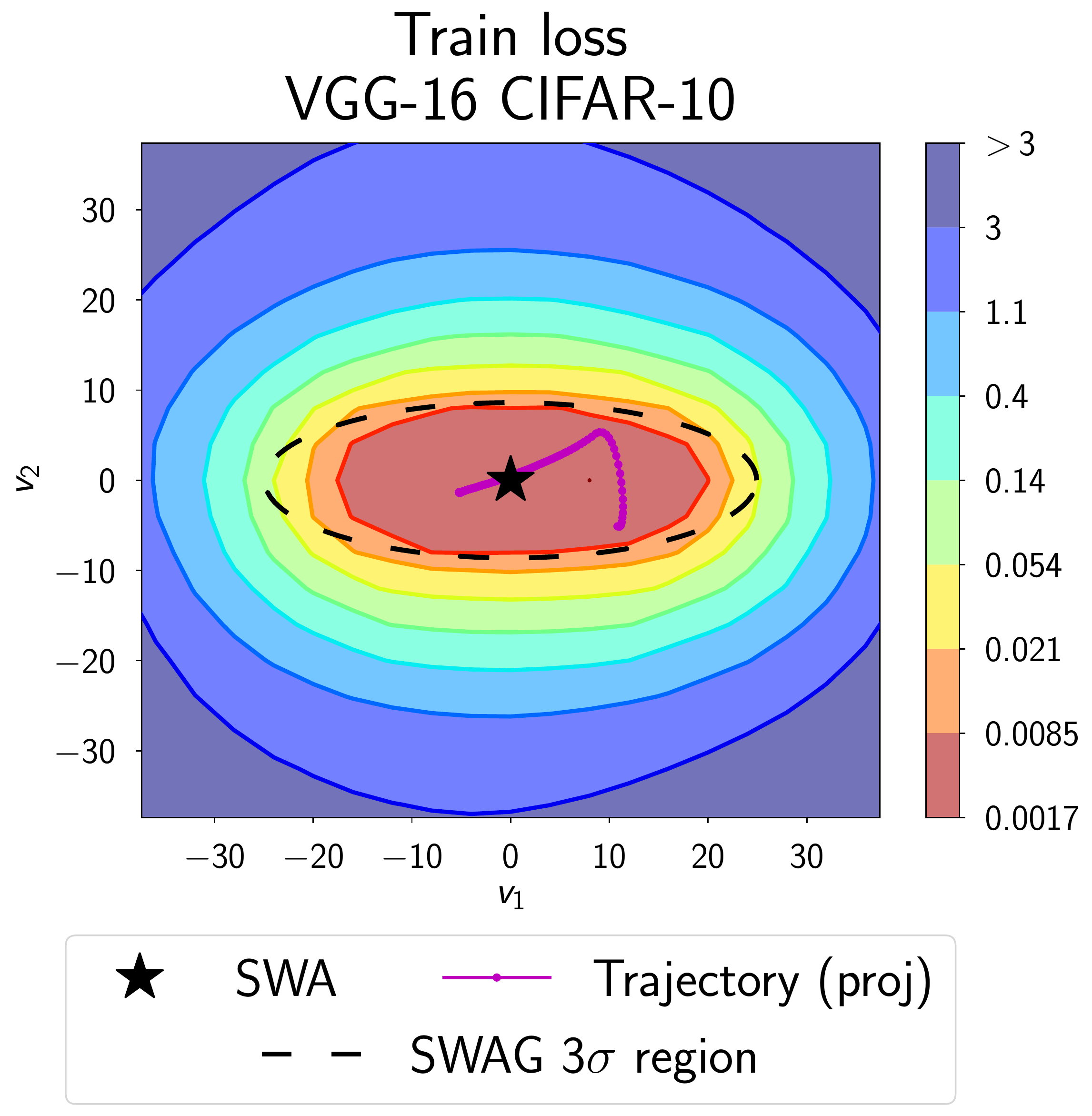}
	\end{subfigure}
	\quad
	\begin{subfigure}[u]{0.30\textwidth}
		\centering
		\includegraphics[width=\textwidth]{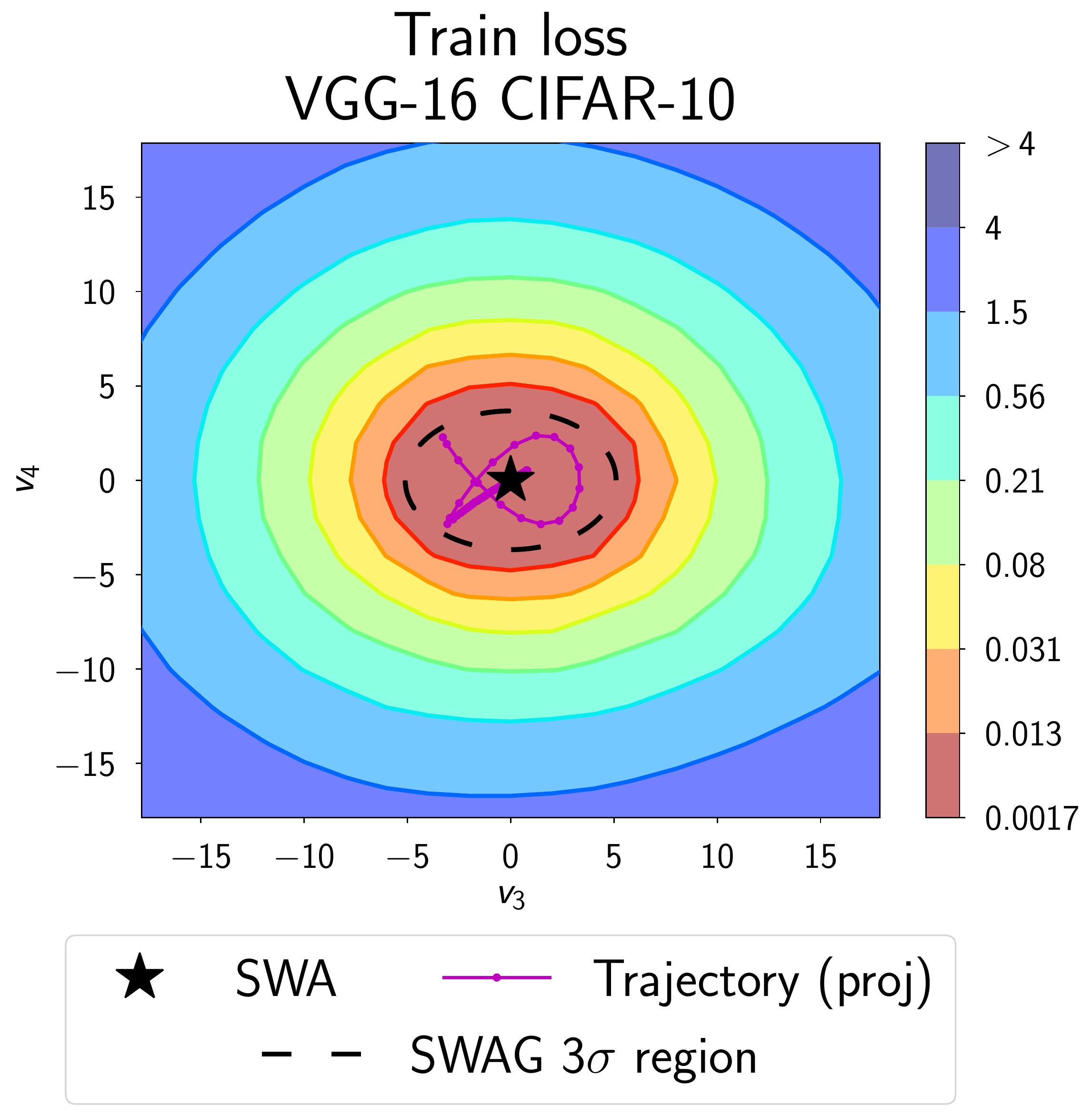}
	\end{subfigure}
	\caption{ 
		\textbf{Left:} Posterior-density cross-sections along the rays corresponding to different
		eigenvectors of the SWAG covariance matrix.
		\textbf{Middle:} Posterior-density surface in the plane spanned by eigenvectors of SWAG 
		covariance matrix corresponding to the first and second largest eigenvalues and
		(\textbf{Right:}) the third and fourth largest eigenvalues. 
		Each row in the figure corresponds to an architecture-dataset pair indicated in the title
		of each panel.}
	\label{fig:allplanes}
\end{figure*}

In Figure \ref{fig:allplanes} we present plots analogous to those in Section
\ref{sec:geometry} for PreResNet-110 and VGG-16 on CIFAR-10 and CIFAR-100.
For all dataset-architecture pairs we see that SWAG is able to capture
the geometry of the posterior in the subspace spanned by SGD trajectory.

\clearpage
\section{Hyper-Parameters and Limitations}\label{app:hypers}
In this section, we discuss the hyper-parameters in SWAG, as well as some current theoretical limitations.

\subsection{Rank of Covariance Matrix}\label{sec:rank}

\begin{figure*}
	\begin{subfigure}{0.245\textwidth}
		\centering
		\includegraphics[width=\textwidth]{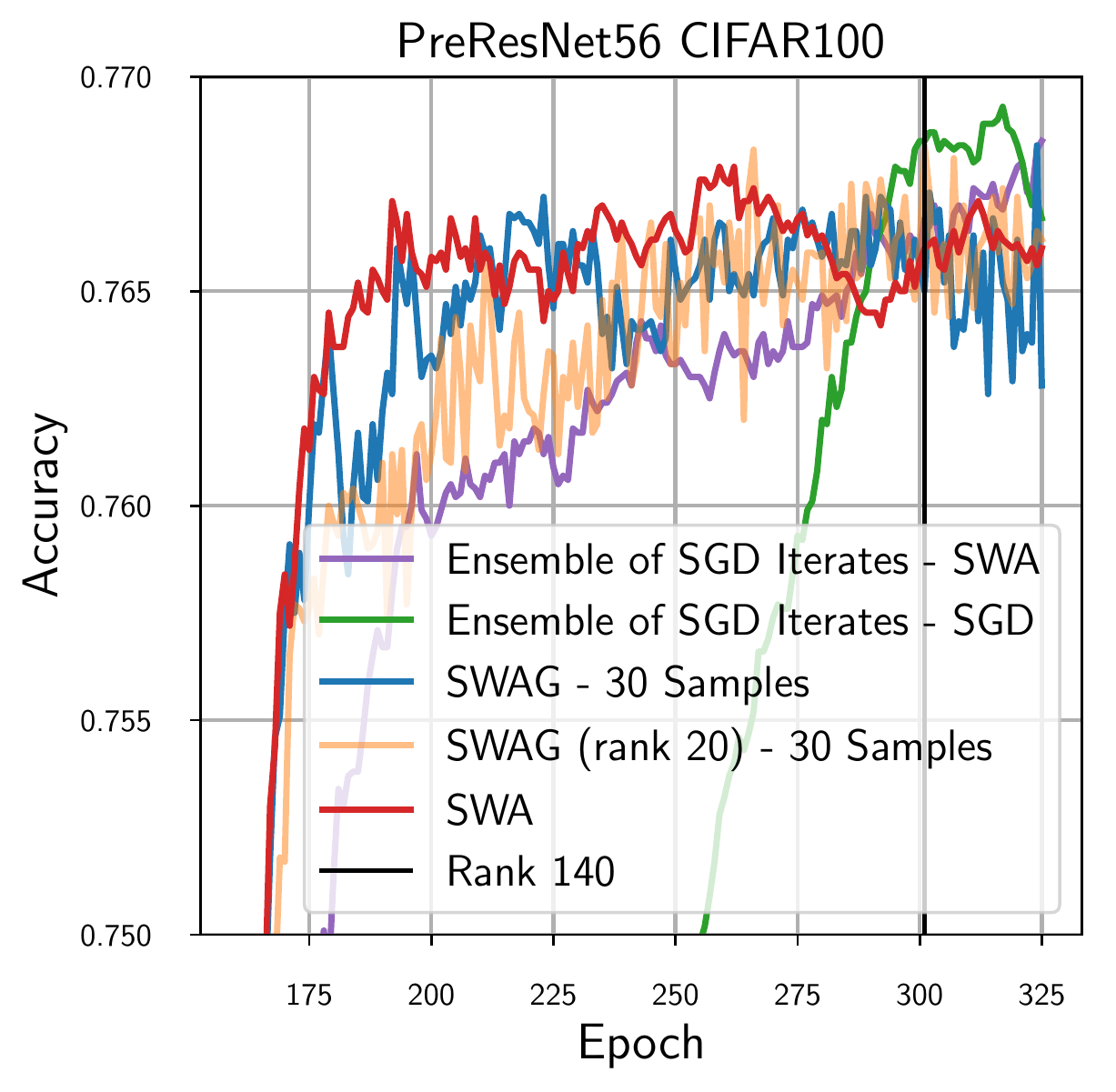}
		\caption{}
		\label{fig:emp_ensembl}
	\end{subfigure}
	\begin{subfigure}{0.245\textwidth}
		\centering
		\includegraphics[width=\textwidth]{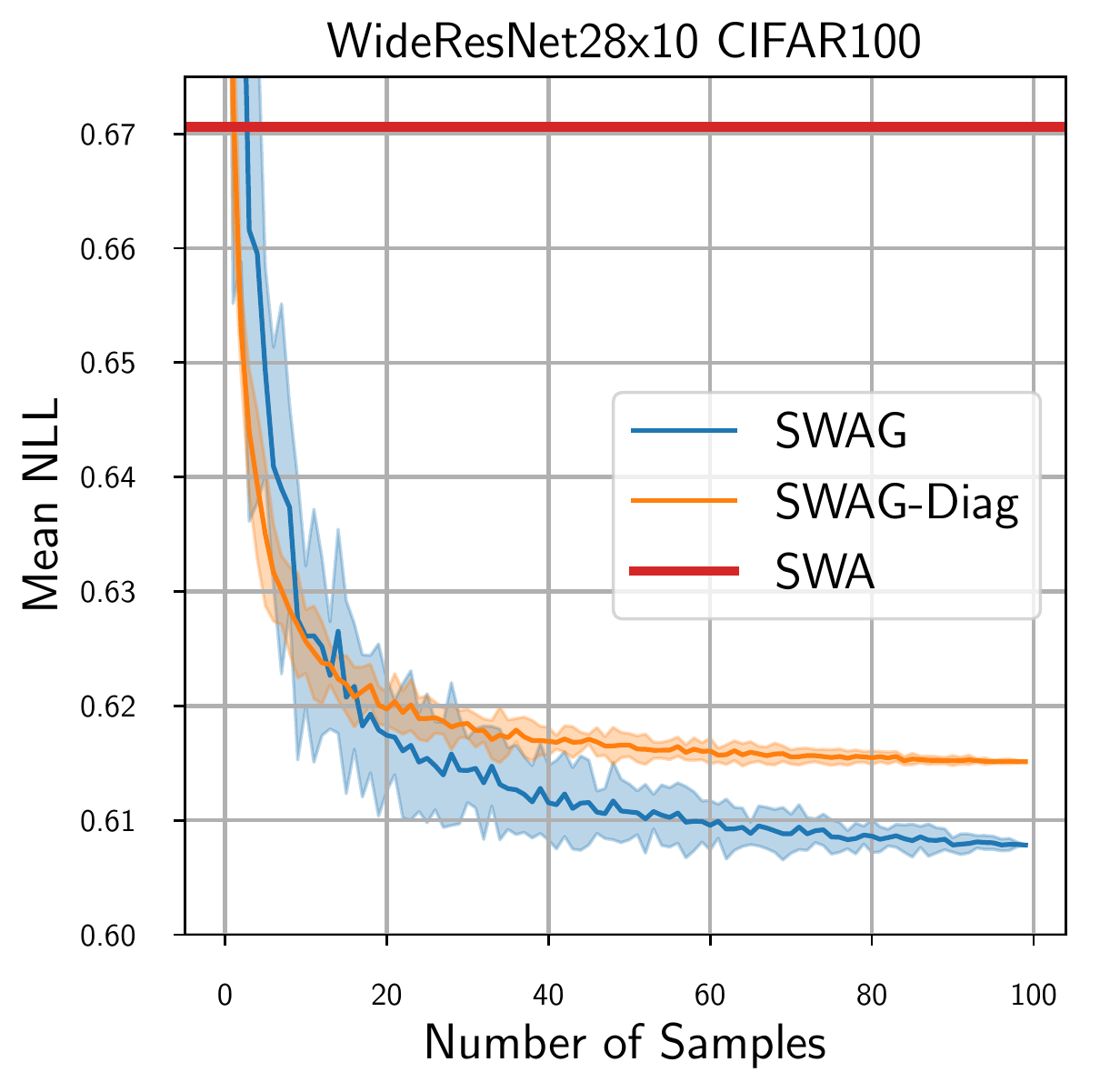}
		\caption{}
		\label{fig:ens_nll}
	\end{subfigure}
	\begin{subfigure}{0.245\textwidth}
		\centering
		\includegraphics[width=\textwidth]{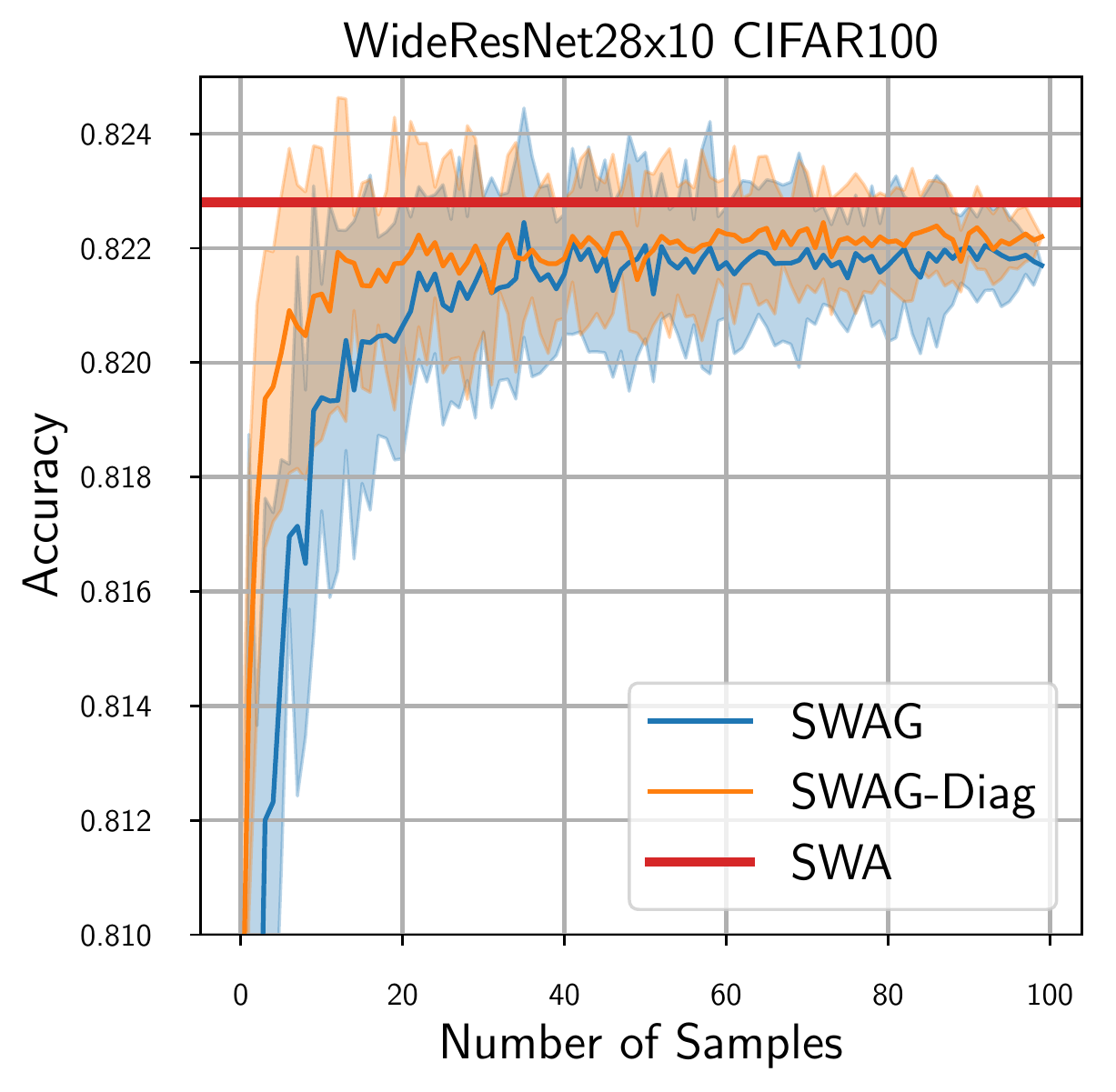}
		\caption{}
		\label{fig:ens_acc}
	\end{subfigure}	
	\begin{subfigure}{0.245\textwidth}
		\includegraphics[width=\textwidth]{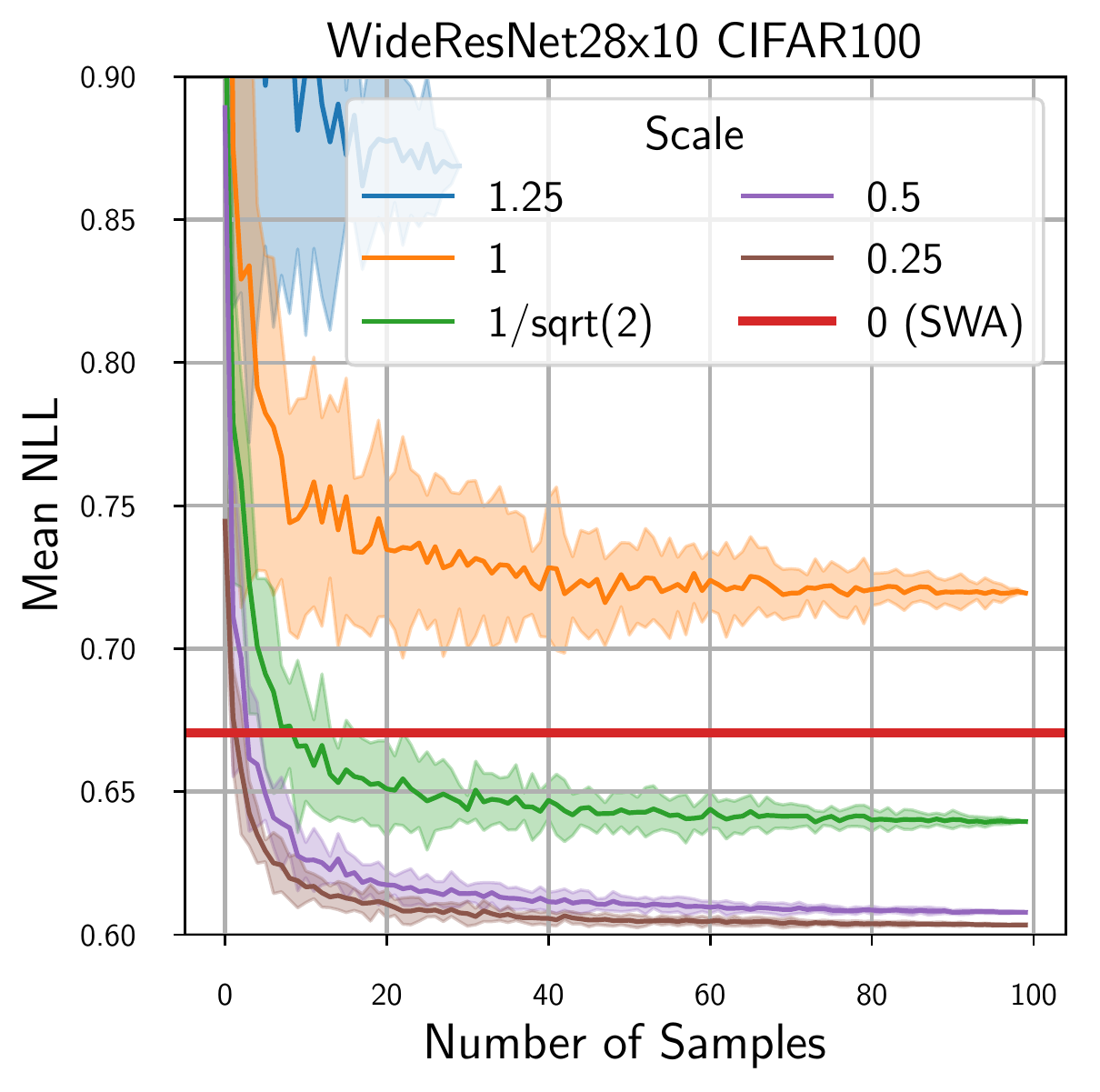}
		\caption{}
		\label{fig:scaling}
	\end{subfigure}	
	
	\caption{\textbf{(a)} 30 samples of SWAG with a rank 20 covariance matches the SWA 
		result over the course of training for PreResNet56 on CIFAR-100.  SWAG with 
		a rank of 140, SWAG with a rank of 20, and SWA all outperform ensembles of 
		SGD iterates from the SWA procedure and from a standard SGD training path.
		\textbf{(b)} NLL and \textbf{(c)} accuracy by number of samples for 
		WideResNet on CIFAR-100 for SWAG, SWAG-Diag, and SWA. 30 samples is adequate 
		for stable accuracies and NLLs. 
		\textbf{(d)} NLL by number of samples for different scales for WideResNet on CIFAR-100 
		for SWAG, SWAG-Diag, and SWA. Scales beneath 1 perform better, with 0.5 and 
		0.25 best.}
	\label{fig:ensemble}
\end{figure*}

We now evaluate the effect of the covariance matrix rank on the SWAG approximation.
To do so, we trained a PreResNet56 on CIFAR-100 with SWAG beginning from epoch 161, 
and evaluated 30 sample Bayesian model averages obtained at different epochs; 
the accuracy plot from this experiment is shown in Figure \ref{fig:ensemble} (a).
The rank of each model after epoch 161 is simply $\min(epoch - 161, 140),$ and 30 samples from even a low rank approximation reach the same predictive accuracy as the SWA model.
Interestingly, both SWAG and SWA outperform ensembles of a SGD run and ensembles of the SGD models in the SWA run.

\subsection{Number of Samples in the Forwards Pass}
In most situations where SWAG will be used, no closed form expression for the integral $\int f(y) q(\theta |y) d\theta,$ will exist.
Thus, Monte Carlo approximations will be used; Monte Carlo integration converges at a rate of $1/\sqrt{K},$ where $K$ is the number of samples used, but practically good results may be found with very few samples (e.g. Chapter 29 of \citet{mackay_information_2003}).

To test how many samples are needed for good predictive accuracy in a Bayesian model averaging task, we used a rank 20 approximation for SWAG and then tested the NLL on the test set as a function of the number of samples for WideResNet28x10 \citep{zagoruyko_wide_2016} on CIFAR-100.

The results from this experiment are shown in Figure \ref{fig:ensemble} (b, c), where it is possible to see that about 3 samples will match the SWA result for NLL, with about 30 samples necessary for stable accuracy (about the same as SWA for this network).
In most of our experiments, we used 30 samples for consistency.
In practice, we suggest tuning this number by looking at a validation set as well as the computational resources available and comparing to the free SWA predictions that come with SWAG.

\subsection{Dependence on Learning Rate}
First, we note that the covariance, $\Sigma,$ estimated using SWAG, is a function of the learning rate (and momentum) for SGD. 
While the theoretical work of \citet{mandt_stochastic_2017} suggests that it is possible to optimally set the learning rate, our experiments in Appendix \ref{app:assumptions} 
show that currently the assumptions of the theory do not match the empirical reality in deep learning.
In practice the learning rate can be chosen to maximize negative log-likelihood 
on a validation set.
In the linear setting as in \citet{mandt_stochastic_2017}, the learning rate controls the scale of the asymptotic covariance matrix. 
If the optimal learning rate (Equation \ref{eq:optlr}) is used in this setting, the covariance matches the true posterior.
To attempt to disassociate the learning rate from the covariance in practice,
we rescale the covariance matrix when sampling by a constant factor for a WideResNet on CIFAR-100 shown in Figure \ref{fig:ensemble} (d).

Over several replications, we found that a scale of 0.5 worked best, which is expected because the low rank plus diagonal covariance incorporates the variance twice (once for the diagonal and once from the low rank component).

\subsection{Necessity of Batch Norm Updates}\label{app:batch_norm}
\begin{figure}[!htb]
	\centering
	\includegraphics[width=0.5\linewidth]{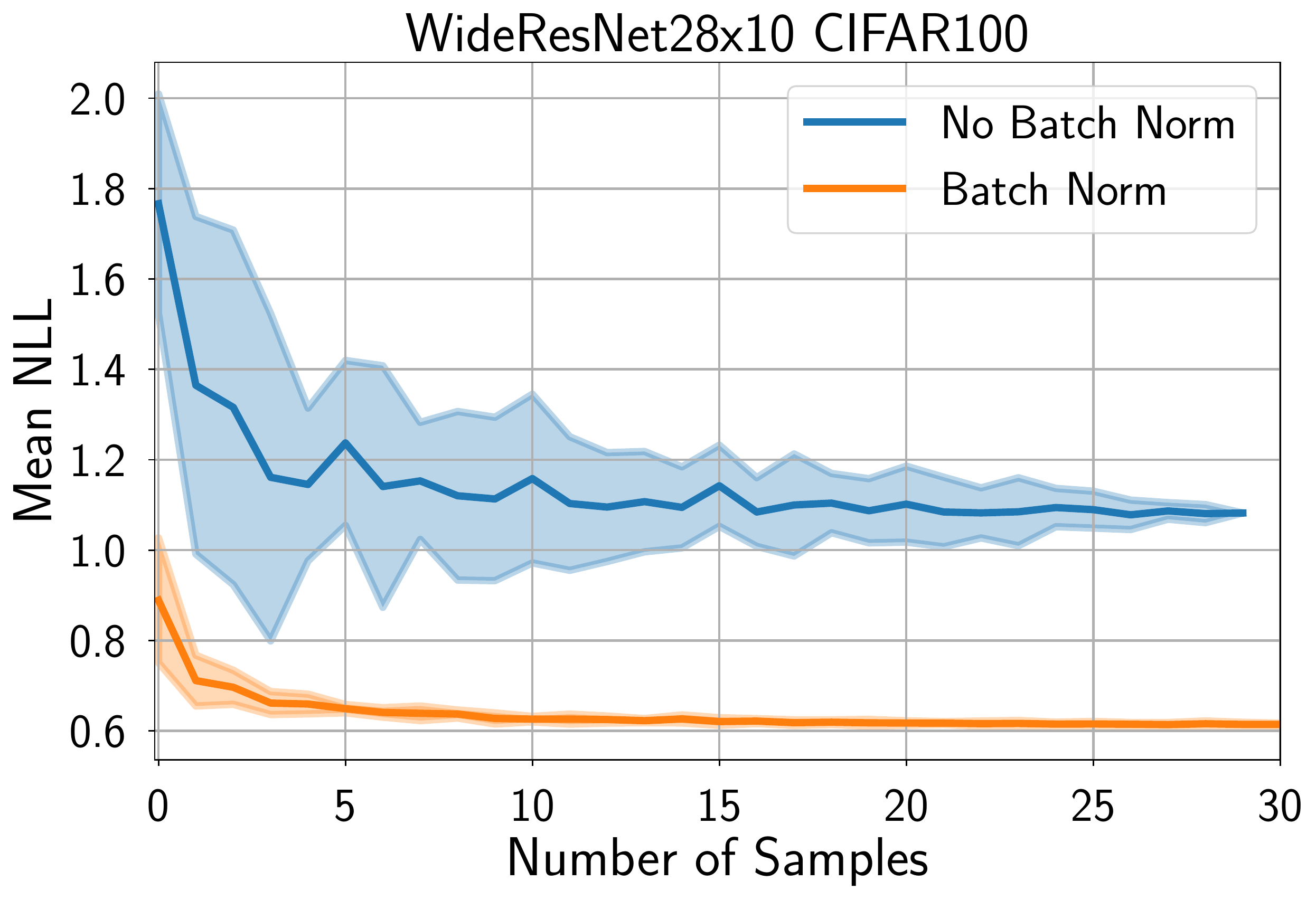}
	\caption{NLL by number of samples for SWAG with and without batch norm updates after sampling. Updating the batch norm parameters after sampling results in a significant improvement in NLL.}
	\label{fig:bn_update}
\end{figure}

One possible slowdown of SWAG at inference time is in the usage of updated batch norm parameters. 
Following \citet{izmailov_averaging_2018}, we found that in order for the averaging and sampling to work well, it was necessary to update the batch norm parameters of networks after sampling a new model. 
This is shown in Figure \ref{fig:bn_update} for a WideResNet on CIFAR-100 for two independently trained models.

\subsection{Usage in Practice}

From our experimental findings, we see that given an equal amount of training time, SWAG typically outperforms other methods for uncertainty calibration.
SWAG additionally does not require a validation set like temperature scaling and Platt scaling (e.g.  \citet{guo_calibration_2017,kuleshov_accurate_2018}).
SWAG also appears to have a distinct advantage over temperature scaling, and other popular alternatives, 
when the target data are from a different distribution than the training
data, as shown by our transfer learning experiments.

Deep ensembles \citep{lakshminarayanan_simple_2017} require several times longer training for equal calibration, but often perform somewhat better due to incorporating several independent training runs. Thus SWAG will be particularly valuable when training time is limited, but inference time may not be.
One possible application is thus in medical applications when image sizes (for semantic segmentation) are large, but predictions can be parallelized and may not have to be instantaneous.

\begin{figure*}
	\centering
	\begin{subfigure}{0.30\textwidth}
		\centering
		\includegraphics[width=\textwidth]{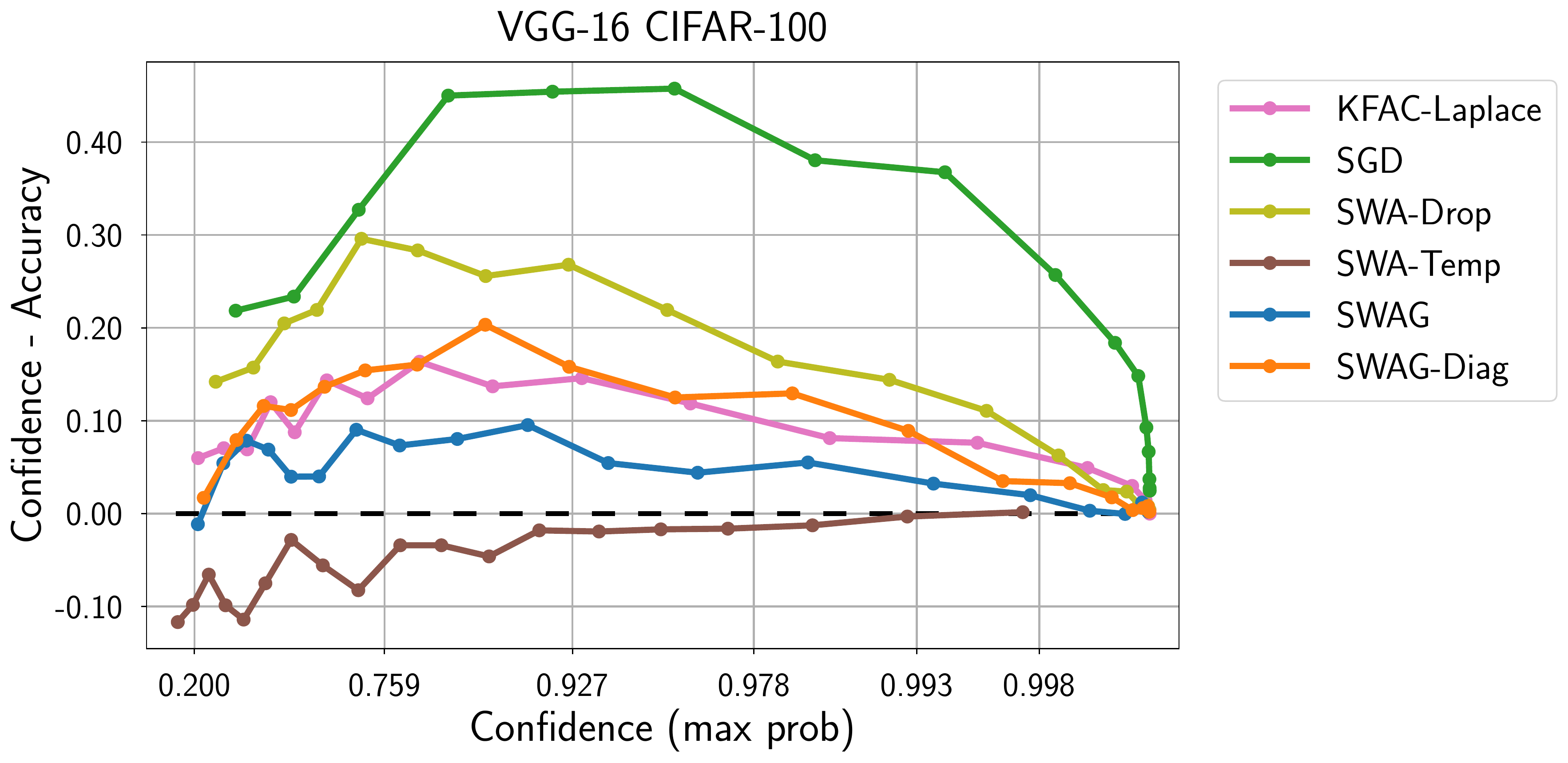}
	\end{subfigure}
	\quad
	\begin{subfigure}{0.30\textwidth}
		\centering
		\includegraphics[width=\textwidth]{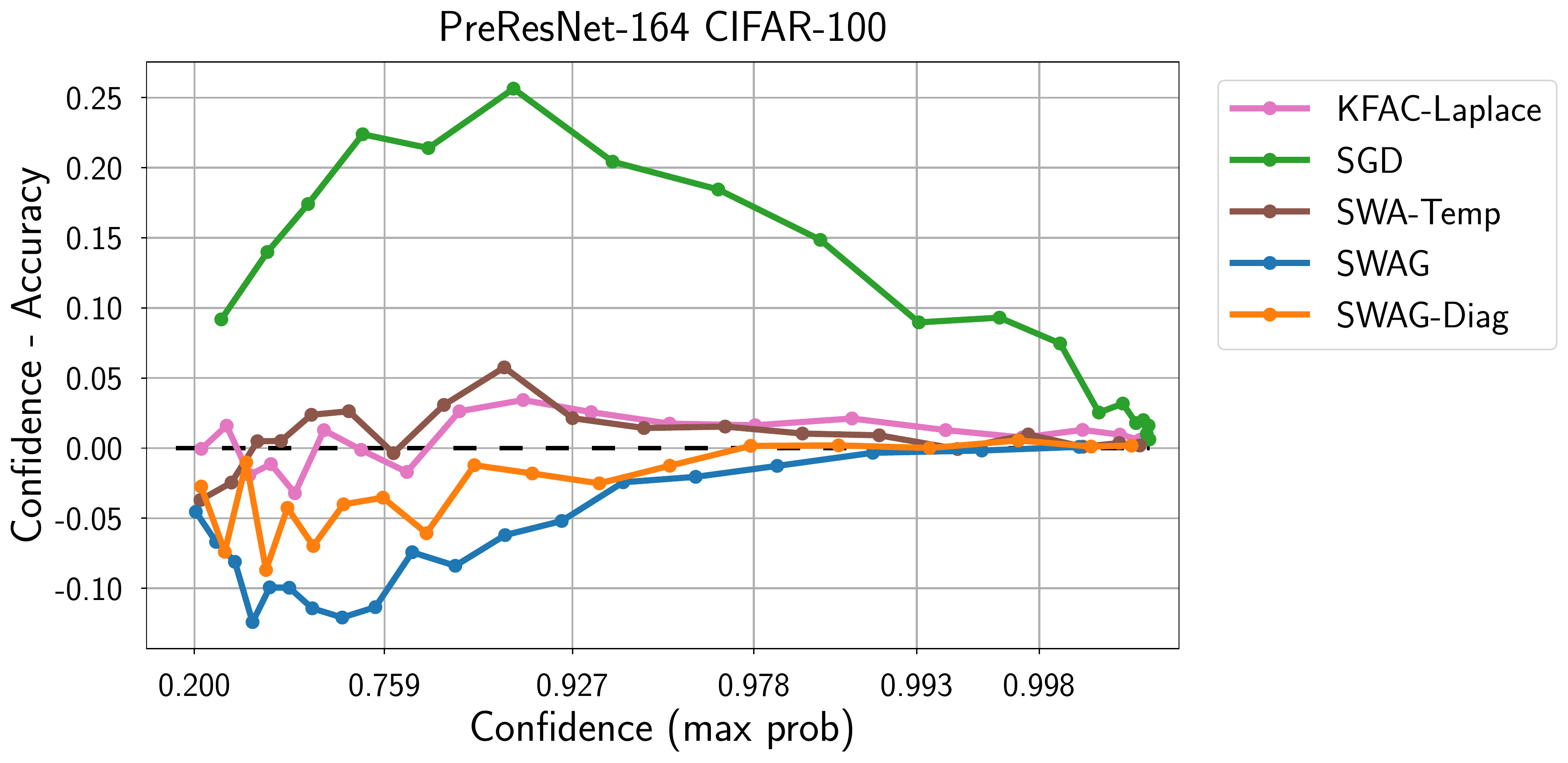}
	\end{subfigure}
	\quad
	\begin{subfigure}{0.30\textwidth}
		\centering
		\includegraphics[width=\textwidth]{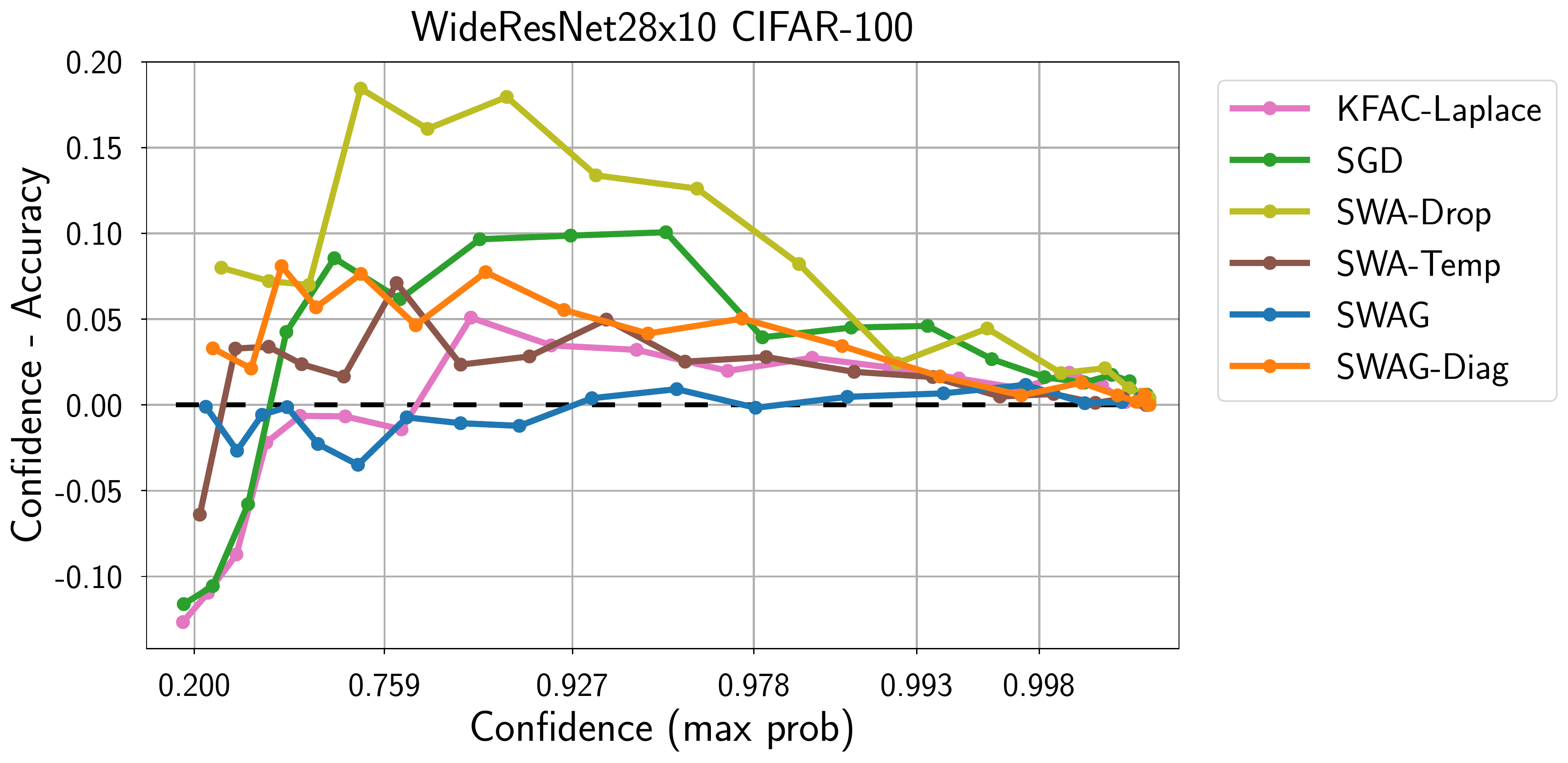}
	\end{subfigure}
	\begin{subfigure}{0.30\textwidth}
		\centering
		\includegraphics[width=\textwidth]{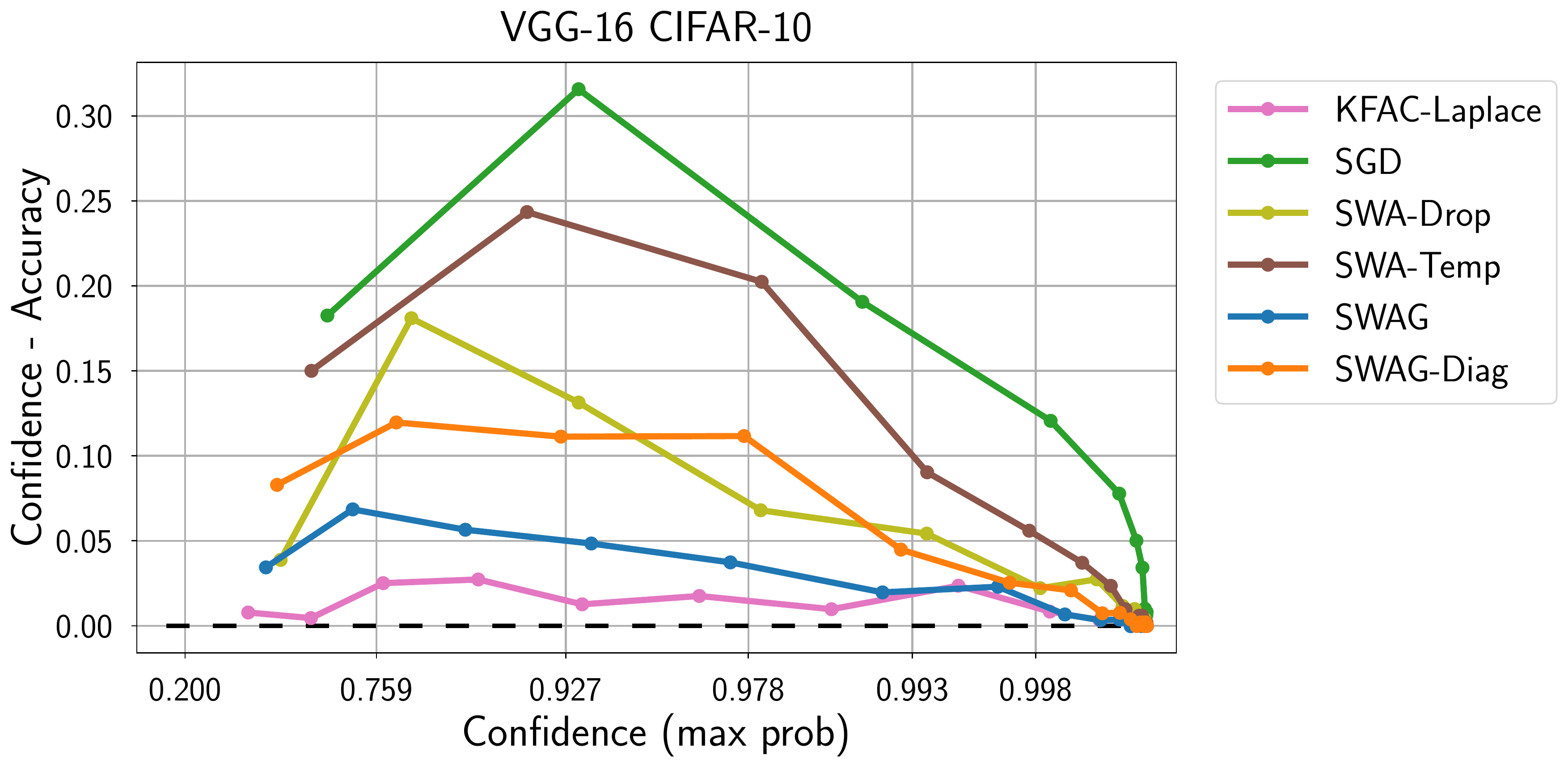}
	\end{subfigure}
	\quad
	\begin{subfigure}{0.30\textwidth}
		\centering
		\includegraphics[width=\textwidth]{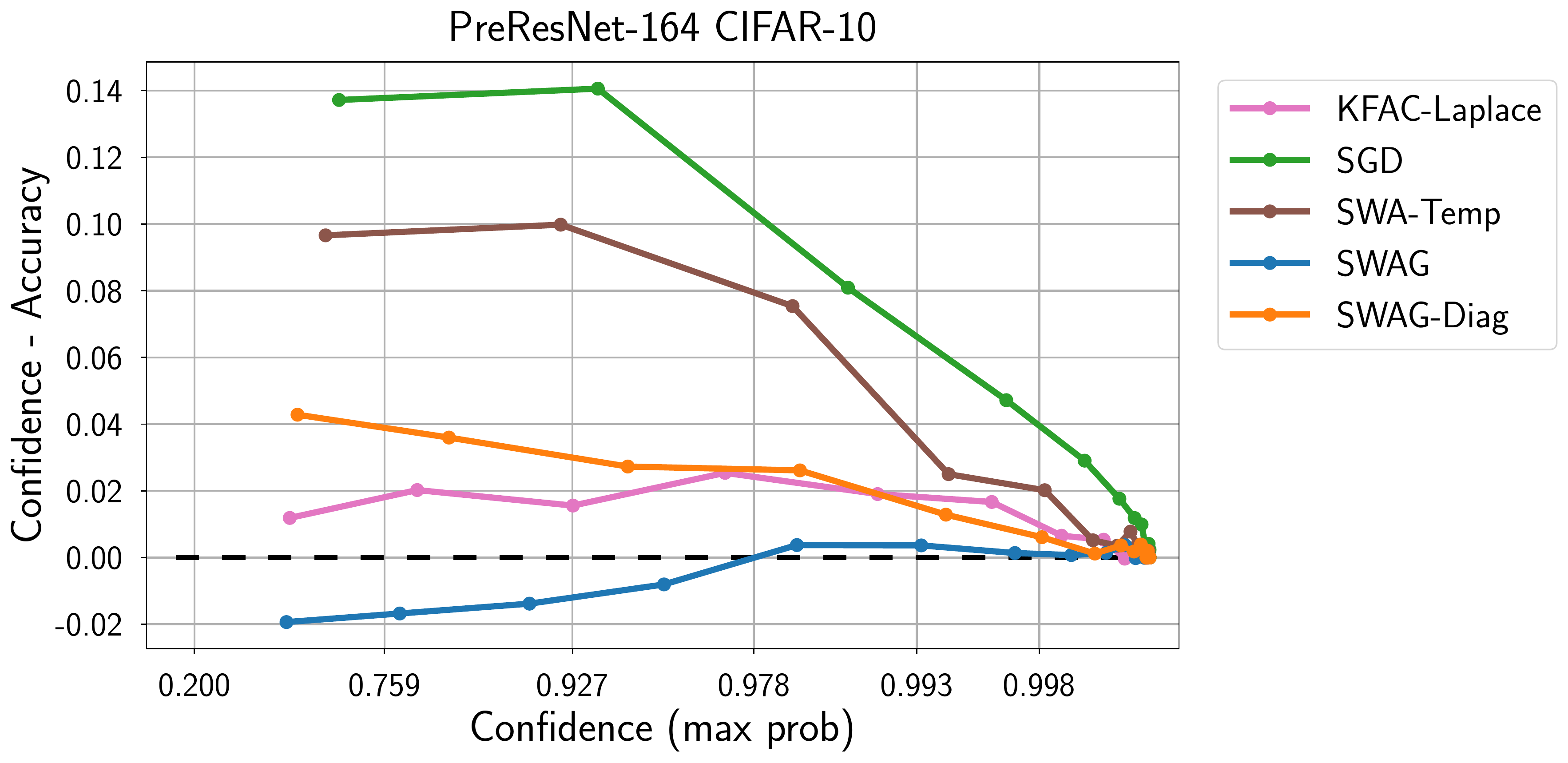}
	\end{subfigure}
	\quad
	\begin{subfigure}{0.30\textwidth}
		\centering
		\includegraphics[width=\textwidth]{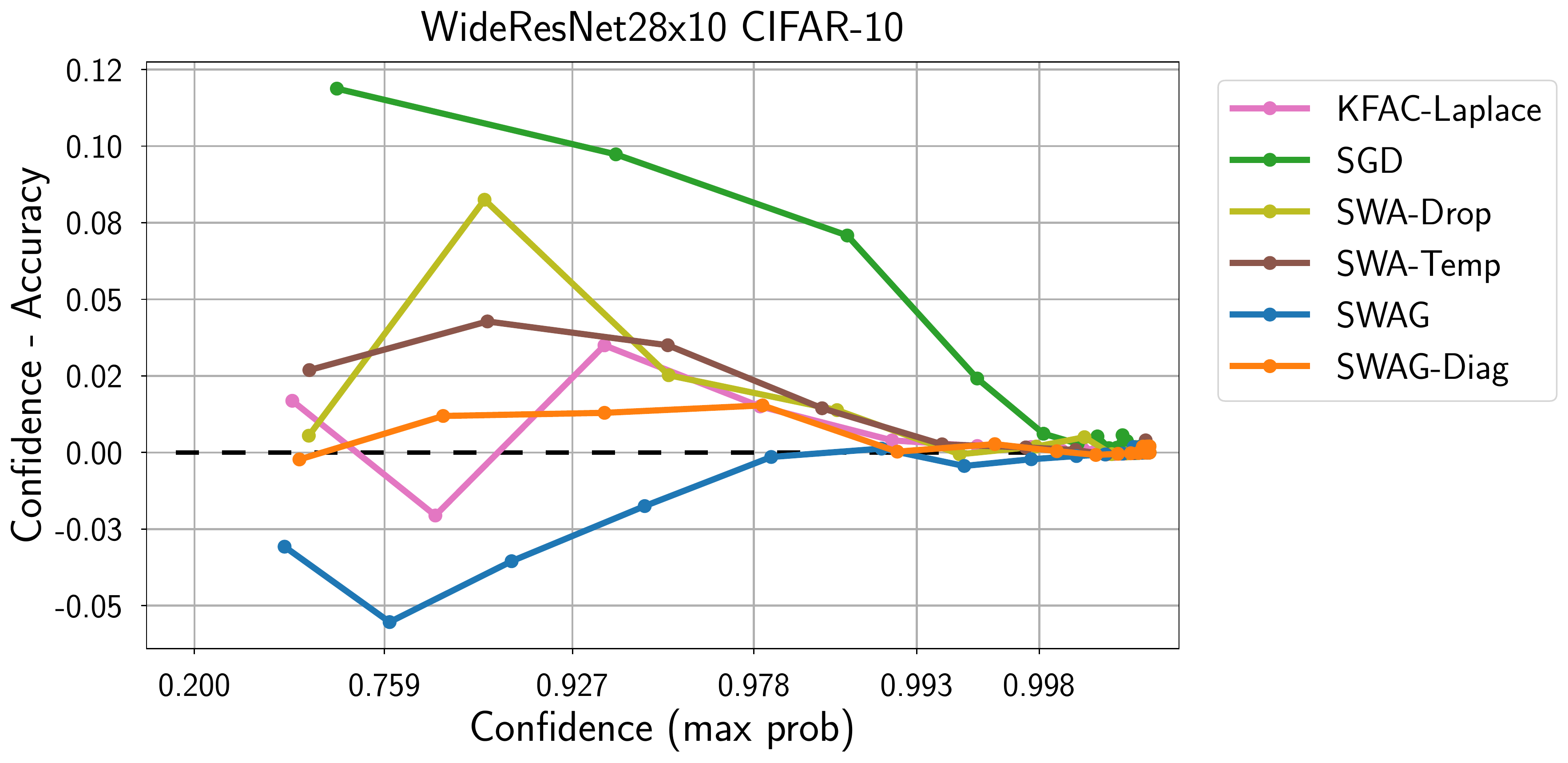}
	\end{subfigure}
	\begin{subfigure}{0.30\textwidth}
		\centering
		\includegraphics[width=\textwidth]{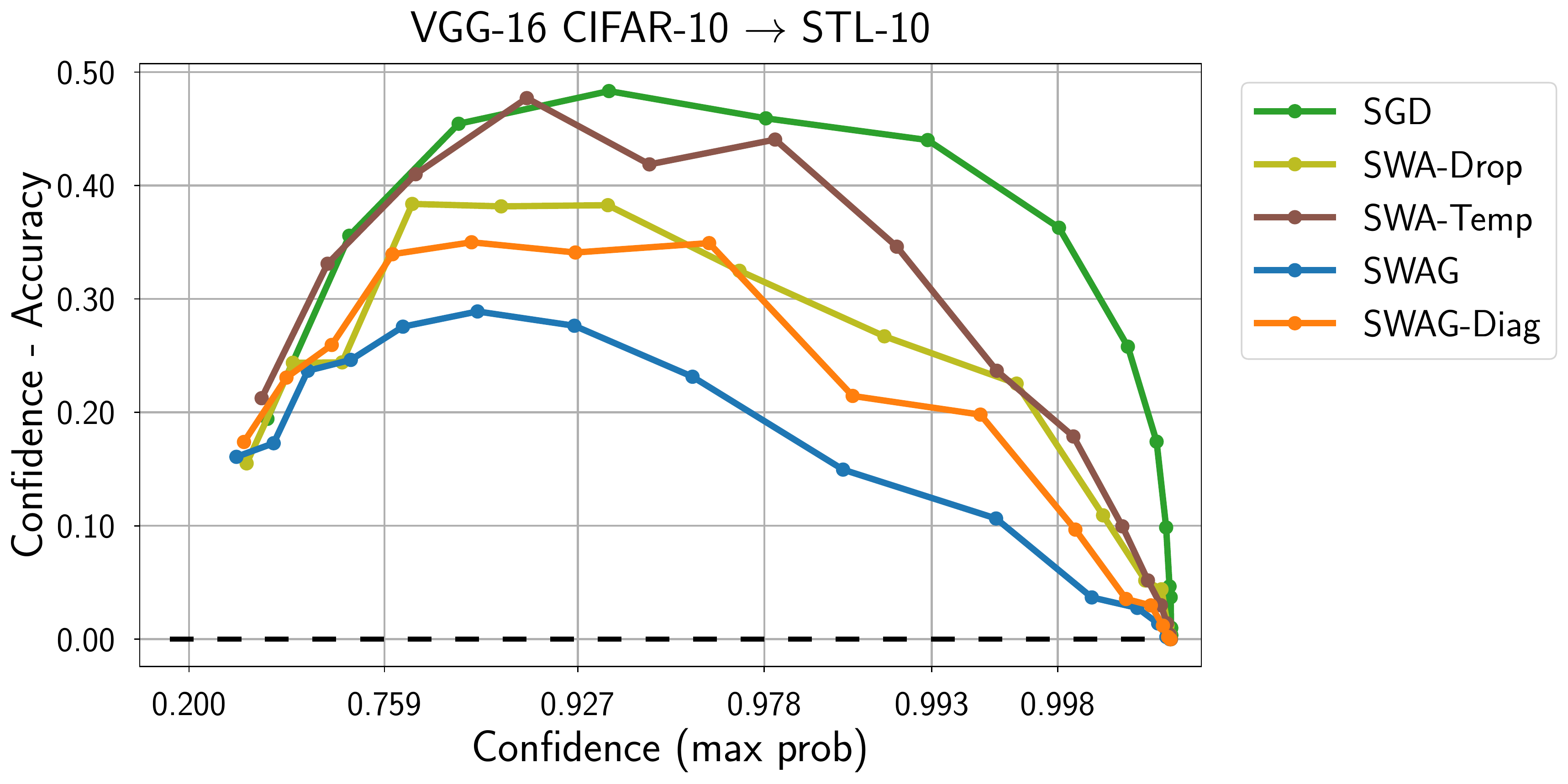}
	\end{subfigure}
	\quad
	\begin{subfigure}{0.30\textwidth}
		\centering
		\includegraphics[width=\textwidth]{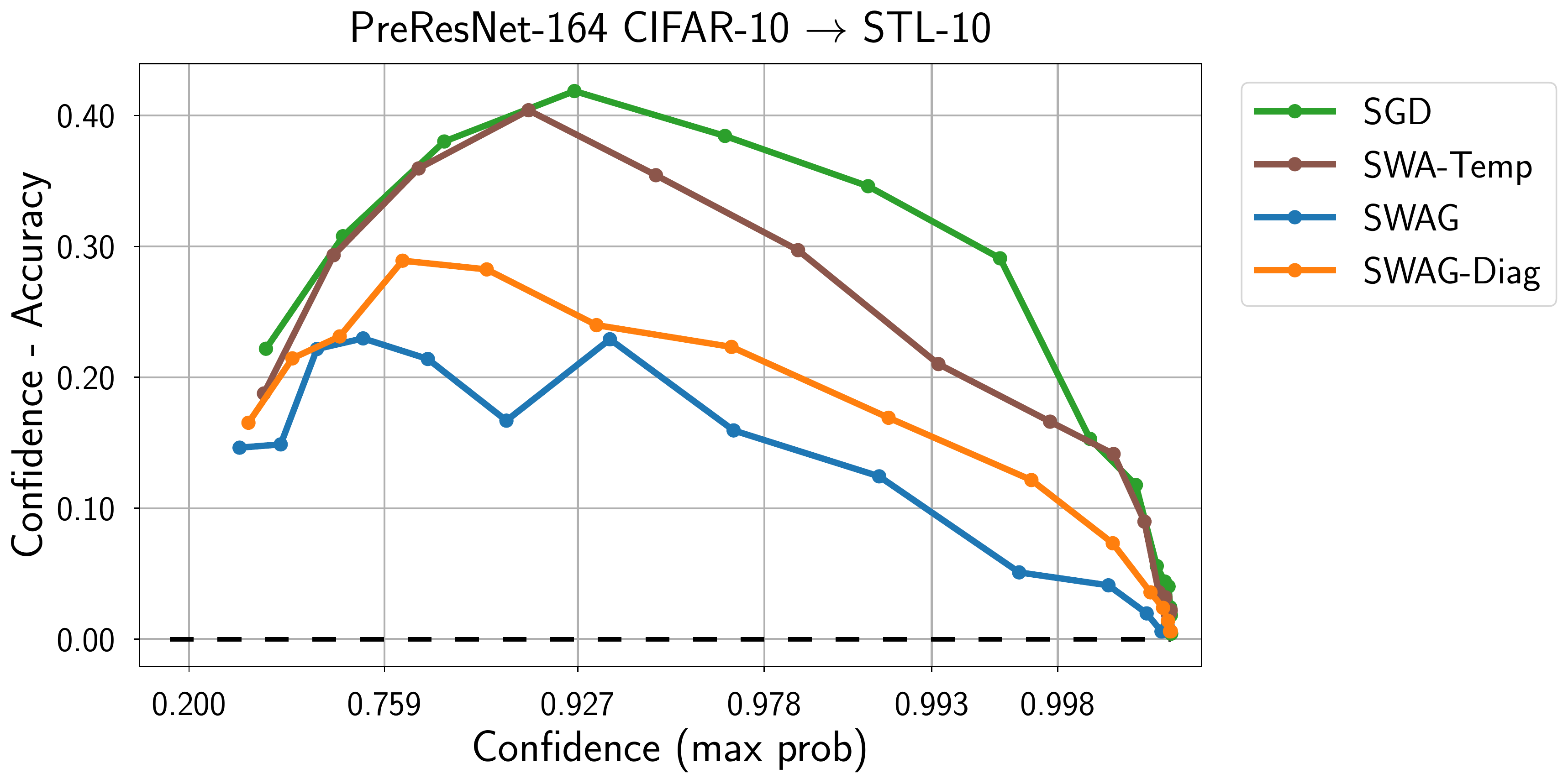}
	\end{subfigure}
	\quad
	\begin{subfigure}{0.30\textwidth}
		\centering
		\includegraphics[width=\textwidth]{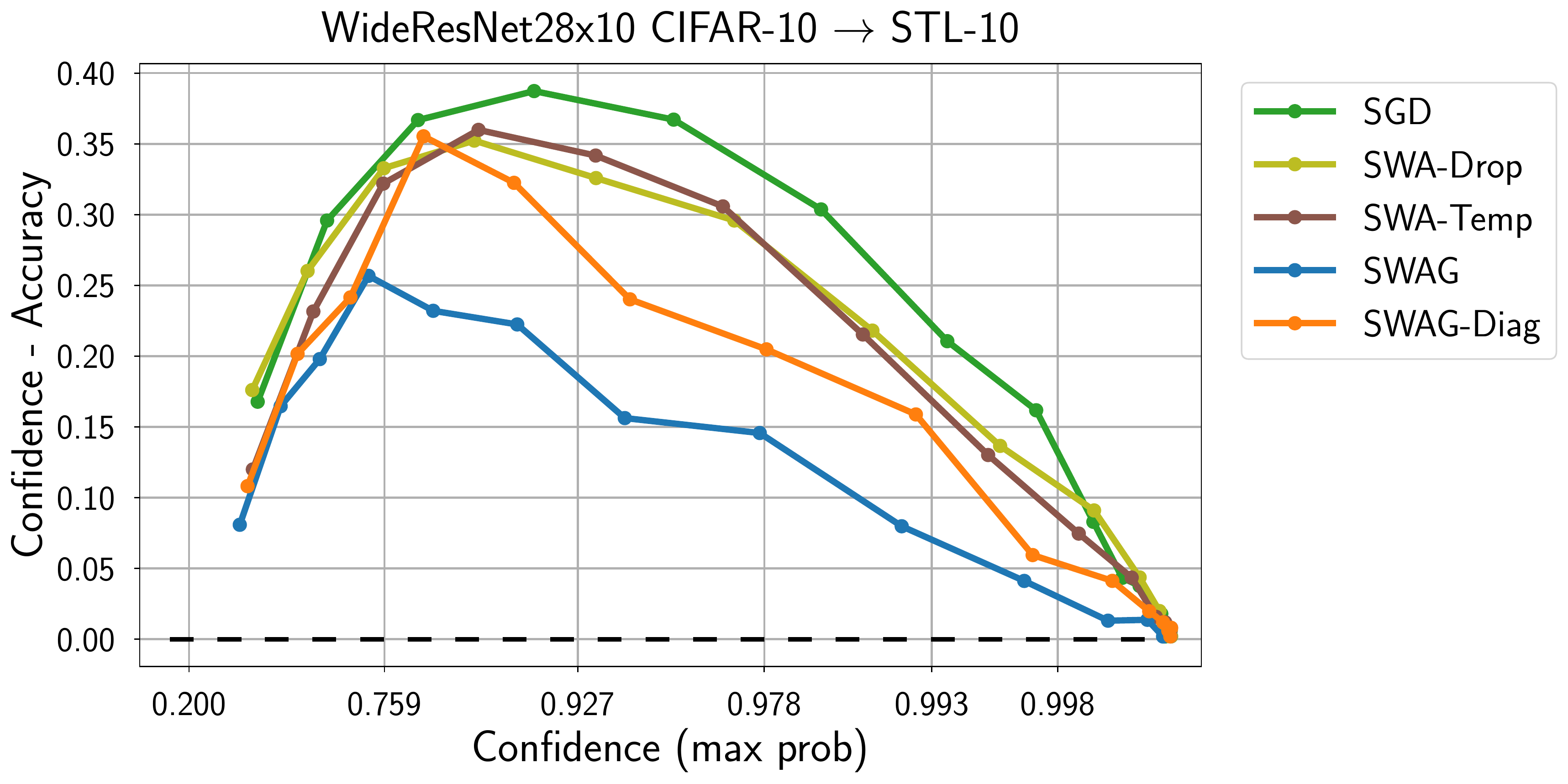}
	\end{subfigure}
	\begin{subfigure}{0.30\textwidth}
		\centering
		\includegraphics[width=\textwidth]{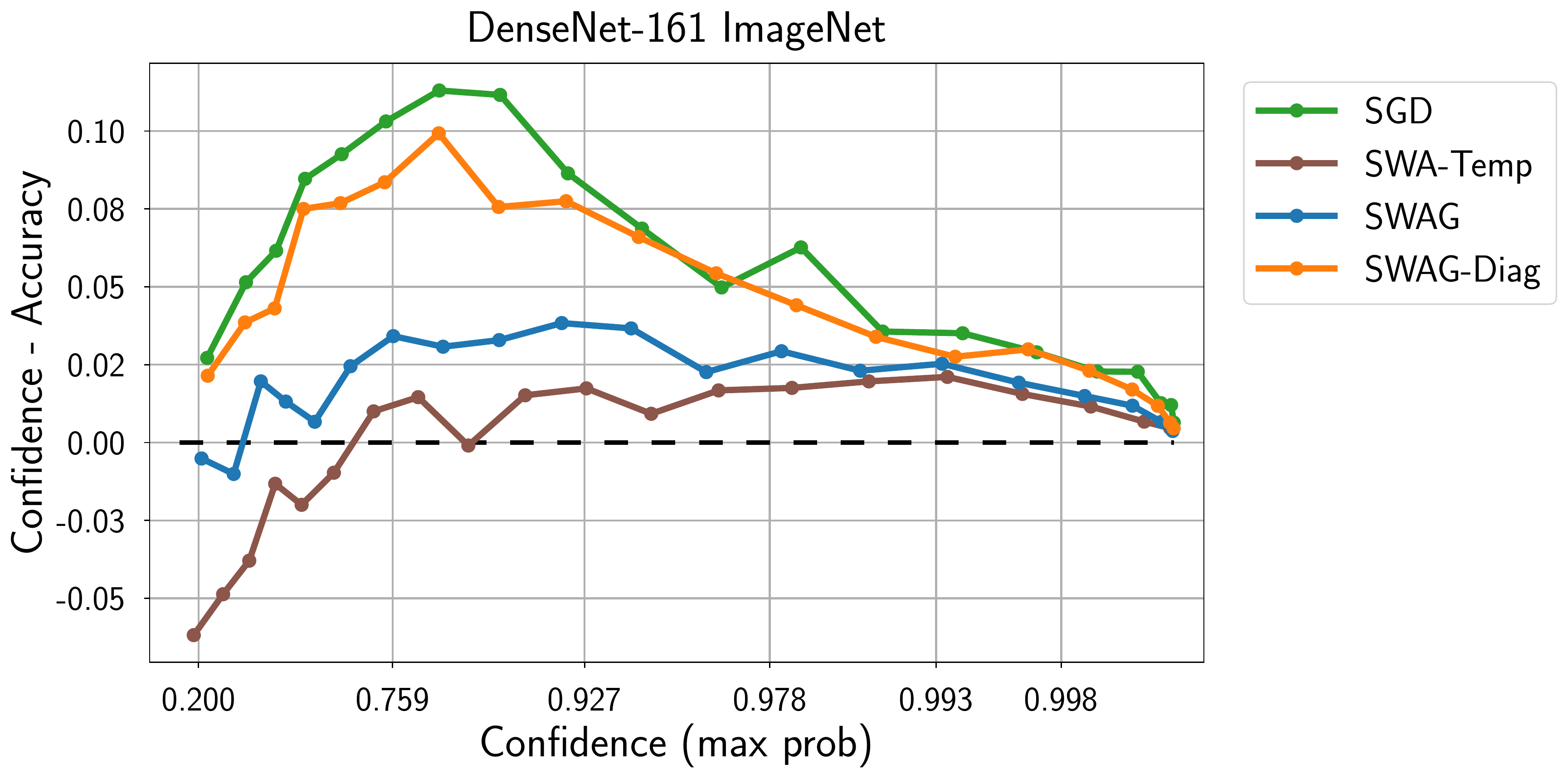}
	\end{subfigure}
	\quad
	\begin{subfigure}{0.30\textwidth}
		\centering
		\includegraphics[width=\textwidth]{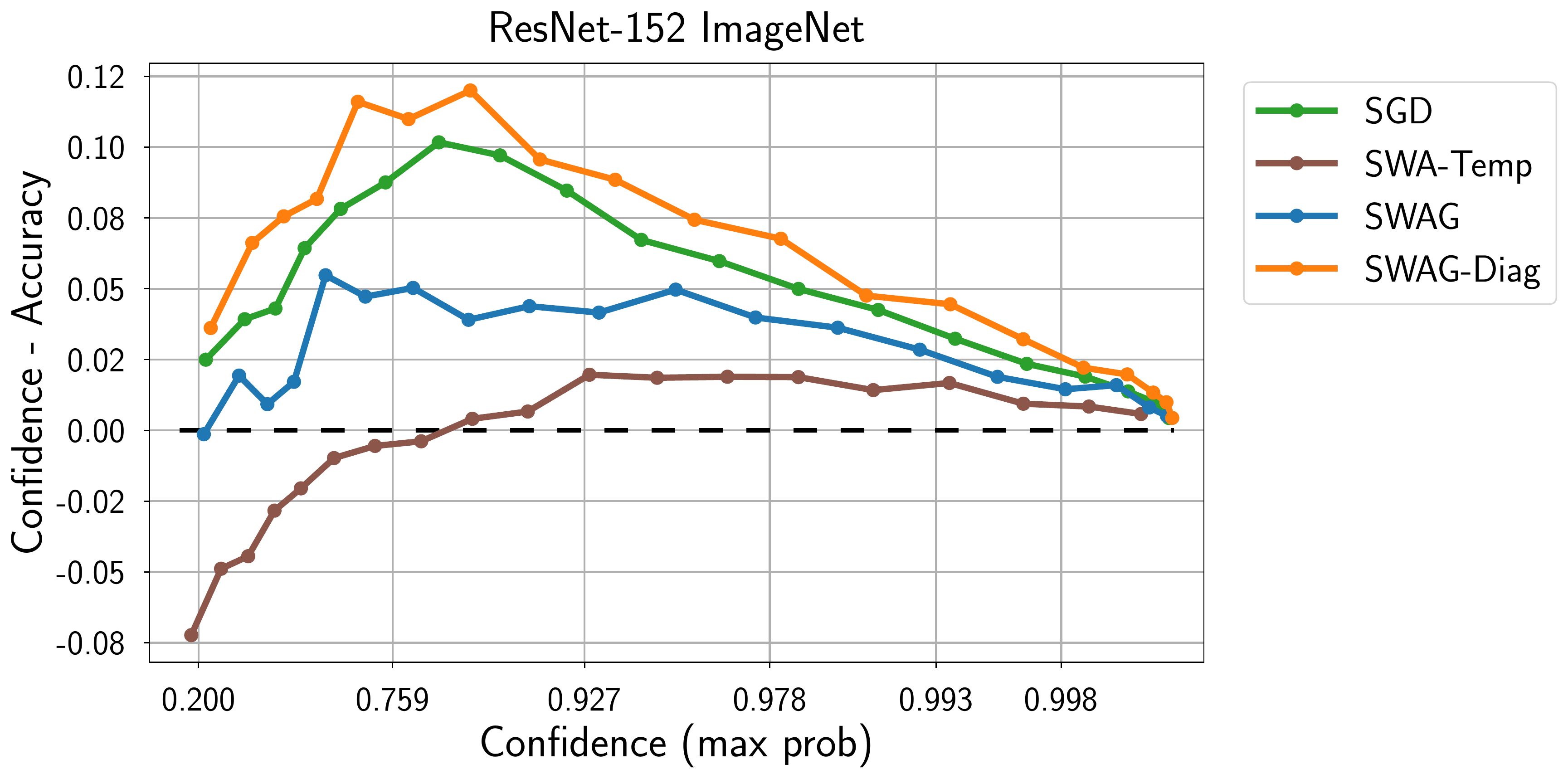}
	\end{subfigure}
	\caption{Reliability diagrams (see Section \ref{sec:exp_calibration}) for all
		models and datasets. The dataset and architecture are listed in the title of
		each panel.} 
	\label{fig:rel_diags}
\end{figure*}

\section{Further Classification Uncertainty Results}

\subsection{Reliability Diagrams}\label{app:reliability}

We provide the additional reliability diagrams for all methods and datasets in
Figure \ref{fig:rel_diags}. SWAG consistently improves calibration
over SWA, and performs on par or better than temperature scaling. In transfer
learning temperature scaling fails to achieve good calibration, while
SWAG still provides a significant improvement over SWA.

\subsection{Out-of-Domain Image Detection}\label{app:oos}
\begin{figure*}[!h]
	\centering
	\includegraphics[width=1.\textwidth]{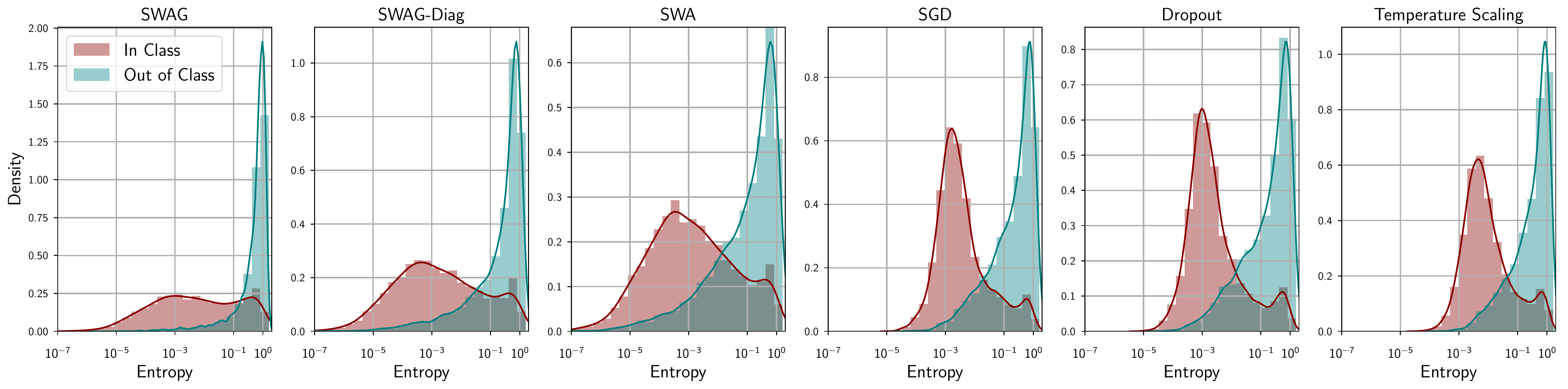}
	\caption{In and out of sample entropy distributions for WideResNet28x10 on CIFAR5 + 5.}
	\label{fig:c5_entropy_dist}
\end{figure*}
\label{sec:exp_out}
\begin{table}[!h]
	\centering
	\caption{Symmetrized, discretized KL divergence between 
		the distributions of predictive entropies for data  from the first and last 
		five classes of CIFAR-10 for models trained only on the first five classes.
		The entropy distributions for SWAG are more different than the baseline models.
	}
	\label{tab:in_out}
	\vspace{0.2cm}
	{\small
		\begin{tabular}{lcccc}
			\toprule
			Method & JS-Distance \\
			\midrule
			SWAG & \textbf{3.31}  \\
			SWAG-Diag & 2.27 \\ 
			MC Dropout &  3.04 \\ 
			SWA &  1.68 \\ 
			SGD (Baseline) & 3.14  \\ 
			SGD + Temp. Scaling & 2.98 \\ \bottomrule
		\end{tabular}
	}
\end{table}
Next, we evaluate the SWAG variants along with the baselines on out-of-domain data detection. 
To do so we train a WideResNet as described in Section \ref{app:details} on the data from five classes of the CIFAR-10 dataset, and then analyze their predictions on the full test set. 
We expect the outputted class probabilities on objects that belong to classes that 
were not present in the training data to have high-entropy reflecting the model's 
high uncertainty in its predictions, and considerably lower entropy on the images 
that are similar to those on which the network was trained. 

To make this comparison quantitative, we computed the symmetrized KL divergence between the binned in and out of sample distributions in Table \ref{tab:in_out}, finding that SWAG and Dropout perform best on this measure.
We plot the histograms of predictive entropies on the in-domain (classes that were trained on) and out-of-domain (classes that were not trained on) in Figure A.\ref{fig:c5_entropy_dist} for a qualitative comparison.

Table \ref{tab:in_out} shows the computed symmetrized, discretized KL distance between in and out of sample distributions for the CIFAR5 out of sample image detection class.
We used the same bins as in Figure \ref{fig:c5_entropy_dist} to discretize the entropy distributions, then smoothed these bins by a factor of 1e-7 before calculating $KL(\text{IN} || \text{OUT}) + KL(\text{OUT} || \text{IN})$ using the \texttt{scipy.stats.entropy} function.
We can see even qualitatively that the distributions are more distinct for SWAG and SWAG-Diagonal than for the other methods, particularly temperature scaling.

\subsection{Tables of ECE, NLL, and Accuracy.}
\label{app:tables}

We provide test accuracies (Tables \ref{tab:acc_cifar},\ref{tab:acc_imagenet},\ref{tab:acc_transfer}) and negative log-likelihoods (NLL) (Tables \ref{tab:nll_cifar},\ref{tab:nll_imagenet},\ref{tab:nll_transfer}) all methods and datasets. 
We observe
that SWAG is competitive with SWA, SWA with temperature scaling and SWA-Dropout in 
terms of test accuracy, and typically outperforms all the baselines in terms
of NLL. SWAG-Diagonal is generally inferior to SWAG for log-likelihood,
but outperforms SWA.

In Tables \ref{tab:ece_cifar},\ref{tab:ece_imagenet},\ref{tab:ece_transfer} we additionally report expected calibration error \citep[ECE,][]{naeini2015obtaining},
a metric of calibration of the predictive uncertainties. To compute ECE for
a given model we split the test points into $20$ bins based on the confidence 
of the model, and we compute the absolute value of the difference of the average 
confidence and accuracy within each bin, and average the obtained values over 
all bins. Please refer to \citep[][]{naeini2015obtaining,guo_calibration_2017}
for more details. We observe that SWAG is competitive with temperature scaling
for ECE. Again, SWAG-Diagonal achieves better calibration than SWA, but 
using the low-rank plus diagonal covariance approximation in SWAG leads to
substantially improved performance.

\begin{table*}[!tb]
	\centering
	\caption{ECE for various versions of SWAG, temperature scaling, and MC Dropout on CIFAR-10 and CIFAR-100.}
	\label{tab:ece_cifar}
	\resizebox{\textwidth}{!}{
	\begin{tabular}{@{}lllllll@{}}
		\cmidrule(r){1-7}
		& CIFAR-10               & CIFAR-10               & CIFAR-10               & CIFAR-100              & CIFAR-100              & CIFAR-100              \\ \cmidrule(r){1-7} 
		Model        & VGG-16                 & PreResNet-164          & WideResNet28x10        & VGG-16                 & PreResNet-164          & WideResNet28x10        \\ \cmidrule{1-7}
		SGD          & $0.0483\pm0.0022$      & $0.0255\pm0.0009$      & $0.0166\pm0.0007$      & $0.1870\pm0.0014$      & $0.1012\pm0.0009$      & $0.0479\pm0.0010$      \\ 
		SWA          & $0.0408\pm0.0019$      & $0.0203\pm0.0010$      & $0.0087\pm0.0002$      & $0.1514\pm0.0032$      & $0.0700\pm0.0056$      & $0.0684\pm0.0022$      \\ 
		SWAG-Diag    & $0.0267\pm0.0025$      & $0.0082\pm0.0008$      & $\bm{0.0047}\pm0.0013$ & $0.0819\pm0.0021$      & $0.0239\pm0.0047$      & $0.0322\pm0.0018$      \\ 
		SWAG         & $0.0158\pm0.0030$      & $\bm{0.0053}\pm0.0004$ & $0.0088\pm0.0006$      & $0.0395\pm0.0061$      & $0.0587\pm0.0048$      & $\bm{0.0113}\pm0.0020$ \\ 
		KFAC-Laplace & $0.0094\pm0.0005$      & $0.0092\pm0.0018$      & $0.0060\pm0.0003$      & $0.0778\pm0.0054$      & $\bm{0.0158}\pm0.0014$ & $0.0379\pm0.0047$      \\ 
		SWA-Dropout  & $0.0284\pm0.0036$      & $0.0162\pm0.0000$      & $0.0094\pm0.0014$      & $0.1108\pm0.0181$      &           *             & $0.0574\pm0.0028$      \\ 
		SWA-Temp     & $0.0366\pm0.0063$      & $0.0172\pm0.0010$      & $0.0080\pm0.0007$      & $\bm{0.0291}\pm0.0097$ & $0.0175\pm0.0037$      & $0.0220\pm0.0007$      \\ 
		SGLD         & $\bm{0.0082}\pm0.0012$ & $0.0251\pm0.0012$      & $0.0192\pm0.0007$      & $0.0424\pm0.0029$      & $0.0363\pm0.0008$      & $0.0296\pm0.0008$      \\ \cmidrule{1-7}
	\end{tabular}
}
\end{table*}

\begin{table*}[!tb]
	\centering
	\caption{ECE on ImageNet.}
	\label{tab:ece_imagenet}
	\begin{tabular}{@{}lll@{}}
		\toprule
		Model     & DenseNet-161           & ResNet-152             \\ \midrule
		SGD       & $0.0545\pm0.0000$      & $0.0478\pm0.0000$      \\
		SWA       & $0.0509\pm0.0000$      & $0.0605\pm0.0000$      \\
		SWAG-Diag & $0.0459\pm0.0000$      & $0.0566\pm0.0000$      \\
		SWAG      & $0.0204\pm0.0000$      & $0.0279\pm0.0000$      \\
		SWA-Temp  & $\bm{0.0190}\pm0.0000$ & $\bm{0.0183}\pm0.0000$ \\ \bottomrule
	\end{tabular}
\end{table*}

\begin{table*}[!tb]
	\centering
	\caption{ECE on CIFAR10 to STL 10.}
	\label{tab:ece_transfer}
	\begin{tabular}{@{}llll@{}}
		\toprule
		Model       & VGG-16                 & PreResNet-164          & WideResNet28x10        \\ \midrule
		SGD         & $0.2149\pm0.0027$      & $0.1758\pm0.0000$      & $0.1561\pm0.0000$      \\
		SWA         & $0.2082\pm0.0056$      & $0.1739\pm0.0000$      & $0.1413\pm0.0000$      \\
		SWAG-Diag   & $0.1719\pm0.0075$      & $0.1312\pm0.0000$      & $0.1241\pm0.0000$      \\
		SWAG        & $\bm{0.1463}\pm0.0075$ & $\bm{0.1110}\pm0.0000$ & $\bm{0.1017}\pm0.0000$ \\
		SWA-Dropout & $0.1803\pm0.0024$      &                        & $0.1421\pm0.0000$      \\
		SWA-Temp    & $0.2089\pm0.0055$      & $0.1646\pm0.0000$      & $0.1371\pm0.0000$      \\ \bottomrule
	\end{tabular}
\end{table*}

\begin{table*}[!tb]
	\caption{NLL on CIFAR10 and CIFAR100.}
	\label{tab:nll_cifar}
	\resizebox{\textwidth}{!}{%
	\begin{tabular}{@{}lllllll@{}}
		\toprule
		Dataset & \multicolumn{3}{|c}{CIFAR-10} &   \multicolumn{3}{|c}{CIFAR-100}  \\ \midrule
		Model                         & VGG-16                        & PreResNet-164          & WideResNet28x10        & VGG-16                         & PreResNet-164          & WideResNet28x10        \\ \midrule
		SGD                           & $0.3285\pm0.0139$             & $0.1814\pm0.0025$      & $0.1294\pm0.0022$      & $1.7308\pm0.0137$              & $0.9465\pm0.0191$      & $0.7958\pm0.0089$      \\
		SWA                           & $0.2621\pm0.0104$             & $0.1450\pm0.0042$      & $0.1075\pm0.0004$      & $1.2780\pm0.0051$              & $0.7370\pm0.0265$      & $0.6684\pm0.0034$      \\
		SWAG-Diag                     & $0.2200\pm0.0078$             & $0.1251\pm0.0029$      & $0.1077\pm0.0009$      & $1.0163\pm0.0032$              & $0.6837\pm0.0186$      & $0.6150\pm0.0029$      \\
		SWAG                          & $0.2016\pm0.0031$             & $\bm{0.1232}\pm0.0022$ & $0.1122\pm0.0009$      & $0.9480\pm0.0038$              & $\bm{0.6595}\pm0.0019$ & $\bm{0.6078}\pm0.0006$ \\
		KFAC-Laplace                  & $0.2252\pm0.0032$             & $0.1471\pm0.0012$      & $0.1210\pm0.0020$      & $1.1915\pm0.0199$              & $0.7881\pm0.0025$      & $0.7692\pm0.0092$      \\
		SWA-Dropout                   & $0.2328\pm0.0049$             & $0.1270\pm0.0000$      & $0.1094\pm0.0021$      & $1.1872\pm0.0524$              &                        & $0.6500\pm0.0049$      \\
		SWA-Temp                      & $0.2481\pm0.0245$             & $0.1347\pm0.0038$      & $\bm{0.1064}\pm0.0004$ & $1.0386\pm0.0126$              & $\bm{0.6770}\pm0.0191$ & $0.6134\pm0.0023$      \\
		SGLD                          & $\bm{0.2001} \pm 0.0059$      & $0.1418 \pm 0.0005$    & $0.1289 \pm 0.0009$    & $0.9699 \pm 0.0057$            & $0.6981 \pm 0.0052$    & $0.678 \pm 0.0022$ \\
		SGD-Ens                       & $0.1881 \pm 0.002$            & $0.1312 \pm 0.0023$    & $0.1855 \pm 0.0014$    & $\bm{0.8979} \pm 0.0065$       & $0.7839 \pm 0.0046$    & $0.7655 \pm 0.0026$ \\
		 \bottomrule   
	\end{tabular}
}
\end{table*}

\begin{table*}[!tb]
	\centering
	\caption{NLL on ImageNet.}
	\label{tab:nll_imagenet}
	\begin{tabular}{@{}lll@{}}
		\toprule
		Model     & DenseNet-161           & ResNet-152             \\ \midrule
		SGD       & $0.9094\pm0.0000$      & $0.8716\pm0.0000$      \\
		SWA       & $0.8655\pm0.0000$      & $0.8682\pm0.0000$      \\
		SWAG-Diag & $0.8559\pm0.0000$      & $0.8584\pm0.0000$      \\
		SWAG      & $\bm{0.8303}\pm0.0000$ & $\bm{0.8205}\pm0.0000$ \\
		SWA-Temp  & $0.8359\pm0.0000$      & $0.8226\pm0.0000$      \\ \bottomrule
	\end{tabular}
\end{table*}

\begin{table*}[!tb]
	\centering
	\caption{NLL when transferring from CIFAR10 to STL10.}
	\label{tab:nll_transfer}
	\begin{tabular}{@{}llll@{}}
		\toprule
		Model       & VGG-16                 & PreResNet-164          & WideResNet28x10        \\ \midrule
		SGD         & $1.6528\pm0.0390$      & $1.4790\pm0.0000$      & $1.1308\pm0.0000$      \\
		SWA         & $1.3993\pm0.0502$      & $1.3552\pm0.0000$      & $1.0047\pm0.0000$      \\
		SWAG-Diag   & $1.2258\pm0.0446$      & $1.0700\pm0.0000$      & $0.9340\pm0.0000$      \\
		SWAG        & $\bm{1.1402}\pm0.0342$ & $\bm{0.9706}\pm0.0000$ & $\bm{0.8710}\pm0.0000$ \\
		SWA-Dropout & $1.3133\pm0.0000$      &                        & $0.9914\pm0.0000$      \\
		SWA-Temp    & $1.4082\pm0.0506$      & $1.2228\pm0.0000$      & $0.9706\pm0.0000$      \\ \bottomrule
	\end{tabular}
\end{table*}

\begin{table*}[!tb]
	\centering
	\caption{Accuracy on CIFAR-10 and CIFAR-100.}
	\label{tab:acc_cifar}
	\resizebox{\textwidth}{!}{%
		\begin{tabular}{@{}lllllll@{}}
			\toprule
			Dataset & \multicolumn{3}{|c}{CIFAR-10} &   \multicolumn{3}{|c}{CIFAR-100}  \\ \midrule
			Model        & VGG-16              & PreResNet-164       & WideResNet28x10     & VGG-16              & PreResNet-164       & WideResNet28x10     \\ \midrule
			SGD          & $93.17\pm0.14$      & $95.49\pm0.06$      & $96.41\pm0.10$      & $73.15\pm0.11$      & $78.50\pm0.32$      & $80.76\pm0.29$      \\
			SWA          & $93.61\pm0.11$      & $96.09\pm0.08$      & $\bm{96.46}\pm0.04$ & $74.30\pm0.22$      & $\bm{80.19}\pm0.52$ & $82.40\pm0.16$      \\
			SWAG-Diag    & $\bm{93.66}\pm0.15$ & $96.03\pm0.10$      & $96.41\pm0.05$      & $74.68\pm0.22$      & $80.18\pm0.50$      & $\bm{82.40}\pm0.09$ \\
			SWAG         & $93.60\pm0.10$      & $96.03\pm0.02$      & $96.32\pm0.08$      & $\bm{74.77}\pm0.09$ & $79.90\pm0.50$      & $82.23\pm0.19$      \\
			KFAC-Laplace & $92.65\pm0.20$      & $95.49\pm0.06$      & $96.17\pm0.00$      & $72.38\pm0.23$      & $78.51\pm0.05$      & $80.94\pm0.41$      \\
			SWA-Dropout  & $93.23\pm0.36$      & $\bm{96.18}\pm0.00$ & $96.39\pm0.09$      & $72.50\pm0.54$      &                     & $82.30\pm0.19$      \\
			SWA-Temp     & $93.61\pm0.11$      & $96.09\pm0.08$      & $96.46\pm0.04$      & $74.30\pm0.22$      & $80.19\pm0.52$      & $82.40\pm0.16$      \\
			SGLD         & $93.55 \pm 0.15$    & $95.55\pm0.04$      & $95.89 \pm 0.02$    & $74.02 \pm 0.30$    & $80.09\pm0.05$      & $80.94 \pm 0.17$    \\ \bottomrule
		\end{tabular}
	}
\end{table*}

\begin{table*}[!tb]
	\centering
	\caption{Accuracy on ImageNet.}
	\label{tab:acc_imagenet}
	\begin{tabular}{@{}lll@{}}
		\toprule
		Model     & DenseNet-161        & ResNet-152          \\ \midrule
		SGD       & $77.79\pm0.00$      & $78.39\pm0.00$      \\
		SWA       & $\bm{78.60}\pm0.00$ & $78.92\pm0.00$      \\
		SWAG-Diag & $78.59\pm0.00$      & $78.96\pm0.00$      \\
		SWAG      & $78.59\pm0.00$      & $\bm{79.08}\pm0.00$ \\
		SWA-Temp  & $78.60\pm0.00$      & $78.92\pm0.00$  \\ \bottomrule
	\end{tabular}
\end{table*}

\begin{table*}[!tb]
	\centering
	\caption{Accuracy when transferring from CIFAR-10 to STL-10.}
	\label{tab:acc_transfer}
	\begin{tabular}{@{}llll@{}}
		\toprule
		Model       & VGG-16              & PreResNet-164       & WideResNet28x10     \\ \midrule
		SGD         & $\bm{72.42}\pm0.07$ & $75.56\pm0.00$      & $76.75\pm0.00$      \\
		SWA         & $71.92\pm0.01$      & $\bm{76.02}\pm0.00$ & $\bm{77.50}\pm0.00$ \\
		SWAG-Diag   & $72.09\pm0.04$      & $75.95\pm0.00$      & $77.26\pm0.00$      \\
		SWAG        & $72.19\pm0.06$      & $75.88\pm0.00$      & $77.09\pm0.00$      \\
		SWA-Dropout & $71.45\pm0.11$      &                     & $76.91\pm0.00$      \\
		SWA-Temp    & $71.92\pm0.01$      & $76.02\pm0.00$      & $77.50\pm0.00$      \\ \bottomrule
	\end{tabular}

\end{table*}


\section{Language Modeling}
\label{sec:lstm}

We evaluate SWAG using standard Penn Treebank and WikiText-2
benchmark language modeling datasets. Following \citep{reglstm} we use a 3-layer LSTM model with 
1150 units in the hidden layer and an embedding of size 400; we apply
dropout, weight-tying, activation regularization (AR) and temporal activation 
regularization (TAR) techniques. We follow \citep{reglstm} for specific
hyper-parameter settings such as dropout rates for different types of layers.
We train all models for language modeling tasks and evaluate validation and 
test perplexity. For SWA and SWAG we pre-train the models using standard SGD
for $500$ epochs, and then run the model for $100$ more 
epochs to estimate the mean $\theta_\swa$ and covariance $\Sigma$ in SWAG.
For this experiment we introduce a small change to SWA and SWAG: to estimate
the mean $\theta_\swa$ we average weights after each mini-batch of data
rather than once per epoch, as we found more frequent averaging to 
greatly improve performance. After SWAG distribution is constructed we
sample and ensemble $30$ models from this distribution. We use rank-$10$ for
the low-rank part of the covariance matrix of SWAG distribution.

\section{Regression}
\label{sec:uci}

For the small UCI regression datasets, we use the architecture from \citet{wu2018fixing} with one hidden layer with 50 units, training for 50 epochs (starting SWAG at epoch 25) and using 20 repetitions of 90/10 train test splits.
We fixed a single seed for tuning before using 20 different seeds for the results in the paper.
 
We use SGD\footnote{Except for concrete where we use Adam due to convergence issues.}, manually tune learning rate and weight decay, and use batch size of $N / 10$ where $N$ is the dataset size. 
All models predict heteroscedastic uncertainty (i.e. output a variance). In Table \ref{tab:small_ll}, we compare subspace inference methods to deterministic VI (DVI, \citet{wu2018fixing}) and deep Gaussian processes with expectation propagation (DGP1-50  \citet{bui2016deep}). SWAG outperforms DVI and the other methods on three of the six datasets and is competitive on the other three despite its vastly reduced computational time (the same as SGD whereas DVI is known to be ~300x slower).
Additionally, we note the strong performance of well-tuned SGD as a baseline against the other approximate inference methods, as it consistently performs nearly as well as both SWAG and DVI.

Finally, in Table \ref{tab:small_ll}, we compare the calibration (coverage of the 95\% credible sets of SWAG and 95\% confidence regions of SGD) of both SWAG and SGD.
Note that neither is ever too over-confident (far beneath 95\% coverage) and that SWAG is considerably better calibrated on four of the six datasets.

\begin{table*}[!tb]
	\centering
	\caption{Unnormalized test log-likelihoods on small UCI datasets for proposed methods, as well as direct comparisons to the numbers reported in deterministic variational inference (DVI, \citet{wu2018fixing}) and Deep Gaussian Processes with expectation propagation (DGP1-50, \citet{bui2016deep}), and variational inference (VI) with the re-parameterization trick \citep{kingma2015variational}. * denotes reproduction from \citep{wu2018fixing}. Note that SWAG wins on two of the six datasets, and that SGD serves as a strong baseline throughout.}
	\label{tab:small_ll}
	\resizebox{\textwidth}{!}{%
		\begin{tabular}{lllcccccccc}
			\toprule
			dataset &           N &   D &       SGD &   SWAG  & DVI* & DGP1-50* & VI* & SGLD* & PBP* \\
			\midrule
			boston   &  506        & 13     &  -2.536 $\pm$ 0.240 &  -2.469 $\pm$ 0.183 & -2.41 $\pm$ 0.02 & \textbf{-2.33} $\pm$ 0.06 & -2.43 $\pm$0.03 & -2.40 $\pm$ 0.05 & -2.57 $\pm$ 0.09 \\
			concrete & 1030     &  8     &  \textbf{-3.02} $\pm$ 0.126 &  -3.05 $\pm$ 0.1  &  -3.06 $\pm$ 0.01  &     -3.13 $\pm$  0.03 & -3.04 $\pm$0.02  & -3.08 $\pm$ 0.03  & -3.16 $\pm$ 0.02 \\
			energy  &    768       &   8  &  -1.736 $\pm$ 1.613 &  -1.679 $\pm$ 1.488      &  \textbf{-1.01} $\pm$ 0.06   &   -1.32 $\pm$ 0.03  & -2.38 $\pm$0.02 & -2.39 $\pm$ 0.01 & -2.04 $\pm$ 0.02 \\
			naval   &     11934   &   16  &   6.567 $\pm$ 0.185 &   \textbf{6.708 $\pm$ 0.105}       &  6.29 $\pm$ 0.04    &  3.60 $\pm$ 0.33  & 5.87 $\pm$0.29 & 3.33 $\pm$ 0.01 & 3.73 $\pm$ 0.01  \\
			yacht    &      308     & 6   &  -0.418 $\pm$ 0.426 &  \textbf{-0.404 $\pm$ 0.418}     &   -0.47 $\pm$ 0.03   &  -1.39 $\pm$ 0.14 & -1.68 $\pm$0.04 & -2.90 $\pm$ 0.01 & -1.63 $\pm$ 0.02 \\
			power & 9568 & 4 & -2.772 $\pm$ 0.04 & -2.775 $\pm$ 0.038 & -2.80 $\pm$ 0.00 & -2.81 $\pm$ 0.01 & \textbf{-2.66} $\pm$ 0.01 & -2.67 $\pm$ 0.00 & -2.84 $\pm$ 0.01 \\
			\bottomrule
		\end{tabular}
	}
\end{table*}

\begin{table*}[!tb]
	\centering
	\caption{Calibration on small-scale UCI datasets. Bolded numbers are those closest to 0.95 \%the predicted coverage).}
	\label{tab:small_ca}
		\begin{tabular}{lllllll}
			\toprule
			{}       & N     & D    &       SGD &               SWAG \\
			\midrule
			boston   & 506   & 13   &    0.913 $\pm$ 0.039 &  \textbf{0.936} $\pm$ 0.036     \\
			concrete & 1030  & 8    &    0.909 $\pm$ 0.032 &  \textbf{0.930} $\pm$ 0.023  \\
			energy   & 768   & 8    & 0.947 $\pm$ 0.026 &   \textbf{0.951} $\pm$ 0.027            \\
			naval    & 11934 & 16   &    \textbf{0.948} $\pm$ 0.051 &    0.967 $\pm$ 0.008        \\
			yacht    & 308   & 6    &        0.895 $\pm$ 0.069  &  \textbf{0.898} $\pm$ 0.067    \\
			power & 9568 & 4 & \textbf{0.956} $\pm$ 0.006 & 0.957 $\pm$ 0.005 \\
			\bottomrule
		\end{tabular}
\end{table*}

\section{Classification Experimental Details and Parameters}\label{app:details}

In this section we describe all of the architectures and hyper-parameters we 
use in Sections \ref{sec:exp_calibration}, \ref{sec:exp_out}.

On ImageNet we use architecture implementations and pre-trained weights from 
{\small\url{https://github.com/pytorch/vision/tree/master/torchvision}}.
For the experiments on CIFAR datasets  we adapted the following implementations:
\begin{itemize}
	\item VGG-$16$: {\small\url{https://github.com/pytorch/vision/blob/master/torchvision/models/vgg.py}}
	\item Preactivation-ResNet-$164$: {\small \url{https://github.com/bearpaw/pytorch-classification/blob/master/models/cifar/preresnet.py}}
	\item WideResNet28x10: {\small \url{https://github.com/meliketoy/wide-resnet.pytorch/blob/master/networks/wide_resnet.py}}
\end{itemize}

For all datasets and architectures we use the same piecewise constant learning
rate schedule and weight decay as in \citet{izmailov_averaging_2018}, except
we train Pre-ResNet for $300$ epochs and start averaging after epoch $160$
in SWAG and SWA.
For all of the methods we are using our own implementations in PyTorch. 
We describe the hyper-parameters for all experiments for each model:

\paragraph{SWA} We use the same hyper-parameters as \citet{izmailov_averaging_2018}
on CIFAR datasets. On ImageNet we used a constant learning rate of $10^{-3}$
instead of the cyclical schedule, and averaged $4$ models per epoch.
We adapt the code from {\small\url{https://github.com/timgaripov/swa}} for our
implementation of SWA.

\paragraph{SWAG} In all experiments we use rank $K = 20$ and use $30$ weight samples
for Bayesian model averaging. We re-use all the other hyper-parameters from 
SWA.

\paragraph{KFAC-Laplace} For our implementation we adapt the code for KFAC Fisher
approximation from {\small \url{https://github.com/Thrandis/EKFAC-pytorch}} and 
implement our own code for sampling. Following \citep{ritter_scalable_2018} we
tune the scale of the approximation on validation set for every model and dataset.

\paragraph{MC-Dropout} In order to implement MC-dropout we add dropout layers
before each weight layer and sample $30$ different dropout masks for Bayesian
model averaging at inference time. To choose the dropout rate, we ran the models
with dropout rates in the set $\{0.1, 0.05, 0.01\}$ and chose the one that
performed best on validation data. For both VGG-16 and WideResNet28x10 we found 
that dropout rate of $0.05$ worked best and used it in all experiments. On
PreResNet-164 we couldn't achieve reasonable performance with any of the 
three dropout rates, which has been reported from the work of \citet{he_deep_2016}. 
We report the results for MC-Dropout in combination with
both SWA (SWA-Drop) and SGD (SGD-Drop) training.

\paragraph{Temperature Scaling}
For SWA and SGD solutions we picked the optimal temperature by minimizing
negative log-likelihood on validation data, adapting the code from 
{\small\url{https://github.com/gpleiss/temperature_scaling}}.

\paragraph{SGLD} 
We initialize SGLD from checkpoints pre-trained with SGD. We run SGLD for
$100$ epochs on WideResNet and for $150$ epochs on PreResNet-156. We use 
the learning rate schedule of \citep{welling2011bayesian}:
\[
\eta_t = \frac{\eta_0}{(\eta_1 + t)^{0.55}}.
\]
We tune constants $a, b$ on validation. For WideResNet we use $a=38.0348$, $b=13928.7$
and for PreResNet we use $a=40.304$, $b=15476.4$; these values are selected so that
the initial learning rate is $0.2$ and final learning rate is $0.1$. 
We also had to rescale the noise in the
gradients by a factor of $5 \cdot 10^{-4}$ compared to \citep{welling2011bayesian}. 
Without this rescaling we found that even with learning rates on the scale of 
$10^{-7}$ SGD diverged. We note that noise rescaling is commonly used with 
stochastic gradient MCMC methods (see e.g. the implementation of \citep{zhang2019cyclical}).

On CIFAR datasets for tuning hyper-parameters we used 
the last $5000$ training data points as a validation set. On ImageNet
we used $5000$ of test data points for validation.
On the transfer task for CIFAR10 to STL10, we report accuracy on all 10 STL10 classes even though frogs are not a part of the STL10 test set (and monkeys are not a part of the CIFAR10 training set).

\end{document}